\documentclass[10pt,twocolumn,letterpaper]{article}

\usepackage[accsupp]{axessibility} 
\usepackage[pagenumbers]{cvpr} 

\usepackage[dvipsnames]{xcolor}

\usepackage{colortbl}

\definecolor{cvprblue}{rgb}{0.21,0.49,0.74}
\definecolor{baselinecolor}{gray}{.92}
\newcommand{\basecell}[1]{\cellcolor{baselinecolor}}
\usepackage{multirow}
\usepackage{booktabs}
\usepackage{graphicx}
\usepackage{math}
  \usepackage{tablefootnote}
 \usepackage{footnote}
\definecolor{cvprblue}{rgb}{0.21,0.49,0.74}
\usepackage[pagebackref,breaklinks,colorlinks,citecolor=cvprblue]{hyperref}
\usepackage{tikz}
\usepackage{pgfplots}

\newcommand{\method}{CLOUDS\xspace}
\newcommand{\miou}{mIoU\xspace}

\title{Collaborating Foundation Models for Domain Generalized\\ Semantic Segmentation}

\author{Yasser Benigmim\textsuperscript{1,2}\quad
Subhankar Roy\textsuperscript{3}\quad
Slim Essid\textsuperscript{1}\quad
Vicky Kalogeiton\textsuperscript{2} \quad
Stéphane Lathuilière\textsuperscript{1} \\
\textsuperscript{1}\,LTCI, Télécom-Paris, Institut Polytechnique de Paris \\
\textsuperscript{2}\,LIX, Ecole Polytechnique, CNRS,  Institut Polytechnique de Paris, 
\textsuperscript{3}\,University of Aberdeen\\
{\tt\small yasser.benigmim@telecom-paris.fr}
}

\begin{document}
\maketitle
\begin{abstract}
Domain Generalized Semantic Segmentation (DGSS) deals with training a model on a labeled source domain with the aim of generalizing to unseen domains during inference. Existing DGSS methods typically effectuate robust features by means of Domain Randomization (DR). Such an approach is often limited as it can only account for style diversification and not content. In this work, we take an orthogonal approach to DGSS and propose to use an assembly of \textbf{C}o\textbf{L}laborative F\textbf{OU}ndation models for \textbf{D}omain Generalized Semantic \textbf{S}egmentation (\textbf{CLOUDS}). In detail, \method is a framework that integrates Foundation Models of various kinds: (i) CLIP backbone for its robust feature representation, (ii) Diffusion Model to diversify the content, thereby covering various modes of the possible target distribution, and (iii) Segment Anything Model (SAM) for iteratively refining the predictions of the segmentation model. Extensive experiments show that our \method excels in adapting from synthetic to real DGSS benchmarks and under varying weather conditions, notably outperforming prior methods by $5.6\%$ and $6.7\%$ on averaged \miou, respectively. Our code is available at \url{https://github.com/yasserben/CLOUDS}
\end{abstract}    

\section{Introduction}
\label{sec:intro}
Deep Neural Networks have showcased remarkable ability in scene understanding tasks like Semantic Segmentation (SS)~\cite{chen2017deeplab,wang2020deep,xie2021segformer}, when the training and test distribution are the same. This dependency reveals a significant vulnerability: their performance substantially diminishes when encountering domain shifts~\cite{torralba2011unbiased}, highlighting a fundamental challenge in generalizing these networks to unseen domains~\cite{saenko2010adapting, luo2019taking, motiian2017few}.
To address this, Domain Generalized Semantic Segmentation (DGSS) has gained prominence~\cite{huang2019iterative,zhao2022style, kim2023texture,jiang2023domain}. DGSS aims to develop models that leverage a source-annotated dataset while remaining effective across unseen domains, thus overcoming the limitations of traditional DNNs in handling unseen environments. 

\begin{figure}[!t]
    \centering
    \includegraphics[width=0.99\linewidth]{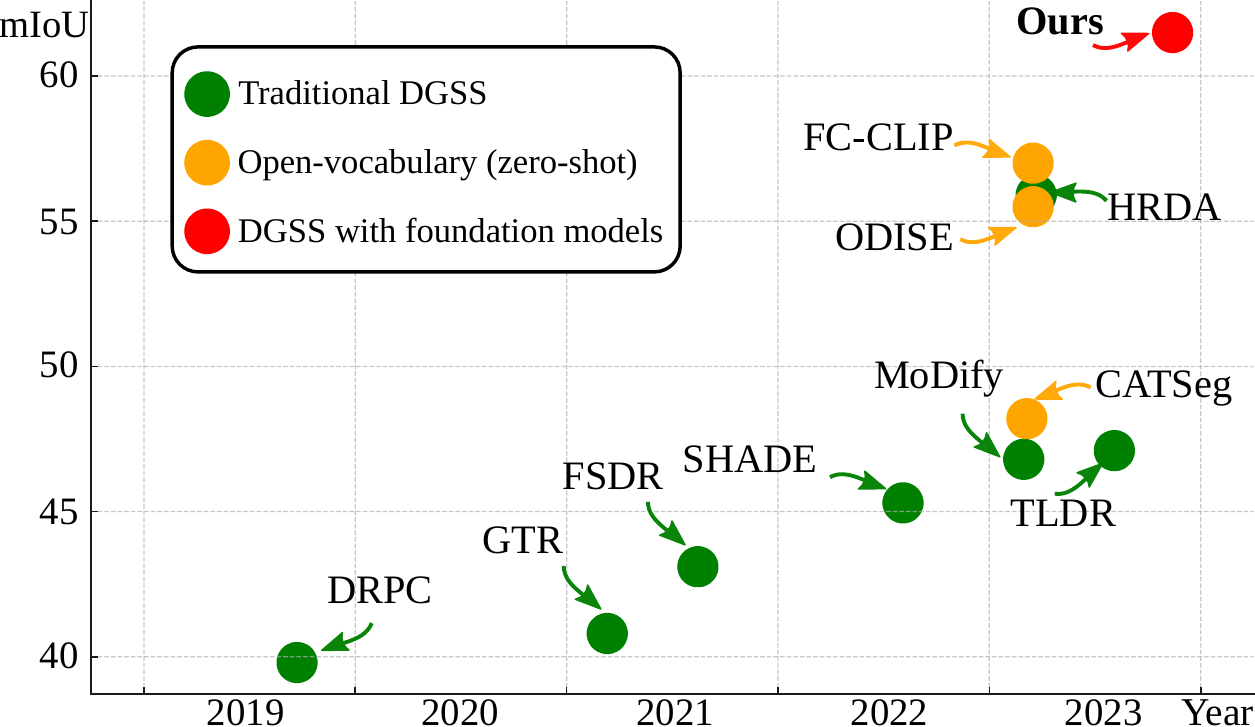}
    \caption{Performance over time by various methods on
the GTA $\rightarrow$ \{Cityscapes, BDD, Mapillary\} benchmark. Recent open-vocabulary approaches, like FC-CLIP, are shown to excel in zero-shot learning and surpass traditional domain generalization methods trained in closed-set scenarios, challenging the relevance of the DGSS setting. \method, by harnessing multiple foundation models, demonstrates its ability to effectively utilize the source dataset, thereby outperforming both conventional DGSS and open-vocabulary methods.}
    \label{fig:teaser}
\end{figure}

\begin{figure*}[!t]
    \centering
        \begin{subfigure}[b]{0.19\linewidth}
    \includegraphics[width=\linewidth]{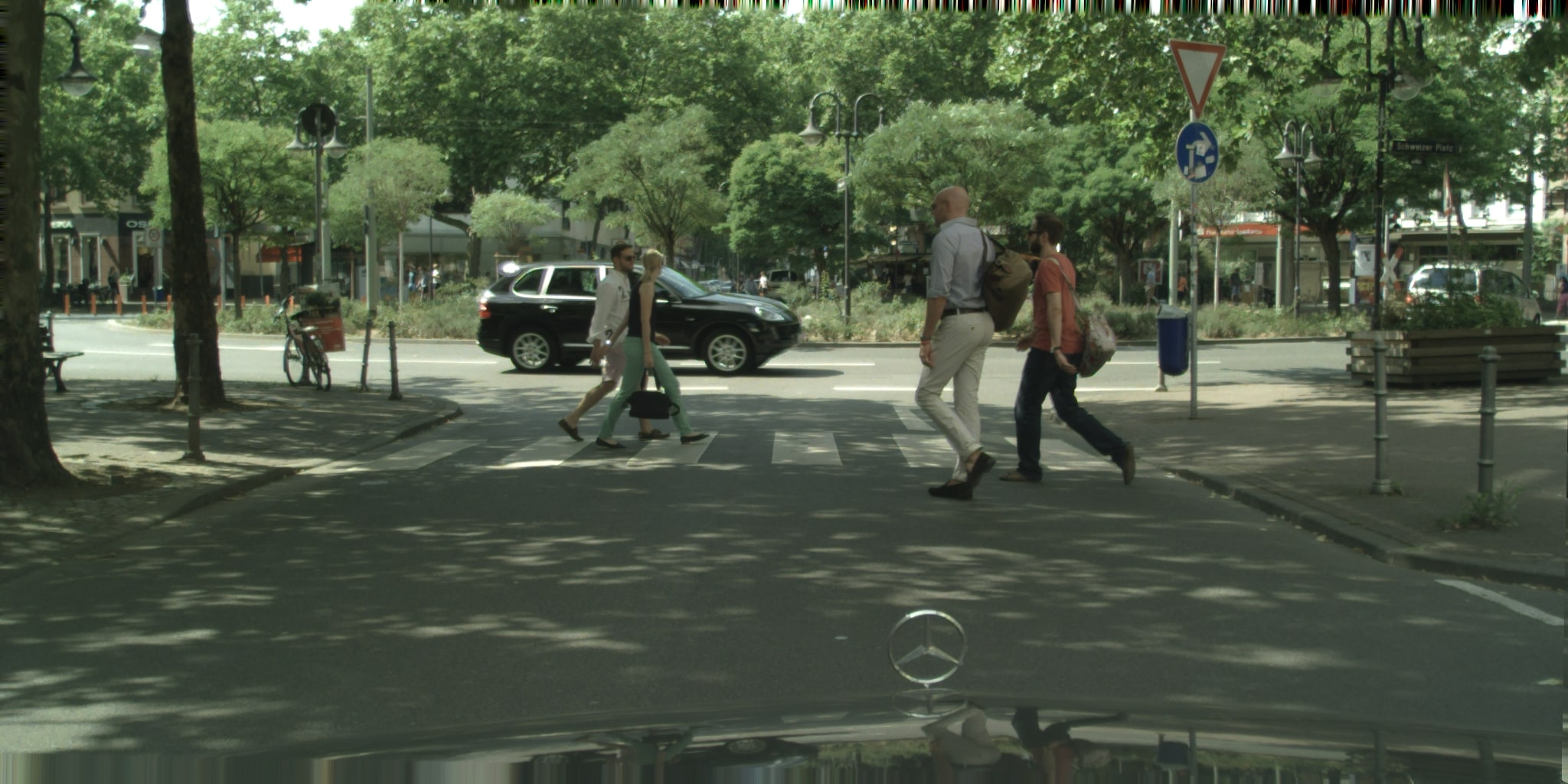}
    \subcaption{Input image}
    \label{fig:img_rgb}
  \end{subfigure}
  \hfill
  \begin{subfigure}[b]{0.19\linewidth}
    \includegraphics[width=\linewidth]{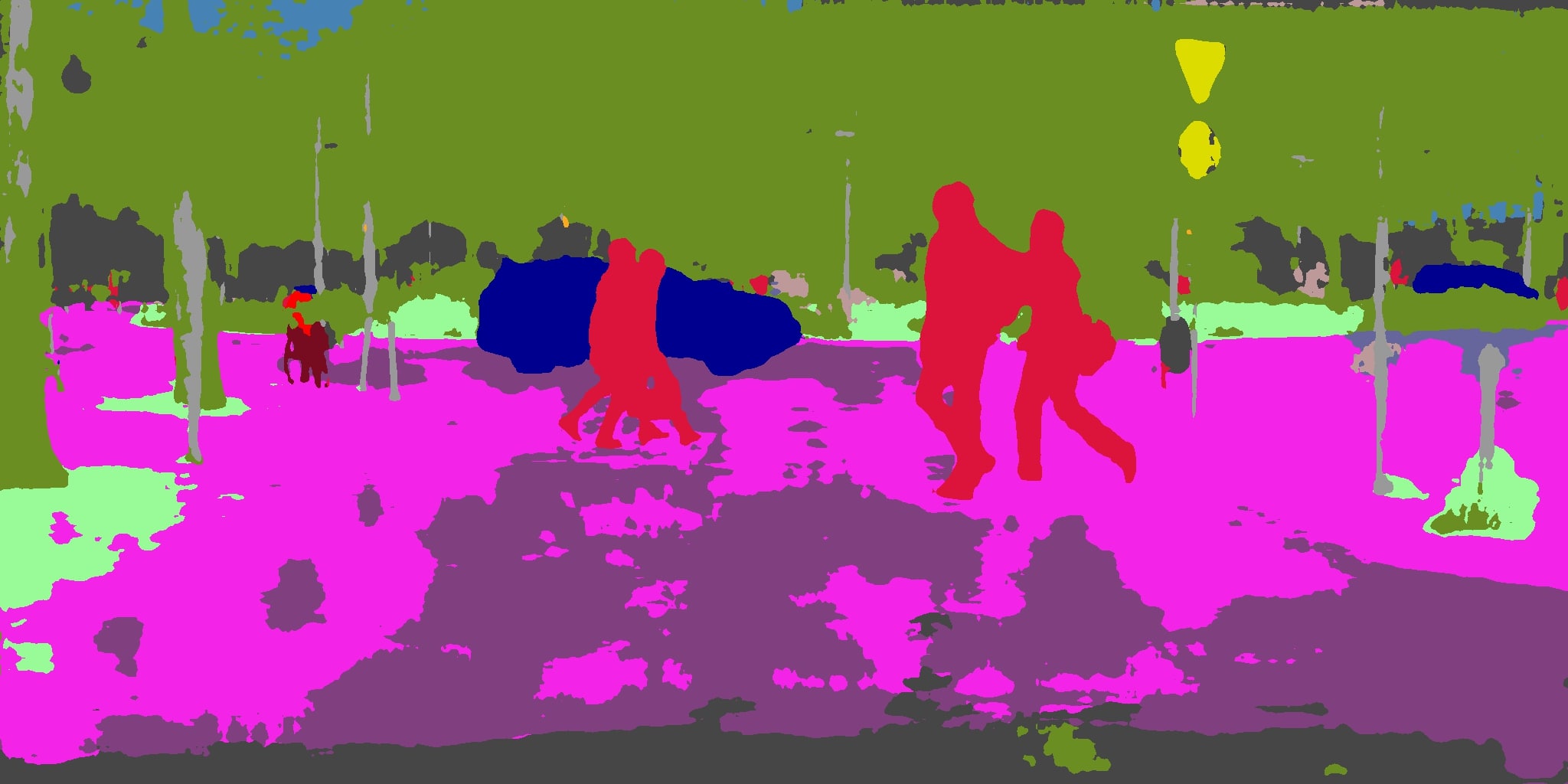}
    \subcaption{SHADE~\cite{zhao2022style}}
    \label{fig:pred_shade}
  \end{subfigure}
  \hfill
  \begin{subfigure}[b]{0.19\linewidth}
    \includegraphics[width=\linewidth]{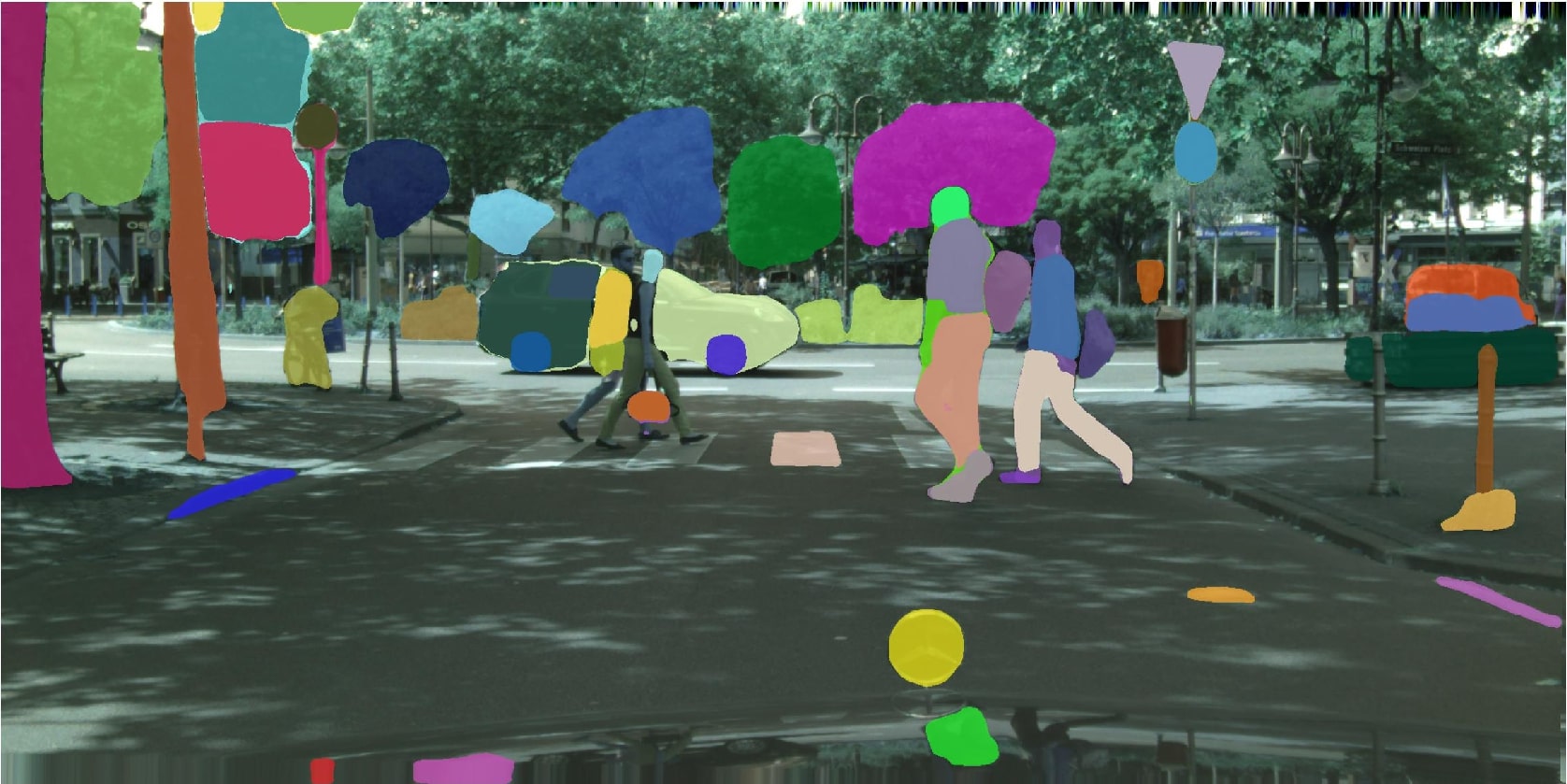}
    \subcaption{SAM~\cite{kirillov2023segment}}
    \label{fig:pred_sam}
  \end{subfigure}
  \hfill
  \begin{subfigure}[b]{0.19\linewidth}
    \includegraphics[width=\linewidth]{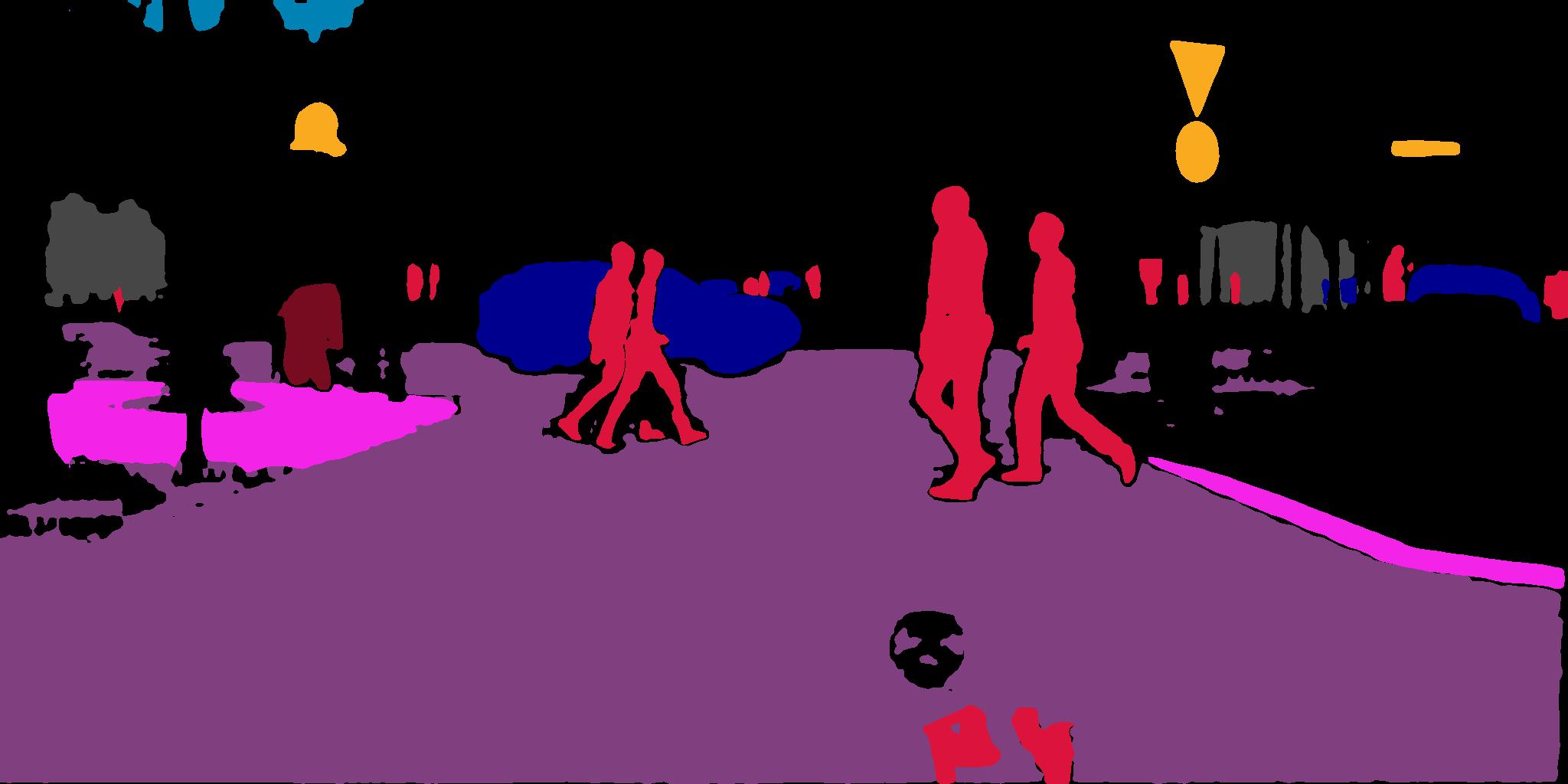}
    \subcaption{GroundingSAM~\cite{liu2023grounding,kirillov2023segment}}
    \label{fig:pred_grsam}
  \end{subfigure}
  \hfill
  \begin{subfigure}[b]{0.19\linewidth}
    \includegraphics[width=\linewidth]{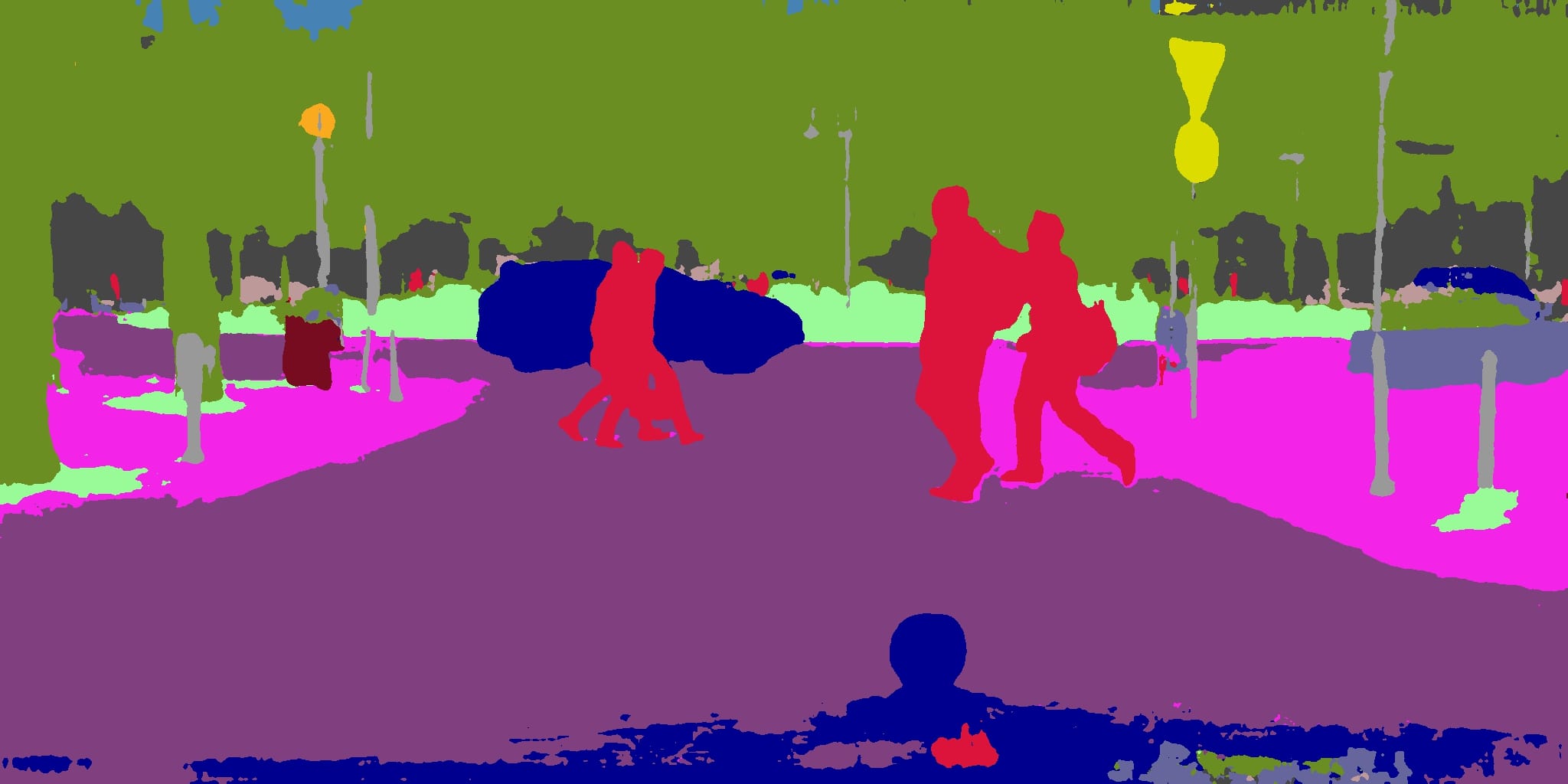}
    \subcaption{\textbf{\method} (Ours)}
    \label{fig:image5}
  \end{subfigure}
    \caption{\textbf{Qualitative comparison at inference}: (a) Input image, (b) SHADE, a traditional style diversification DGSS method, (c) SAM, a foundation model that predicts precise class-agnostic maps, (d) GroundingSAM that leverages SAM and text-prompts to output semantic maps, and (d) Our proposed \method that leverages an assembly of foundation models to predict high-quality semantic maps.}
    \label{fig:pred_ours}
\end{figure*}

Recently, the advent of large-scale pretrained models, often referred to as Foundation Models (FMs)~\cite{bommasani2021opportunities,radford2021learning,rombach2022high,kirillov2023segment,touvron2023llama}, have brought a paradigm shift in computer vision tasks. The FMs comprise of contrastively trained image classification models (e.g., CLIP \cite{radford2021learning}, ALIGN \cite{jia2021scaling}), text-conditioned generative models (e.g., Stable Diffusion \cite{rombach2022high}, DALL-E \cite{ramesh2022hierarchical}), vision transformer-based segmentation model trained on mammoth dataset (e.g., Segment Anything Model (SAM) \cite{kirillov2023segment}), to name a few. Of particular interest to SS, SAM is a promptable segmentation model that accepts as input an image and geometric prompts (points, scribbles, boxes, masks) and outputs class-agnostic masks. Owing to its pre-training on billion-scale dataset~\cite{kirillov2023segment}, SAM has demonstrated excellent performance on varied out-of-distribution tasks namely, medical imaging \cite{ma2023segment,wu2023medical,he2023accuracy}, crater detection \cite{giannakis2023deep}, and so on. In summary, large-scale pre-training holds promise for DGSS as robustness is greatly improved when a model is trained on large datasets that cover various possible distributions \cite{zara2023unreasonable}.

While the FMs present an excellent solution for addressing DGSS, they have not been adopted so far in the literature. A great majority of DGSS methods rely on the technique of Domain Randomization (DR)~\cite{qiao2020learning,volpi2018generalizing,xu2021fourier,zhao2022style}, where the goal is to diversify the labeled source domain images by photometric and geometric image transformations. Examples of such transformations include style diversification~\cite{zhao2022style,zhong2022adversarial}, adversarial style augmentation~\cite{zhong2022adversarial}, and difficulty-aware photometric augmentations~\cite{jiang2023domain}, among others~\cite{yue2019domain,kim2023texture}. Albeit effective to some extent, these techniques can not bring content augmentation, resulting in sub-optimal performance (see Fig. \ref{fig:pred_shade}). Moreover, adopting SAM for the task of DGSS is not straightforward as it outputs class-agnostic masks, making it unsuitable for DGSS (see Fig. \ref{fig:pred_sam}). While there are recent works, such as Grounding-SAM \cite{kirillov2023segment,liu2023grounding}, that equip SAM with semantic predictions, it misses out on several important objects (\textit{e.g.}, ``poles") and \textit{stuff} classes (\textit{e.g.}, ``tree", ``sidewalk") (Fig. \ref{fig:pred_grsam}). The lack of semantic awareness can be attributed to ambiguous text prompts or poor vision and text alignment. These results indicate that DGSS is far from solved and accommodating FMs in DGSS is an open question.

Driven by the motivation of narrowing down the research gap, in this work we propose to amalgamate an \textit{assembly} of FMs into a cohesive system to address DGSS. In detail, we propose to exploit: (\textbf{i}) strong representations of contrastive FMs, (\textbf{ii}) diverse content augmentation using generative FMs, and (\textbf{iii}) near accurate class-agnostic mask prediction of SAM and turn them into semantic predictions. Concretely, for the segmentation network we use \textbf{CLIP}~\cite{radford2021learning} as a backbone, serving as a robust feature extractor. To introduce content diversification, we generate synthetic images using a combination of \textbf{Large Language Models} (LLMs)~\cite{touvron2023llama} and text-to-image \textbf{Diffusion Model} (DM)~\cite{rombach2022high}. The LLM generates task-aware diverse textual prompts which then guide the DM in producing photorealistic synthetic images. Since the generated images are unlabeled, we employ self-training strategy \cite{zou2018unsupervised,zou2019confidence,zhang2021prototypical,hoyer2022daformer} where we use the pseudo labels from the teacher network to train the student model. However, as the pseudo labels from the teacher can be noisy, we leverage \textbf{SAM}'s excellent class-agnostic mask predictions to refine the pseudo labels. As shown in Fig. \ref{fig:image5}, the pseudo-label refinement greatly assists in improving the reliability of self-training.

In summary, in this work we make the following contributions: (\textbf{i}) We highlight for the first time in DGSS the importance of content diversification, which is more effective than traditional DGSS methods relying on style diversification; (\textbf{ii}) We propose \method, a system of collaborative Foundation models for DGSS. We run extensive experiments on several DGSS benchmarks and demonstrate its effectiveness. As shown in Fig. \ref{fig:teaser}, our FM-based DGSS method outperforms the traditional DGSS and open-vocabulary segmentation models by a non-trivial margin.

\section{Related Works}
\label{sec:related}

\textbf{Domain Generalization (DG).}
DG focuses on training models robust against domain shift~\cite{volpi2021continual,muandet2013domain}.
In the field of Domain Generalized Semantic Segmentation (DGSS), methods can be classified into two main categories. The first involves integrating tailor-made modules and transformations to explicitly eliminate domain-specific features~\cite{pan2018two,pan2019switchable,choi2021robustnet,peng2022semantic}, such as IBN-Net~\cite{pan2018two} that uses instance normalization blocks and ISW~\cite{choi2021robustnet} that applies whitening transformations.
The second category relies on Domain Randomization (DR) by diversifying image styles~\cite{tang2020selfnorm,zhao2022source,zhao2022style,jiang2023domain,peng2021global}. SHADE \cite{zhao2022style} generates new styles from basis styles of the source domain, and MoDify \cite{jiang2023domain} leverages difficulty-aware photometric augmentations. HRDA~\cite{hoyer2023domain}, designed for Unsupervised Domain Adaptation~\cite{hoyer2022hrda}, shows remarkable performance in DGSS, combining the VIT-based model~\cite{dosovitskiy2020image} and multiple training strategies for SS.
While our approach aligns with DR, it differs in two aspects: (i) {\method} emphasizes data generation over stylization for greater diversification, and (ii) the use of generated data enables the introduction of a self-training strategy which, to our knowledge, has never been used for DGSS.

\paragraph{Foundation Models for Segmentation.}
Foundation models~\cite{bommasani2021opportunities,rombach2022high,radford2021learning,kirillov2023segment} have recently garnered significant interest, and their application across a wide range of downstream tasks constitutes an active area of research.

\noindent \textbf{Diffusion Models (DM)}, a key type among generative models~\cite{creswell2018generative,rombach2022high,touvron2023llama,brown2020language}, stand out in producing photorealistic images, thereby enhancing synthetic dataset generation and boosting vision task performance~\cite{sariyildiz2023fake,Benigmim_2023_CVPR,Niemeijer_2024_WACV}. Yet, this approach remains underexplored in DGSS, a gap our paper aims to address.
Diffusion models have also been effectively used for discriminative tasks~\cite{karazija2023diffusion,xu2023open,gong2023prompting}. In particular, the U-Net architecture~\cite{lin2017feature} of these models, known for its rich and robust image representation, is particularly well suited for pixel-level prediction tasks like semantic segmentation. ODISE~\cite{xu2023open}, for instance, adopts this approach for open-vocabulary semantic segmentation. The study closely related to ours~\cite{gong2023prompting} investigates both DGSS and test-time domain adaptation using a diffusion-based backbone.

\noindent\textbf{CLIP} also offers robust feature representations and image-text alignment, which have been instrumental in various vision applications \cite{xu2023open,dong2023maskclip,gong2023prompting,cho2023promptstyler,yu2023convolutions,cho2023cat}. CLIP has emerged as a crucial component in open-vocabulary segmentation models which segment unseen classes via the introduction of a masked self-distillation mechanism\cite{dong2023maskclip} or a novel cost aggregation layer~\cite{cho2023cat}. Drawing inspiration from this success, our work employs a CNN-based~\cite{lecun1998gradient} CLIP backbone, as recommended in \cite{yu2023convolutions}.

\noindent\textbf{Segment Anything Model (SAM)}~\cite{kirillov2023segment}, a prominent vision foundation model, is trained for promptable segmentation tasks. SAM excels in producing high-quality masks for any segmentation prompt. Various studies \cite{liu2023grounding,kirillov2023segment,li2023semantic,chen2023semantic} have proposed solutions to address SAM's primary limitation of generating class-agnostic outputs. One notable approach is GroundingSAM~\cite{liu2023grounding,kirillov2023segment}, which synergizes an open-set object detector with SAM to yield labeled masks. In contrast, our work proposes leveraging SAM to enhance pseudo labels derived from a generated dataset.

\paragraph{Large Language Models.} LLMs have also impacted computer vision in many ways~\cite{Yang_2023_CVPR,brooks2023instructpix2pix,momeni2023verbs}. For instance, CuPL~\cite{pratt2023does} leverages the knowledge of GPT-3~\cite{brown2020language} to generate rich text descriptions which are prompted to CLIP and improve performance on the zero-shot image classification task. 
In \method, we use an LLM to increase diversity in the textual prompts conditioning the generation of images using a diffusion model, offering a notably more cost-effective approach than collecting and curating real data.

\section{Collaborating Foundation Models }
  \begin{figure*}[ht]
    \centering
    
    \includegraphics[width=0.95\linewidth]{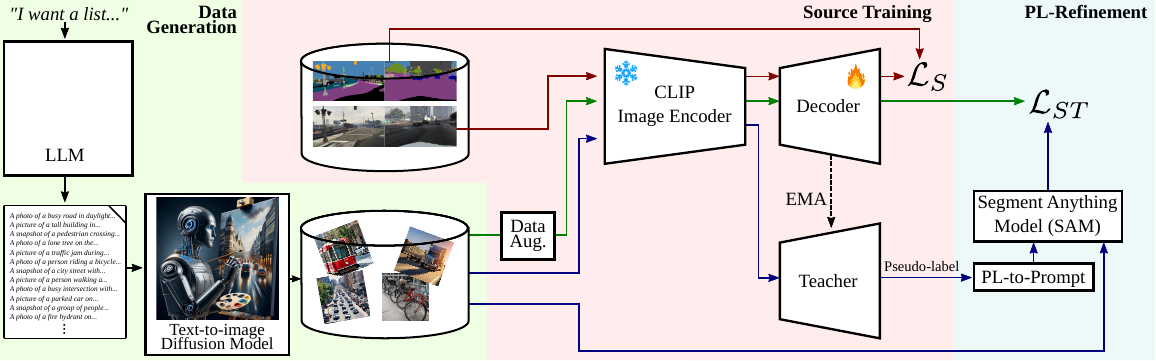}
      
      \caption{\textbf{Training pipeline of \method}: The model integrates a CLIP image encoder with a MaskFormer decoder (Sec.~\ref{sub:clip}). Our domain randomization strategy is based on a data generalization module (Sec.~\ref{sub:diffusion}) that combines a Large Language Model (LLM) with a text-to-image diffusion model to generate a varied dataset, representative of potential target datasets. This data is then employed in a Self-Training framework (Sec.~\ref{sub:self_training}), where initial pseudo labels (PL) prompt the Segment Anything Model (SAM) for refined pseudo labels, thereby fortifying the decoder's robustness.}
      \label{fig:method_diag}
    \end{figure*}

\label{sec:method}

The goal of Domain Generalized Semantic Segmentation (DGSS) is to train a segmentation model $g \circ f$, where $f$ corresponds to the feature extractor and $g$ the decoder, on a source domain $\source$ such that it can generalize to any unseen domain $\target$ that follows a different distribution. We assume having access only to a labeled source domain $\source {=} \{(\xmat_{i}^{\mathrm{S}}, \ymat_{i}^{\mathrm{S}})\}_{i=1}^{N^{\mathrm{S}}}$ of size $N^{\mathrm{S}}$ where $\xmat_{i}^{\mathrm{S}} \in \mathbb{R}^{H \cdot W \cdot 3}$ represents an RGB source image and $\ymat_{i}^{\mathrm{S}} \in \{0,1\}^{H \cdot W \cdot \mathcal{C}}$ the corresponding one-hot encoded ground-truth label.
The proposed approach shown in Fig.~\ref{fig:method_diag}, leverages the power of established foundation models to advance on DGSS.

\vspace{1mm}
\noindent\textbf{Overview:} \method uses a CNN-based CLIP backbone, harnessing its capability to extract robust features (Section~\ref{sub:clip}). To enable Domain Randomization (DR) with content diversity, we employ a text-to-image diffusion model, which generates photo-realistic images while being conditioned on textual prompts which are also generated using a Large Language Model to broaden the diversity of generated images used for self-training (Section~\ref{sub:diffusion}). Lastly, the predicted pseudo labels are refined with the help of the Segment-Anything Model (SAM) in a student-teacher fashion (Section~\ref{sub:self_training}).
This collaborative strategy aims to fortify the model's generalizability across diverse domains. 

\subsection{CLIP-based Student-Teacher Network}
\label{sub:clip}
The segmentation model is composed of an encoder $E$ and a decoder $D$. For the encoder $E$, we use a CNN-based CLIP instead of a ViT-based one, as it has been demonstrated in ~\cite{yu2023convolutions} that it produces better semantic features, and performs better on high-resolution images for semantic segmentation.
For the decoder $D$, we employ the one from Mask2Former~\cite{cheng2022masked}, a refined version of MaskFormer~\cite{cheng2021per}. This model redefines the task of semantic segmentation by treating it as mask classification. It achieves this by dissociating the division of the image into regions from their subsequent classification, thereby stepping aside from the conventional per-pixel classification in semantic segmentation 
The decoder comprises two components: a pixel decoder, and a transformer decoder, which perform mask prediction and classification. More details about the architecture can be found in~\cite{cheng2022masked}.

\paragraph{Training procedure:} We initially train our model on the labeled source dataset $\source$, while  keeping the backbone $f$ frozen as this offers: 
Firstly, it leads to faster and more efficient training. Secondly, it leverages pre-established, robust feature representations which ensure a better generalization while not overfitting on the source domain. Lastly, it preserves the image-text-aligned representations.

\noindent Thus, we only optimize the model's decoder and employ the loss function used in Mask2Former, which is composed of two parts: a mask loss $\mathcal{L}_{\text{mask}}$ and a classification loss $\mathcal{L}_{\text{cls}}$. The mask loss $\mathcal{L}_{\text{mask}}$ is defined as a linear combination of binary cross-entropy and dice losses. The total supervised loss for the source domain, denoted as $\mathcal{L}_{S}$, is formulated as:
\begin{equation}
\mathcal{L}_{S} = \mathcal{L}_{\text{mask}} + \lambda_{\text{cls}} \mathcal{L}_{\text{cls}} \quad ,
\end{equation}
where $\lambda_{\text{cls}}$ is a factor balancing the two loss components. 
\paragraph{Self-Training:}
The generated data are leveraged via a self-training strategy that trains the model using its own predictions, referred to as \textit{pseudo labels}. 
The pseudo labels are continuously updated during the training process, allowing them to evolve and become more accurate as the model's predictions improve.
Specifically, our approach employs a student-teacher framework to ensure training stability. The encoder remains frozen, which means our student-teacher framework only involves the decoder. The decoder $D$ (Section~\ref{sub:clip}) assumes the student's role. Concurrently, the teacher network $D_T$ is derived from the Exponential Moving Average (EMA) of the student's parameters, as  in~\cite{hoyer2023domain}. The update rule for the teacher's parameters $\theta_T$ is:
\begin{equation}
\theta'_{T} \leftarrow \alpha \theta_T + (1 - \alpha) \theta \quad ,
\end{equation}
where $\theta$ represents the student's parameters, and $\alpha$ is the factor controlling the teacher's momentum.
Following~\cite{hoyer2023domain}, data augmentation is applied to the student branch to enhance the robustness of the student decoder. Meanwhile, the teacher branch receives the original images, facilitating the generation of more reliable pseudo labels.
\subsection{Domain Randomization with Generative FMs}
\label{sub:diffusion}
We use text-conditioned DMs for generating images and leverage LLMs for creating the text-prompts.

\paragraph{Diffusion models:} To achieve domain randomization with high content diversity, we leverage a pretrained text-conditional diffusion model, as it excels in generating photorealistic images. Specifically, we employ the latent diffusion model~\cite{rombach2022high} Stable Diffusion trained on LAION-5B~\cite{schuhmann2022laion}. We provide the model with textual prompts describing scenes relevant to potential future target domains. In our experiments, we focus on prompts depicting urban street scenes, aligning with our benchmark scenarios. 
Note, this is in line with real-world scenarios, where we may not know the exact test environment, but we know the generic deployment setup (e.g. indoor, outdoor, or driving settings).
To obtain diverse prompts without effort, we rely on a LLM.

\paragraph{LLM for diversifying prompts:} 
Given their semantic diversity capabilities, we leverage LLMs (Llama-2~\cite{touvron2023llama}), to enrich prompts for image generation. 
Specifically, we employ an LLM to both create a wide range of synonyms for predefined class names and increase the semantic compositionality of the scene by varying descriptions of environmental factors such as lighting and weather conditions.
Our goal is to generate prompts formatted as `a photo of X in Z', where `X' represents any class name from our source dataset or its synonym, and `Z' encompasses any contextual information describing the environment. 
This leverages prior knowledge of LLMs, which possess an understanding of the contexts in which objects typically appear. Thus, we employ the LLM to enable the creation of synthetic target data, mirroring plausible future target environments.

To obtain such prompts, we prompt the LLM with the following text: \textit{
I want a list of prompts that can be used by an image generation model to generate synthetic images [...] The prompt should strictly follow this template: ``a photo of X in Z'' where X contains one or multiple class names within \texttt{\textbf{C}} [...]. Can you provide 100 diverse and simple prompts.} Where \texttt{\textbf{C}} is replaced by the list of all the class names of our source dataset.

\begin{figure}[!t]
    \centering
    
    \includegraphics[width=0.99\linewidth]{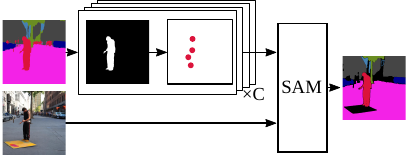}
    \caption{Pseudo Label refinement with SAM. We extract binary masks from the predicted segmentation. After labeling connected components and filtering noisy ones, we select random points within each binary mask. These points, along with the corresponding RGB images, are then used to prompt SAM, enabling the generation of more accurate segmentation maps}
    \label{fig:method_sam}
\end{figure}

\subsection{Pseudo Label refinement using FMs}
\label{sub:self_training}

The pseudo labels generated by the teacher decoder demonstrate prediction inaccuracies. For their refinement, we use SAM \cite{kirillov2023segment}, capable of generating multiple detailed masks for any image within a zero-shot framework. SAM, originally trained for promptable segmentation, is adapted to our task through the use of point-based prompting (Fig. ~\ref{fig:method_sam}).

Specifically, we adopt the following methodology:
Given a generated image \(\xvect\), we sequentially derive its pseudo labels using the CLIP encoder and the teacher decoder \(\pvect = D_T(E(\xvect))\). These pseudo labels are assembled into \(C\) class-specific binary masks \(\mvect_i, \; 1 \leq i \leq C\), each representing one of the predefined classes. For each class \(i\), the mask \(\mvect_i\) might encompass multiple objects, as they are not differentiated by instances. To resolve this, we apply the Hoshen–Kopelman algorithm, known for its efficiency in labeling connected components in masks \cite{Hoshen1976PercolationAC}. Consequently, for every class \(i\) segmented in the image, we obtain \(M_i\) distinct binary masks. Additionally, to further refine the pseudo labels, a filtering process is implemented to remove regions falling below a set size threshold.

Then, we randomly select points within each binary mask, utilizing these points to prompt SAM. For each set of points, SAM provides an enhanced binary mask. The final enhanced pseudo label map is obtained by aggregating all the predicted objects across all classes. SAM may predict overlapping masks for points from different classes. In such instances, the intersecting pixels are categorized as an ``unlabeled'' class to prevent the student model's training with erroneous pseudo labels.

The Student model is then trained using these refined pseudo labels. The self-training loss \(\mathcal{L}_{\text{st}}\) for generated images is the same as the source loss $\mathcal{L}_{S}$ of the source dataset.
\section{Experiments}
\label{sec:exps}

\subsection{Experimental Setups}

\noindent \textbf{Datasets:}
To evaluate the model's adaptability to new environments, we train on a single source domain and test on multiple unseen domains. We assess two scenarios:
(a) Synthetic-to-Real domain generalization, where we rely on GTA \cite{richter2016playing}, which comprises 24,966 images at a resolution of 1914×1052, and SYNTHIA \cite{ros2016synthia} which contains 9,400 images at a resolution of 1280×760. In the real-world domain, we use Cityscapes \cite{cordts2016cityscapes}, featuring 2,975 training images and 500 validation images at a resolution of 2048×1024. We also consider BDD \cite{yu2020bdd100k} which involves 1,000 validation images at a resolution of 1280×720, and Mapillary \cite{neuhold2017mapillary} including 2,000 validation images across diverse resolutions. For brevity, we denote `C' for Cityscapes, `B' for BDD100K, and `M' for Mapillary.
(b) Clear-to-adverse weather scenarios, where we incorporate the ACDC dataset \cite{sakaridis2021acdc} containing 406 validation images of 1920×1080 resolution.

\vspace{1mm}
\noindent \textbf{Implementation details:}
The network is trained for 40K iterations on GTA alone for initial reliable pseudo labels and an additional 40K iterations on both GTA and the generated dataset. On SYNTHIA, a smaller dataset, each step consists of 30K iterations. For Cityscapes, an even smaller dataset, we use 10K iterations for each step.
The learning rate is $lr=1\times10^{-4}$ for Cityscapes and $lr=1\times10^{-5}$ for GTA5 and SYNTHIA and the batch size is 8.
For other hyperparameters, we follow Mask2Former~\cite{cheng2022masked}, using the same loss functions with AdamW~\cite{kinga2015method} optimizer with a decay of $0.05$.
We use a random scaling in the range of [0.5,2.0] and random cropping of 768×768 size, and then we apply data augmentation (random flipping, color jittering).
Our assessment involves ResNet-50~\cite{he2016deep} and ResNet-101~\cite{he2016deep} encoders pretrained on both ImageNet~\cite{deng2009imagenet} and WIT~\cite{radford2021learning} from OpenAI.\footnote{https://github.com/openai/CLIP} We also use a ConvNext-Large encoder\cite{liu2022convnet, radford2021learning} pretrained on the LAION-2B~\cite{schuhmann2022laion} from OpenCLIP~\cite{ilharco2021openclip}. For the decoder, we use the one from Mask2Former \cite{cheng2022masked}.

\vspace{1mm}
\noindent \textbf{Evaluation protocol:}
We measure model performance using the mean Intersection over Union (\miou). For GTA, \miou is computed across 19 classes; for SYNTHIA, we report \miou for 16 shared classes with Cityscapes, BDD, and Mapillary.
We consistently evaluate the model's \miou using the final trained model across all experiments.

\subsection{State-of-the-art comparison}
\begin{table*}[ht]
    \centering
    \small
    \renewcommand{\arraystretch}{1.16}
   \small
    \begin{tabular}{>{\cellcolor{baselinecolor}}ll ccc|ccc|c}
    \toprule
    \cellcolor{white}{}   &\multirow{2}{*}{Method} & \multirow{2}{*}{Encoder} & \multirow{2}{*}{Pre-training} & \multirow{2}{*}{Training} & \multicolumn{3}{c|}{Target Domains (\miou in \%)} & \multirow{2}{*}{Avg.}\\
    \cellcolor{white}{}  &  & & & & Cityscapes & BDD100K & Mapillary  & \\
       \midrule
       & IBN-Net~\cite{pan2018two} & \multirow{6}{*}{ResNet-50} & \multirow{6}{*}{ImageNet}  &  \multirow{6}{*}{GTA5} & 33.9 & 32.3 & 37.8 & 34.6 \\
      &  ISW~\cite{choi2021robustnet} &   &  &   & 36.6 & 35.2 & 40.3 & 37.4 \\
     &   SHADE~\cite{zhao2022style} &  & &  & 44.7 & 39.3 & 43.3 & 42.4\\
      &  TLDR~\cite{kim2023texture} &  &  &  & 46.5 & 42.6 & 46.2 & 45.1 \\
      &  MoDify~\cite{jiang2023domain} & &  &   & 45.7 & 40.1 & 46.2 & 44.0 \\
      &  \textbf{\method} (Ours) &
       & WIT$^\dagger$ & & \textbf{54.6} & \textbf{46.7} & \textbf{58.6} & \textbf{53.3} \\
       \cline{2-9}
       
      &  IBN-Net~\cite{pan2018two} & \multirow{7}{*}{ResNet-101} & \multirow{7}{*}{ImageNet}  &  \multirow{7}{*}{GTA5}   &  37.4 & 34.2 & 36.8 & 36.1 \\
      &  ISW~\cite{choi2021robustnet} & &  &   & 37.2 & 33.4 & 35.6 & 35.4 \\
      &  SHADE~\cite{zhao2022style} &  &  & & 46.6 & 43.7 & 45.5 & 45.3 \\
      &  TLDR~\cite{kim2023texture} &  &  &  & 47.6 & 44.9 & 48.8 & 47.1 \\
      &  MoDify~\cite{jiang2023domain} & &  &  & 48.8 & 44.2 & 47.5 & 46.8 \\
      &  HRDA *~\cite{hoyer2023domain} & &  &  & 39.6 & 38.7 & 42.2 & 40.1 \\
      & \textbf{\method} (Ours) &  & 
       &  & 50.6 & 44.8 & 56.6 & 50.7 \\
      & \textbf{\method} (Ours) & & 
      WIT$^\dagger$ &  & \textbf{55.7} & \textbf{49.3} & \textbf{59.0} & \textbf{54.7} \\
      
      \cline{2-9}
      
      & HRDA~\cite{hoyer2023domain} & \multirow{2}{*}{MiT-B5} & \multirow{2}{*}{ImageNet} & \multirow{2}{*}{GTA5}  & 57.4 & 49.1 & 61.1 & 55.9 \\

       \multirow{-14}{*}{\rotatebox[origin=c]{90}{\textbf{Domain Generalization}}} & \textbf{\method} (Ours) &  & 
       &  & \textbf{58.1} & \textbf{53.8} & \textbf{62.3} & \textbf{58.1} \\
       \cline{2-9}
\addlinespace[1mm]
      
    &  PTDiffSeg~\cite{gong2023prompting} & Diffusion & LAION-5B & GTA5 &  52.0 & -- & -- & --  \\ 
     &  Grounding-SAM~\cite{kirillov2023segment, liu2023grounding} & ViT-H & SA-1B & -- & 43.5 & 39.4 & 48.4 & 43.8 \\  
    &   CAT-Seg~\cite{cho2023cat} & ViT-G/14 & \hspace{1mm}\multirow{5}{*}{LAION-2B} & COCO-Stuff~\cite{caesar2018coco}  & 45.0 & 47.6  & 51.8 & 48.2  \\  
    &  ODISE~\cite{xu2023open} & Diffusion &   & COCO Panoptic~\cite{lin2014microsoft} &  53.8 & 53.6 & 59.1 & 55.5 \\
      &  FC-CLIP~\cite{yu2023convolutions} & ConvNeXt-L &  & COCO Panoptic~\cite{lin2014microsoft} &  56.2 & 54.2 & 60.6 & 57.0  \\
      &  FC-CLIP * ~\cite{yu2023convolutions} & ConvNeXt-L &  & GTA5 &  53.6 & 47.6 & 57.4 & 52.9  \\
    \multirow{-7}{*}{\rotatebox[origin=c]{90}{{\textbf{Foundation Models}}}}  &  \textbf{\method} (Ours) & ConvNext-L & & GTA5 & \textbf{60.2} & \textbf{57.4} & \textbf{67.0} & \textbf{61.5} \\
       \bottomrule
    \end{tabular}
    \caption{Comparison with state of the art on  GTA $\rightarrow$ \{Cityscapes, BDD, Mapillary\} with leading DGSS methods and foundation models. $*$ denotes experiment obtained using the official code. $\dagger$ denotes Web Image-Text dataset used to train CLIP model from OpenAI. }
    \label{tab:merged_sota_no_category}
\end{table*}

We compare \method against both DGSS methods and methods based on FMs. We first describe the compared methods and then report and analyze the mIoU results in two settings: training on GTA and SYNTHIA, when the target domains are Cityscapes, BDD100K, and Mapillary. Similar to prior work in DGSS~\cite{zhao2022style,jiang2023domain,kim2023texture}, we assess the performance using ResNet-50, ResNet-101, and MiT-B5. We also extend our evaluation to include ConvNext-Large.

\paragraph{Compared Methods:} 
For \emph{traditional DGSS}, we include all leading DGSS methods: IBN-Net~\cite{pan2018two}, ISW~\cite{choi2021robustnet}, SHADE~\cite{zhao2022style}, TLDR~\cite{kim2023texture} and MoDify~\cite{jiang2023domain}. We also compare against HRDA~\cite{hoyer2023domain}, initially designed for Unsupervised Domain Adaptation and adjusted for DGSS in \cite{hoyer2022hrda}.

For \emph{Foundation models}, we compare against three categories: (i) SAM-based \cite{kirillov2023segment} (GroundingSAM\ \cite{kirillov2023segment,liu2023grounding}), (ii) CLIP-based\cite{radford2021learning} (FC-CLIP\cite{yu2023convolutions}, CATSeg \cite{cho2023cat} ), and (iii) Diffusion-based\cite{rombach2022high}  (ODISE\cite{xu2023open}, PTDiffSeg \cite{gong2023prompting}) methods. 
(i) For SAM-based methods, GroundingSAM merges SAM's segmentation power with the GroundingDINO open-set object detector that uses text prompts to predict bounding boxes. SAM encodes these boxes to predict their corresponding semantic mask. 
(ii) For CLIP-based methods, FC-CLIP uses a frozen CNN-based backbone to extract semantic features for both mask generation and CLIP classification. CAT-Seg uses cost aggregation to finetune CLIP image embeddings for better generalization and robustness to unseen domains.
(iii) For Diffusion-based methods, PTDiffSeg~\cite{gong2023prompting} is tailored for DGSS and uses the feature representations of a diffusion model for semantic segmentation. We present results only on Cityscapes, as the paper does not report results on other datasets and the code is not available. 
While ODISE $(55.5\%)$ falls into this category, we note that during inference, to improve its performance it uses CLIP.

\paragraph{GTA as source domain:} Table \ref{tab:merged_sota_no_category} reports the results when benchmarking {GTA $\rightarrow$ \{C, B, M\}}. 
Overall, we observe that \method outperforms both all DGSS methods and FM-based methods for all types of encoders and pre-training setups. 
For traditional DGSS methods, when using ResNet-50, \method achieves $53.3\%$, i.e. $+8.1\%$ compared to the second competitor MoDify, whereas with ResNet-101, it reaches $54.7\%$, improving the benchmark by $+7.6\%$. 
When using ConvNext-L, \method achieves $61.5\%$, outperforming the previously strongest DGSS model HRDA ($55.9\%)$ by +5.5\%. Additionally, \method notably outperforms HRDA by 10\% when using ResNet-101 as backbone and by 3\% when using MiT-B5~\cite{xie2021segformer}. 
These results show that our method efficiently leverages several FMs, leading to improved features and enhanced training.

\noindent Furthermore, \method consistently outperforms all methods based on FMs, on all three datasets, achieving an average \miou of $61.5\%$. 
GroundingSAM alongside open-vocabulary methods (ODISE, FC-CLIP, and CAT-Seg) face a significant limitation due to their reliance on text prompts, which leads to ambiguity when it comes to segmenting classes that can hardly be described with words only (e.g., classes "pole" and "terrain"). Text prompts alone fall short in fully capturing all visual elements, particularly in complex scenes like autonomous driving, as the same class may cover diverse objects with different characteristics. 
Instead, \method leverages image data by training on the source domain and harnessing visual cues that are essential for accurately segmenting these complex classes.
\begin{table}[!h]
\centering
\small
\resizebox{\linewidth}{!}{
\begin{tabular}{lccccc}
\toprule
Method & Encoder & C & B & M & Avg \\
\midrule
DRPC \cite{yue2019domain} & \multirow{5}{*}{ResNet-50} & 35.7 & 31.5 & 32.7 & 33.3\\
SAN-SAW \cite{peng2022semantic} &  & 38.9 & 35.2 & 34.5 & 36.2 \\
MoDify~\cite{jiang2023domain} & & 38.9 & 33.7 & 36.2 & 36.3 \\
TLDR~\cite{kim2023texture} &  & 41.9 & 34.4 & 36.8 & 37.7 \\
\textbf{\method} (Ours) &  & \textbf{46.1} & \textbf{37.6} & \textbf{48.1} & \textbf{43.9} \\
\midrule
DRPC \cite{yue2019domain} & \multirow{8}{*}{ResNet-101} & 37.6 & 34.4 & 34.1 & 35.3 \\
GTR~\cite{peng2021global} & & 39.7 & 35.3 & 36.4 & 37.1 \\
 FSDR~\cite{huang2021fsdr} & & 40.8 & 37.4 & 39.6 & 39.3 \\
SAN-SAW \cite{peng2022semantic} &  & 40.9 & 36.0 & 37.3 & 38.0 \\
TLDR~\cite{kim2023texture} &  & 42.6 & 35.5 & 37.5 & 38.5\\
HRDA *~\cite{hoyer2023domain} &  & 34.9 & 25.0 & 34.0 & 31.3\\
MoDify~\cite{jiang2023domain} & & 43.4 & 39.5 & 42.3 & 41.7 \\
\textbf{\method} (Ours) &  & \textbf{49.1} & \textbf{40.3} & \textbf{50.1} & \textbf{46.5} \\
\midrule
HRDA *~\cite{hoyer2023domain} & \multirow{2}{*}{MiT-B5}  & 39.6 & 32.6 & 40.0 & 37.4\\
\textbf{\method} (Ours) &  & \textbf{42.2} & \textbf{38.3} & \textbf{43.6} & \textbf{41.4}\\
\midrule
FC-CLIP *~\cite{yu2023convolutions} & \multirow{2}{*}{ConvNext-L} & 38.0 & 29.9 & 39.0 & 35.6\\
\textbf{\method} (Ours) &  & \textbf{53.4} & \textbf{47.0} & \textbf{55.8} & \textbf{52.1} \\
\bottomrule
\end{tabular}}
\caption{Comparison with state-of-the-art methods for DGSS on Synthia $\rightarrow$ \{Cityscapes (C), BDD (B), Mapillary (M)\}. * denotes experiment obtained using the official code}
\label{tab:syn}
\end{table}

\noindent \textbf{SYNTHIA as source domain:}
In Table \ref{tab:syn}, we address the setting {SYNTHIA $\rightarrow$ \{C, B, M\}}.  \method achieves $43.9\%$ using ResNet-50 and outperforms the previously leading method (TLDR) by $+6.2\%$. When using ResNet-101, \method achieves $46.5\%$ and improves the previously best method (MoDify) by $+4.8\%$. Moreover, \method outperforms HRDA with MiT-B5 backbone by $+4\%$. When using ConvNext-L we obtain the best results, i.e. $52.1\%$ \miou, outperforming FC-CLIP by a large margin. 
\begin{table}[t]
\centering
\begin{tabular}{lcccc}
\toprule
Method & Encoder & Night & Snow & Rain \\
\midrule
CLIPStyler~\cite{kwon2022clipstyler} & \multirow{3}{*}{ResNet-50} & 21.3 & 41.0 & 38.7  \\
PODA~\cite{fahes2023poda} & & 25.0 & 43.9 & 42.3  \\
\textbf{\method} (Ours)  && $\textbf{29.4}$ & $\textbf{52.1}$ & $\textbf{49.7}$ \\
\midrule
\multirow{2}{*}{\textbf{\method} (Ours)}  & ResNet-101 & $\textbf{33.0}$ & $\textbf{51.3}$ & $\textbf{53.4}$ \\
  & ConvNext-L& $\textbf{45.1}$ & $\textbf{65.3}$ & $\textbf{64.4}$ \\
\bottomrule
\end{tabular}%
\caption{Comparison with zero-shot Domain Adaptation methods. evaluation on ACDC dataset.}
\label{tab:zeroshot}
\end{table}

\paragraph{Comparison with zero-shot Domain Adaptation:}
Zero-shot Domain Adaptation assumes having access to the annotated source domain and a domain description of the target using natural language, in the form of a text prompt.
The task of prompt-driven zero-shot domain adaptation was introduced by P{\O}DA\cite{fahes2023poda}, a method that harnesses CLIP's robust feature representations. It adjusts source feature statistics to align with the target domain, guided by natural language domain prompts.
To ensure a fair comparison with P{\O}DA~\cite{fahes2023poda}, we employed the same ResNet-50 backbone pretrained on CLIP. The results in Table \ref{tab:zeroshot} demonstrate that our method consistently outperforms P{\O}DA, achieving improvements of $+4.4\%$, $+8.2\%$ and $+7.4\%$ across three scenarios: {day $\rightarrow$ night}, {clear $\rightarrow$ snow} and {clear $\rightarrow$ rain}, respectively.
We also provide results using ResNet-101 pretrained on CLIP and ConvNext-L on LAION-2B.

\subsection{Ablation studies}
We ablate the various components of \method.
We focus our evaluation on the setting GTA $\rightarrow$ \{C, B, M\}.

\begin{table}[!h]
\centering
\small
\resizebox{\linewidth}{!}{
\begin{tabular}{lcccc}
\toprule
Backbone & CLIP & \{LLM, Diffusion\} & SAM & Avg \\
\midrule
    \multirow{3}{*}{ResNet-50} & \checkmark &  & & 50.0  \\
    &   \checkmark  & \checkmark &  & 50.7  \\
    & \checkmark & \checkmark  & \checkmark & \textbf{53.3 } \\
\midrule
    \multirow{3}{*}{ResNet-101} & \checkmark &  &   & 51.9  \\
    &   \checkmark  & \checkmark     &   &   53.3  \\
    & \checkmark & \checkmark  & \checkmark &   \textbf{54.7}  \\
\midrule
    \multirow{3}{*}{ConvNext-L} & \checkmark &  &  & 58.5  \\
    & \checkmark & \checkmark  &  & 58.6  \\
    & \checkmark & \checkmark  & \checkmark & \textbf{61.5}  \\
\bottomrule
\end{tabular}}
\caption{Ablation of the key components of \method on GTA $\rightarrow$ \{Cityscapes, BDD, Mapillary\} using three different backbones.}
\label{tab:mainablation}
\end{table}

\noindent \textbf{Effect of each Foundation model:}
We investigate the impact of integrating various FMs on the performance of DGSS. Table \ref{tab:mainablation} reports the results, i.e., average \miou across three diverse datasets: C, B, and M. Notably, employing CLIP as a standalone feature extractor with various backbones, including ResNet-50, ResNet-101 and ConvNext-L, yields state-of-the-art results, reaching an average \miou of $50.0\%$, $51.9\%$ and $58.5\%$, respectively. This impressive performance underscores CLIP's rich feature representation, hence leading to robustness in handling unseen environments. 
Including the diffusion model conditioned by the LLM-generated prompt brings marginal improvements with ConvNext-L and ResNet-50 backbones. 
This can be attributed to the noise in the pseudo labels predicted by the Teacher model, which comes from the domain shift in the generated images produced by the text-to-image model, thus posing difficulties in self-training.
Adding SAM leads to the largest performance gains, i.e. approx.\ +3\% in mIoU for all models. This affirms the critical role of SAM in refining pseudo labels and enhancing their accuracy before using them in supervision. This gain is evident in all backbones, highlighting SAM's effectiveness in addressing the challenges of self-training.

\begin{figure}[t]
  \centering
  \includegraphics[width=\linewidth]{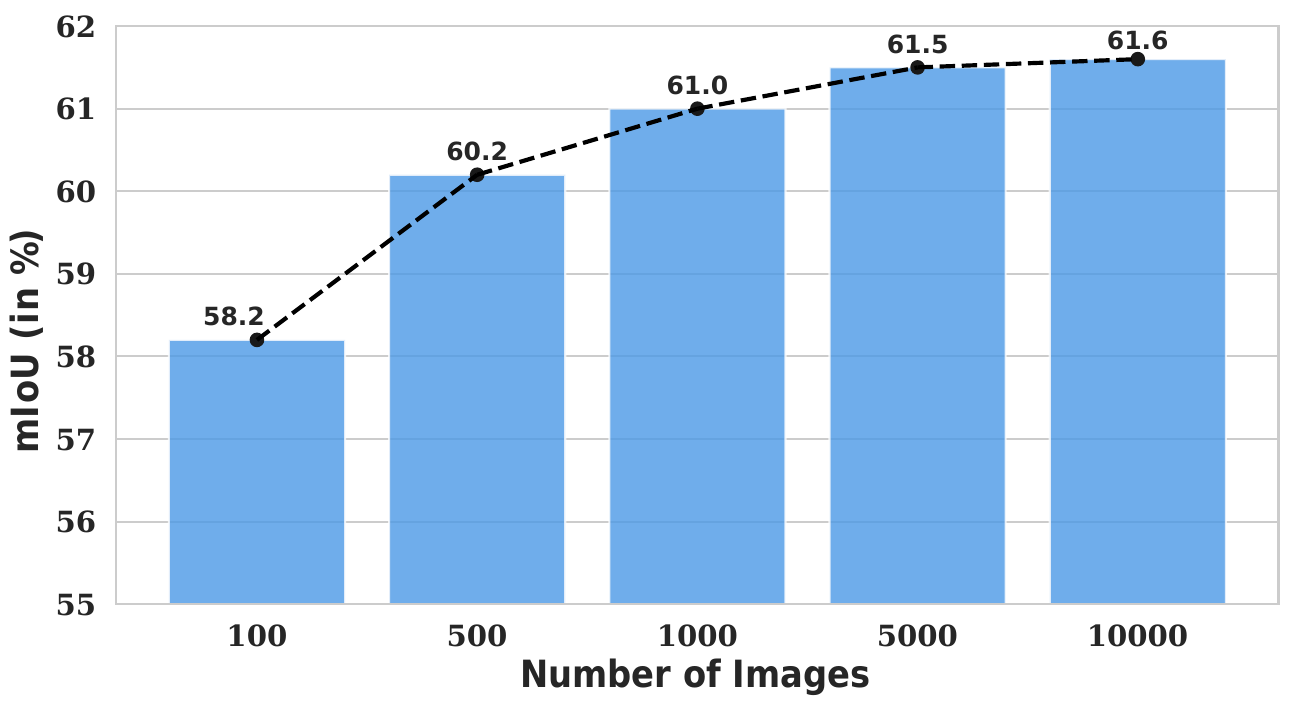} 
  \caption{Effect of generated dataset size on \miou. Experiments performed on GTA $\rightarrow$ \{Cityscapes, BDD, Mapillary\}.}
  \label{fig:miou_plot}
\end{figure}

\paragraph{Effect of Scaling:}
Here, we examine the impact of the generated dataset size on the performance. Figure~\ref{fig:miou_plot} shows that increasing the size of the generated dataset improves the \miou significantly (up to +3\%) up approximately 5K images, where a plateau is reached. Therefore, in our experiments, we use a dataset size of 5000 images.

\begin{figure}[ht]
  \centering
  \includegraphics[width=\linewidth]{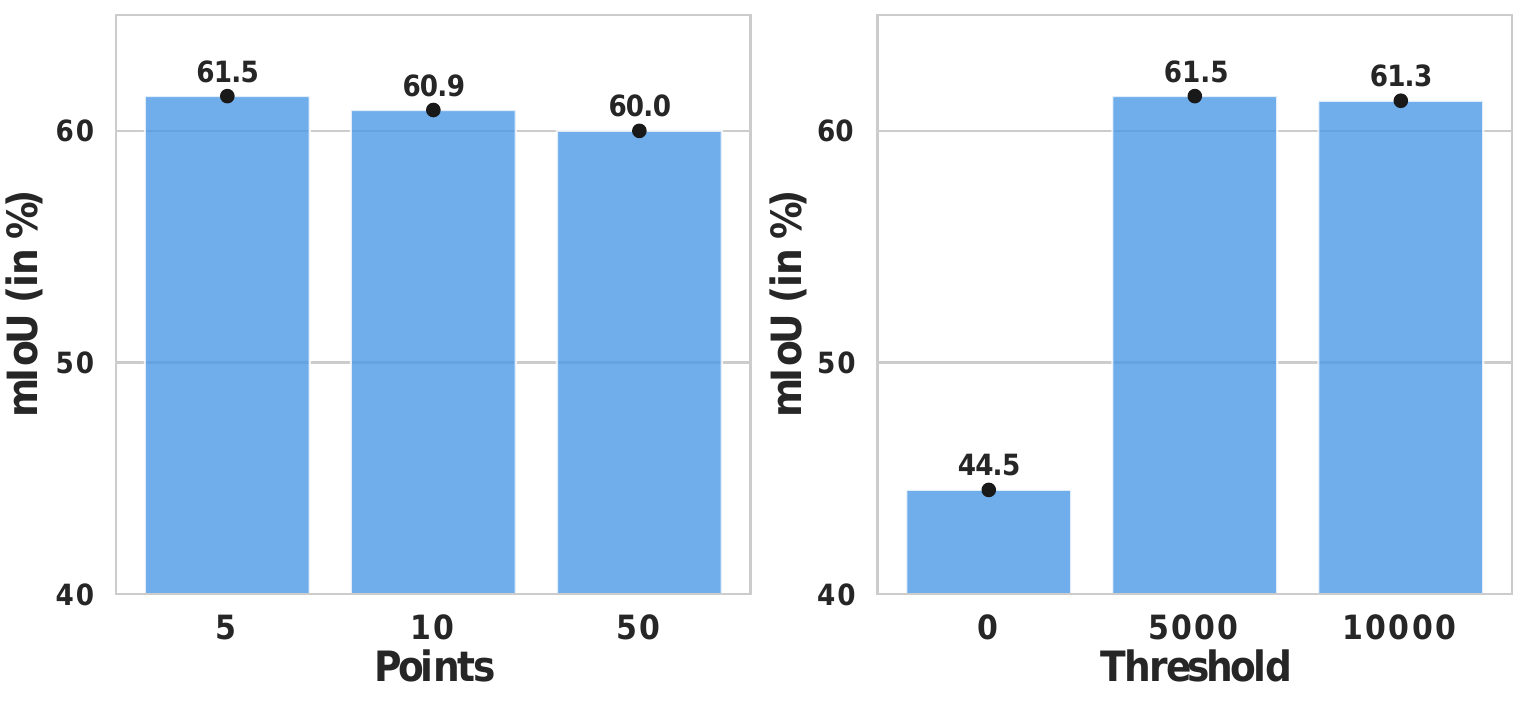} 
  \caption{Ablation Study on SAM Prompting: This study examines the effects of varying the number of sampled points and the threshold used to separate connected components.}
  \label{fig:miou_plot_sam}
\end{figure}

\noindent \textbf{Prompting SAM:}
We examine how various prompts used in SAM affect the performance, Fig. ~\ref{fig:miou_plot_sam} reports the results. 
The left barplot displays the impact of varying the number of points used to prompt SAM for predicting masks. Increasing them leads to a decrease in \miou. This is most likely due to the noise in pseudo labels and the possibility of points being placed on different objects on the image, resulting in less accurate masks, which leads to noisy pseudo labels.
The right barplot illustrates the relationship between the threshold size used in the Hoshen–Kopelman algorithm (which separates the connected components of the binary mask) and the performance of our model. A higher threshold helps eliminate small regions that might be incorrectly identified as objects belonging to a certain class, thereby enhancing the precision of the pseudo labels. Therefore, the model's performance improves.

\begin{table}[!h]
\centering
\resizebox{\linewidth}{!}{
\begin{tabular}{lcccc}
\toprule
Backbone & Cityscapes & BDD100K & Mapillary & Avg. \\
\midrule
Trainable & 58.6 & 53.0 & 62.8 & 58.1  \\
Frozen & \textbf{60.2} & \textbf{57.4} & \textbf{67.0} & \textbf{61.5}  \\
\bottomrule
\end{tabular}}
\caption{Impact of fine-tuning CLIP on the performance.}
\label{tab:ablation_clip}
\end{table}

\noindent \textbf{Fine-tuning the CLIP encoder:}
Here, we examine the impact of fine-tuning CLIP on the performance. 
Table~\ref{tab:ablation_clip} reports the results. We observe that keeping the backbone frozen during training, our method effectively retains the encoder’s ability to generalize and perform well in unseen environments. 
Contrarily, finetuning the CLIP encoder resulted in poorer performance, likely due to overfitting on the source domain, which consequently led to diminished generalization capabilities on the target domains.

\section{Conclusion}
In this work, we proposed \method, a novel approach to Domain Generalized Semantic Segmentation that uniquely integrates various Foundation Models in a collaborative manner. By leveraging the robust features of CLIP to unseen domains and the content diversification capabilities of diffusion models, enriched by LLM-generated text prompts, and employing the ability of SAM to enhance pseudo labels for self-training, we effectively address key challenges in DGSS. Our approach introduces a self-training strategy using generated data. Through extensive experimentation, \method demonstrates superior performance over traditional DGSS and open-vocabulary segmentation models on various benchmarks, marking a significant advancement in the field. This work not only bridges a crucial research gap but also establishes a new standard for robustness and adaptability in semantic segmentation across diverse domains in the era of Foundation Models.

\paragraph{Acknowledgements.}
{This work has been supported by the French National Research Agency (ANR) with the ANR-20-CE23-0027 project and was granted access to the HPC resources of IDRIS under the allocation AD011013071 made by GENCI. We would like to thank T.Delatolas and I.Marouf for proofreading.
}

{
    \small
    \bibliographystyle{ieeenat_fullname}
    \bibliography{main}

\begin{thebibliography}{86}
\providecommand{\natexlab}[1]{#1}
\providecommand{\url}[1]{\texttt{#1}}
\expandafter\ifx\csname urlstyle\endcsname\relax
  \providecommand{\doi}[1]{doi: #1}\else
  \providecommand{\doi}{doi: \begingroup \urlstyle{rm}\Url}\fi

\bibitem[Benigmim et~al.(2023)Benigmim, Roy, Essid, Kalogeiton, and Lathuili\`ere]{Benigmim_2023_CVPR}
Yasser Benigmim, Subhankar Roy, Slim Essid, Vicky Kalogeiton, and St\'ephane Lathuili\`ere.
\newblock One-shot unsupervised domain adaptation with personalized diffusion models.
\newblock In \emph{Proceedings of the IEEE/CVF Conference on Computer Vision and Pattern Recognition (CVPR) Workshops}, pages 698--708, 2023.

\bibitem[Bommasani et~al.(2021)Bommasani, Hudson, Adeli, Altman, Arora, von Arx, Bernstein, Bohg, Bosselut, Brunskill, et~al.]{bommasani2021opportunities}
Rishi Bommasani, Drew~A Hudson, Ehsan Adeli, Russ Altman, Simran Arora, Sydney von Arx, Michael~S Bernstein, Jeannette Bohg, Antoine Bosselut, Emma Brunskill, et~al.
\newblock On the opportunities and risks of foundation models.
\newblock \emph{arXiv preprint arXiv:2108.07258}, 2021.

\bibitem[Brooks et~al.(2023)Brooks, Holynski, and Efros]{brooks2023instructpix2pix}
Tim Brooks, Aleksander Holynski, and Alexei~A Efros.
\newblock Instructpix2pix: Learning to follow image editing instructions.
\newblock In \emph{Proceedings of the IEEE/CVF Conference on Computer Vision and Pattern Recognition}, pages 18392--18402, 2023.

\bibitem[Brown et~al.(2020)Brown, Mann, Ryder, Subbiah, Kaplan, Dhariwal, Neelakantan, Shyam, Sastry, Askell, et~al.]{brown2020language}
Tom Brown, Benjamin Mann, Nick Ryder, Melanie Subbiah, Jared~D Kaplan, Prafulla Dhariwal, Arvind Neelakantan, Pranav Shyam, Girish Sastry, Amanda Askell, et~al.
\newblock Language models are few-shot learners.
\newblock \emph{Advances in neural information processing systems}, 33:\penalty0 1877--1901, 2020.

\bibitem[Caesar et~al.(2018)Caesar, Uijlings, and Ferrari]{caesar2018coco}
Holger Caesar, Jasper Uijlings, and Vittorio Ferrari.
\newblock Coco-stuff: Thing and stuff classes in context.
\newblock In \emph{Proceedings of the IEEE conference on computer vision and pattern recognition}, pages 1209--1218, 2018.

\bibitem[Chen et~al.(2023)Chen, Yang, and Zhang]{chen2023semantic}
Jiaqi Chen, Zeyu Yang, and Li Zhang.
\newblock Semantic segment anything.
\newblock \url{https://github.com/fudan-zvg/Semantic-Segment-Anything}, 2023.

\bibitem[Chen et~al.(2017)Chen, Papandreou, Kokkinos, Murphy, and Yuille]{chen2017deeplab}
Liang-Chieh Chen, George Papandreou, Iasonas Kokkinos, Kevin Murphy, and Alan~L Yuille.
\newblock Deeplab: Semantic image segmentation with deep convolutional nets, atrous convolution, and fully connected crfs.
\newblock \emph{IEEE transactions on pattern analysis and machine intelligence}, 40\penalty0 (4):\penalty0 834--848, 2017.

\bibitem[Cheng et~al.(2021)Cheng, Schwing, and Kirillov]{cheng2021per}
Bowen Cheng, Alex Schwing, and Alexander Kirillov.
\newblock Per-pixel classification is not all you need for semantic segmentation.
\newblock \emph{Advances in Neural Information Processing Systems}, 34:\penalty0 17864--17875, 2021.

\bibitem[Cheng et~al.(2022)Cheng, Misra, Schwing, Kirillov, and Girdhar]{cheng2022masked}
Bowen Cheng, Ishan Misra, Alexander~G Schwing, Alexander Kirillov, and Rohit Girdhar.
\newblock Masked-attention mask transformer for universal image segmentation.
\newblock In \emph{Proceedings of the IEEE/CVF conference on computer vision and pattern recognition}, pages 1290--1299, 2022.

\bibitem[Cho et~al.(2023{\natexlab{a}})Cho, Nam, Kim, Yang, and Kwak]{cho2023promptstyler}
Junhyeong Cho, Gilhyun Nam, Sungyeon Kim, Hunmin Yang, and Suha Kwak.
\newblock Promptstyler: Prompt-driven style generation for source-free domain generalization.
\newblock In \emph{Proceedings of the IEEE/CVF International Conference on Computer Vision}, pages 15702--15712, 2023{\natexlab{a}}.

\bibitem[Cho et~al.(2023{\natexlab{b}})Cho, Shin, Hong, An, Lee, Arnab, Seo, and Kim]{cho2023cat}
Seokju Cho, Heeseong Shin, Sunghwan Hong, Seungjun An, Seungjun Lee, Anurag Arnab, Paul~Hongsuck Seo, and Seungryong Kim.
\newblock Cat-seg: Cost aggregation for open-vocabulary semantic segmentation.
\newblock \emph{arXiv preprint arXiv:2303.11797}, 2023{\natexlab{b}}.

\bibitem[Choi et~al.(2021)Choi, Jung, Yun, Kim, Kim, and Choo]{choi2021robustnet}
Sungha Choi, Sanghun Jung, Huiwon Yun, Joanne~T Kim, Seungryong Kim, and Jaegul Choo.
\newblock Robustnet: Improving domain generalization in urban-scene segmentation via instance selective whitening.
\newblock In \emph{CVPR}, pages 11580--11590, 2021.

\bibitem[Cordts et~al.(2016)Cordts, Omran, Ramos, Rehfeld, Enzweiler, Benenson, Franke, Roth, and Schiele]{cordts2016cityscapes}
Marius Cordts, Mohamed Omran, Sebastian Ramos, Timo Rehfeld, Markus Enzweiler, Rodrigo Benenson, Uwe Franke, Stefan Roth, and Bernt Schiele.
\newblock The cityscapes dataset for semantic urban scene understanding.
\newblock In \emph{CVPR}, pages 3213--3223, 2016.

\bibitem[Creswell et~al.(2018)Creswell, White, Dumoulin, Arulkumaran, Sengupta, and Bharath]{creswell2018generative}
Antonia Creswell, Tom White, Vincent Dumoulin, Kai Arulkumaran, Biswa Sengupta, and Anil~A Bharath.
\newblock Generative adversarial networks: An overview.
\newblock \emph{IEEE signal processing magazine}, 35\penalty0 (1):\penalty0 53--65, 2018.

\bibitem[Deng et~al.(2009)Deng, Dong, Socher, Li, Li, and Fei-Fei]{deng2009imagenet}
Jia Deng, Wei Dong, Richard Socher, Li-Jia Li, Kai Li, and Li Fei-Fei.
\newblock Imagenet: A large-scale hierarchical image database.
\newblock In \emph{2009 IEEE conference on computer vision and pattern recognition}, pages 248--255. Ieee, 2009.

\bibitem[Dong et~al.(2023)Dong, Bao, Zheng, Zhang, Chen, Yang, Zeng, Zhang, Yuan, Chen, et~al.]{dong2023maskclip}
Xiaoyi Dong, Jianmin Bao, Yinglin Zheng, Ting Zhang, Dongdong Chen, Hao Yang, Ming Zeng, Weiming Zhang, Lu Yuan, Dong Chen, et~al.
\newblock Maskclip: Masked self-distillation advances contrastive language-image pretraining.
\newblock In \emph{Proceedings of the IEEE/CVF Conference on Computer Vision and Pattern Recognition}, pages 10995--11005, 2023.

\bibitem[Dosovitskiy et~al.(2020)Dosovitskiy, Beyer, Kolesnikov, Weissenborn, Zhai, Unterthiner, Dehghani, Minderer, Heigold, Gelly, et~al.]{dosovitskiy2020image}
Alexey Dosovitskiy, Lucas Beyer, Alexander Kolesnikov, Dirk Weissenborn, Xiaohua Zhai, Thomas Unterthiner, Mostafa Dehghani, Matthias Minderer, Georg Heigold, Sylvain Gelly, et~al.
\newblock An image is worth 16x16 words: Transformers for image recognition at scale.
\newblock \emph{arXiv preprint arXiv:2010.11929}, 2020.

\bibitem[Fahes et~al.(2023)Fahes, Vu, Bursuc, P{\'e}rez, and de~Charette]{fahes2023poda}
Mohammad Fahes, Tuan-Hung Vu, Andrei Bursuc, Patrick P{\'e}rez, and Raoul de Charette.
\newblock Poda: Prompt-driven zero-shot domain adaptation.
\newblock In \emph{Proceedings of the IEEE/CVF International Conference on Computer Vision}, pages 18623--18633, 2023.

\bibitem[Giannakis et~al.(2023)Giannakis, Bhardwaj, Sam, and Leontidis]{giannakis2023deep}
Iraklis Giannakis, Anshuman Bhardwaj, Lydia Sam, and Georgios Leontidis.
\newblock Deep learning universal crater detection using segment anything model (sam).
\newblock \emph{arXiv preprint arXiv:2304.07764}, 2023.

\bibitem[Gong et~al.(2023)Gong, Danelljan, Sun, Mangas, and Van~Gool]{gong2023prompting}
Rui Gong, Martin Danelljan, Han Sun, Julio~Delgado Mangas, and Luc Van~Gool.
\newblock Prompting diffusion representations for cross-domain semantic segmentation.
\newblock \emph{arXiv preprint arXiv:2307.02138}, 2023.

\bibitem[He et~al.(2016)He, Zhang, Ren, and Sun]{he2016deep}
Kaiming He, Xiangyu Zhang, Shaoqing Ren, and Jian Sun.
\newblock Deep residual learning for image recognition.
\newblock In \emph{Proceedings of the IEEE conference on computer vision and pattern recognition}, pages 770--778, 2016.

\bibitem[He et~al.(2023)He, Bao, Li, Grant, and Ou]{he2023accuracy}
Sheng He, Rina Bao, Jingpeng Li, P~Ellen Grant, and Yangming Ou.
\newblock Accuracy of segment-anything model (sam) in medical image segmentation tasks.
\newblock \emph{arXiv preprint arXiv:2304.09324}, 2023.

\bibitem[Hoshen and Kopelman(1976)]{Hoshen1976PercolationAC}
Joseph Hoshen and Raoul Kopelman.
\newblock Percolation and cluster distribution. i. cluster multiple labeling technique and critical concentration algorithm.
\newblock \emph{Physical Review B}, 14:\penalty0 3438--3445, 1976.

\bibitem[Hoyer et~al.(2022{\natexlab{a}})Hoyer, Dai, and Van~Gool]{hoyer2022daformer}
Lukas Hoyer, Dengxin Dai, and Luc Van~Gool.
\newblock Daformer: Improving network architectures and training strategies for domain-adaptive semantic segmentation.
\newblock In \emph{Proceedings of the IEEE/CVF Conference on Computer Vision and Pattern Recognition}, pages 9924--9935, 2022{\natexlab{a}}.

\bibitem[Hoyer et~al.(2022{\natexlab{b}})Hoyer, Dai, and Van~Gool]{hoyer2022hrda}
Lukas Hoyer, Dengxin Dai, and Luc Van~Gool.
\newblock Hrda: Context-aware high-resolution domain-adaptive semantic segmentation.
\newblock In \emph{Computer Vision--ECCV 2022: 17th European Conference, Tel Aviv, Israel, October 23--27, 2022, Proceedings, Part XXX}, pages 372--391. Springer, 2022{\natexlab{b}}.

\bibitem[Hoyer et~al.(2023)Hoyer, Dai, and Van~Gool]{hoyer2023domain}
Lukas Hoyer, Dengxin Dai, and Luc Van~Gool.
\newblock Domain adaptive and generalizable network architectures and training strategies for semantic image segmentation.
\newblock \emph{T-PAMI}, 2023.

\bibitem[Huang et~al.(2021)Huang, Guan, Xiao, and Lu]{huang2021fsdr}
Jiaxing Huang, Dayan Guan, Aoran Xiao, and Shijian Lu.
\newblock Fsdr: Frequency space domain randomization for domain generalization.
\newblock In \emph{CVPR}, pages 6891--6902, 2021.

\bibitem[Huang et~al.(2019)Huang, Zhou, Zhu, Liu, and Shao]{huang2019iterative}
Lei Huang, Yi Zhou, Fan Zhu, Li Liu, and Ling Shao.
\newblock Iterative normalization: Beyond standardization towards efficient whitening.
\newblock In \emph{CVPR}, pages 4874--4883, 2019.

\bibitem[Ilharco et~al.(2021)Ilharco, Wortsman, Wightman, Gordon, Carlini, Taori, Dave, Shankar, Namkoong, Miller, et~al.]{ilharco2021openclip}
Gabriel Ilharco, Mitchell Wortsman, Ross Wightman, Cade Gordon, Nicholas Carlini, Rohan Taori, Achal Dave, Vaishaal Shankar, Hongseok Namkoong, John Miller, et~al.
\newblock Openclip.
\newblock \emph{Zenodo}, 4:\penalty0 5, 2021.

\bibitem[Jia et~al.(2021)Jia, Yang, Xia, Chen, Parekh, Pham, Le, Sung, Li, and Duerig]{jia2021scaling}
Chao Jia, Yinfei Yang, Ye Xia, Yi-Ting Chen, Zarana Parekh, Hieu Pham, Quoc Le, Yun-Hsuan Sung, Zhen Li, and Tom Duerig.
\newblock Scaling up visual and vision-language representation learning with noisy text supervision.
\newblock In \emph{International conference on machine learning}, pages 4904--4916. PMLR, 2021.

\bibitem[Jiang et~al.(2023)Jiang, Huang, Jin, and Lu]{jiang2023domain}
Xueying Jiang, Jiaxing Huang, Sheng Jin, and Shijian Lu.
\newblock Domain generalization via balancing training difficulty and model capability.
\newblock In \emph{Proceedings of the IEEE/CVF International Conference on Computer Vision}, pages 18993--19003, 2023.

\bibitem[Karazija et~al.(2023)Karazija, Laina, Vedaldi, and Rupprecht]{karazija2023diffusion}
Laurynas Karazija, Iro Laina, Andrea Vedaldi, and Christian Rupprecht.
\newblock Diffusion models for zero-shot open-vocabulary segmentation.
\newblock \emph{arXiv preprint arXiv:2306.09316}, 2023.

\bibitem[Kim et~al.(2023)Kim, Kim, and Kim]{kim2023texture}
Sunghwan Kim, Dae-hwan Kim, and Hoseong Kim.
\newblock Texture learning domain randomization for domain generalized segmentation.
\newblock \emph{Proceedings of the IEEE/CVF International Conference on Computer Vision}, 2023.

\bibitem[Kingma and Ba(2015)]{kinga2015method}
Diederik~P Kingma and Jimmy Ba.
\newblock A method for stochastic optimization.
\newblock In \emph{International conference on learning representations (ICLR)}, page~6, 2015.

\bibitem[Kirillov et~al.(2023)Kirillov, Mintun, Ravi, Mao, Rolland, Gustafson, Xiao, Whitehead, Berg, Lo, et~al.]{kirillov2023segment}
Alexander Kirillov, Eric Mintun, Nikhila Ravi, Hanzi Mao, Chloe Rolland, Laura Gustafson, Tete Xiao, Spencer Whitehead, Alexander~C Berg, Wan-Yen Lo, et~al.
\newblock Segment anything.
\newblock \emph{arXiv preprint arXiv:2304.02643}, 2023.

\bibitem[Kwon and Ye(2022)]{kwon2022clipstyler}
Gihyun Kwon and Jong~Chul Ye.
\newblock Clipstyler: Image style transfer with a single text condition.
\newblock In \emph{Proceedings of the IEEE/CVF Conference on Computer Vision and Pattern Recognition}, pages 18062--18071, 2022.

\bibitem[LeCun et~al.(1998)LeCun, Bottou, Bengio, and Haffner]{lecun1998gradient}
Yann LeCun, L{\'e}on Bottou, Yoshua Bengio, and Patrick Haffner.
\newblock Gradient-based learning applied to document recognition.
\newblock \emph{Proceedings of the IEEE}, 86\penalty0 (11):\penalty0 2278--2324, 1998.

\bibitem[Li et~al.(2023)Li, Zhang, Sun, Zou, Liu, Yang, Li, Zhang, and Gao]{li2023semantic}
Feng Li, Hao Zhang, Peize Sun, Xueyan Zou, Shilong Liu, Jianwei Yang, Chunyuan Li, Lei Zhang, and Jianfeng Gao.
\newblock Semantic-sam: Segment and recognize anything at any granularity.
\newblock \emph{arXiv preprint arXiv:2307.04767}, 2023.

\bibitem[Lin et~al.(2014)Lin, Maire, Belongie, Hays, Perona, Ramanan, Doll{\'a}r, and Zitnick]{lin2014microsoft}
Tsung-Yi Lin, Michael Maire, Serge Belongie, James Hays, Pietro Perona, Deva Ramanan, Piotr Doll{\'a}r, and C~Lawrence Zitnick.
\newblock Microsoft coco: Common objects in context.
\newblock In \emph{Computer Vision--ECCV 2014: 13th European Conference, Zurich, Switzerland, September 6-12, 2014, Proceedings, Part V 13}, pages 740--755. Springer, 2014.

\bibitem[Lin et~al.(2017)Lin, Doll{\'a}r, Girshick, He, Hariharan, and Belongie]{lin2017feature}
Tsung-Yi Lin, Piotr Doll{\'a}r, Ross Girshick, Kaiming He, Bharath Hariharan, and Serge Belongie.
\newblock Feature pyramid networks for object detection.
\newblock In \emph{Proceedings of the IEEE conference on computer vision and pattern recognition}, pages 2117--2125, 2017.

\bibitem[Liu et~al.(2023)Liu, Zeng, Ren, Li, Zhang, Yang, Li, Yang, Su, Zhu, et~al.]{liu2023grounding}
Shilong Liu, Zhaoyang Zeng, Tianhe Ren, Feng Li, Hao Zhang, Jie Yang, Chunyuan Li, Jianwei Yang, Hang Su, Jun Zhu, et~al.
\newblock Grounding dino: Marrying dino with grounded pre-training for open-set object detection.
\newblock \emph{arXiv preprint arXiv:2303.05499}, 2023.

\bibitem[Liu et~al.(2022)Liu, Mao, Wu, Feichtenhofer, Darrell, and Xie]{liu2022convnet}
Zhuang Liu, Hanzi Mao, Chao-Yuan Wu, Christoph Feichtenhofer, Trevor Darrell, and Saining Xie.
\newblock A convnet for the 2020s.
\newblock In \emph{Proceedings of the IEEE/CVF conference on computer vision and pattern recognition}, pages 11976--11986, 2022.

\bibitem[Luo et~al.(2019)Luo, Zheng, Guan, Yu, and Yang]{luo2019taking}
Yawei Luo, Liang Zheng, Tao Guan, Junqing Yu, and Yi Yang.
\newblock Taking a closer look at domain shift: Category-level adversaries for semantics consistent domain adaptation.
\newblock In \emph{Proceedings of the IEEE/CVF Conference on Computer Vision and Pattern Recognition}, pages 2507--2516, 2019.

\bibitem[Ma and Wang(2023)]{ma2023segment}
Jun Ma and Bo Wang.
\newblock Segment anything in medical images.
\newblock \emph{arXiv preprint arXiv:2304.12306}, 2023.

\bibitem[Momeni et~al.(2023)Momeni, Caron, Nagrani, Zisserman, and Schmid]{momeni2023verbs}
Liliane Momeni, Mathilde Caron, Arsha Nagrani, Andrew Zisserman, and Cordelia Schmid.
\newblock Verbs in action: Improving verb understanding in video-language models.
\newblock In \emph{Proceedings of the IEEE/CVF International Conference on Computer Vision}, pages 15579--15591, 2023.

\bibitem[Motiian et~al.(2017)Motiian, Jones, Iranmanesh, and Doretto]{motiian2017few}
Saeid Motiian, Quinn Jones, Seyed Iranmanesh, and Gianfranco Doretto.
\newblock Few-shot adversarial domain adaptation.
\newblock \emph{Advances in neural information processing systems}, 30, 2017.

\bibitem[Muandet et~al.(2013)Muandet, Balduzzi, and Sch{\"o}lkopf]{muandet2013domain}
Krikamol Muandet, David Balduzzi, and Bernhard Sch{\"o}lkopf.
\newblock Domain generalization via invariant feature representation.
\newblock In \emph{International conference on machine learning}, pages 10--18. PMLR, 2013.

\bibitem[Neuhold et~al.(2017)Neuhold, Ollmann, Rota~Bulo, and Kontschieder]{neuhold2017mapillary}
Gerhard Neuhold, Tobias Ollmann, Samuel Rota~Bulo, and Peter Kontschieder.
\newblock The mapillary vistas dataset for semantic understanding of street scenes.
\newblock In \emph{ICCV}, pages 4990--4999, 2017.

\bibitem[Niemeijer et~al.(2024)Niemeijer, Schwonberg, Term\"ohlen, Schmidt, and Fingscheidt]{Niemeijer_2024_WACV}
Joshua Niemeijer, Manuel Schwonberg, Jan-Aike Term\"ohlen, Nico~M. Schmidt, and Tim Fingscheidt.
\newblock Generalization by adaptation: Diffusion-based domain extension for domain-generalized semantic segmentation.
\newblock In \emph{Proceedings of the IEEE/CVF Winter Conference on Applications of Computer Vision (WACV)}, pages 2830--2840, 2024.

\bibitem[Pan et~al.(2018)Pan, Luo, Shi, and Tang]{pan2018two}
Xingang Pan, Ping Luo, Jianping Shi, and Xiaoou Tang.
\newblock Two at once: Enhancing learning and generalization capacities via ibn-net.
\newblock In \emph{Proceedings of the European Conference on Computer Vision (ECCV)}, pages 464--479, 2018.

\bibitem[Pan et~al.(2019)Pan, Zhan, Shi, Tang, and Luo]{pan2019switchable}
Xingang Pan, Xiaohang Zhan, Jianping Shi, Xiaoou Tang, and Ping Luo.
\newblock Switchable whitening for deep representation learning.
\newblock In \emph{ICCV}, pages 1863--1871, 2019.

\bibitem[Peng et~al.(2021)Peng, Lei, Liu, Zhang, and Liu]{peng2021global}
Duo Peng, Yinjie Lei, Lingqiao Liu, Pingping Zhang, and Jun Liu.
\newblock Global and local texture randomization for synthetic-to-real semantic segmentation.
\newblock \emph{IEEE TIP}, pages 6594--6608, 2021.

\bibitem[Peng et~al.(2022)Peng, Lei, Hayat, Guo, and Li]{peng2022semantic}
Duo Peng, Yinjie Lei, Munawar Hayat, Yulan Guo, and Wen Li.
\newblock Semantic-aware domain generalized segmentation.
\newblock In \emph{Proceedings of the IEEE/CVF Conference on Computer Vision and Pattern Recognition}, pages 2594--2605, 2022.

\bibitem[Pratt et~al.(2023)Pratt, Covert, Liu, and Farhadi]{pratt2023does}
Sarah Pratt, Ian Covert, Rosanne Liu, and Ali Farhadi.
\newblock What does a platypus look like? generating customized prompts for zero-shot image classification.
\newblock In \emph{Proceedings of the IEEE/CVF International Conference on Computer Vision}, pages 15691--15701, 2023.

\bibitem[Qiao et~al.(2020)Qiao, Zhao, and Peng]{qiao2020learning}
Fengchun Qiao, Long Zhao, and Xi Peng.
\newblock Learning to learn single domain generalization.
\newblock In \emph{CVPR}, pages 12556--12565, 2020.

\bibitem[Radford et~al.(2021)Radford, Kim, Hallacy, Ramesh, Goh, Agarwal, Sastry, Askell, Mishkin, Clark, et~al.]{radford2021learning}
Alec Radford, Jong~Wook Kim, Chris Hallacy, Aditya Ramesh, Gabriel Goh, Sandhini Agarwal, Girish Sastry, Amanda Askell, Pamela Mishkin, Jack Clark, et~al.
\newblock Learning transferable visual models from natural language supervision.
\newblock In \emph{International conference on machine learning}, pages 8748--8763. PMLR, 2021.

\bibitem[Ramesh et~al.(2022)Ramesh, Dhariwal, Nichol, Chu, and Chen]{ramesh2022hierarchical}
Aditya Ramesh, Prafulla Dhariwal, Alex Nichol, Casey Chu, and Mark Chen.
\newblock Hierarchical text-conditional image generation with clip latents.
\newblock \emph{arXiv preprint arXiv:2204.06125}, 2022.

\bibitem[Richter et~al.(2016)Richter, Vineet, Roth, and Koltun]{richter2016playing}
Stephan~R Richter, Vibhav Vineet, Stefan Roth, and Vladlen Koltun.
\newblock Playing for data: Ground truth from computer games.
\newblock In \emph{ECCV}, pages 102--118. Springer, 2016.

\bibitem[Rombach and Esser(2023)]{sdv1_4}
Robin Rombach and Patrick Esser.
\newblock Compvis/stable-diffusion-v1-4.
\newblock \url{https://huggingface.co/CompVis/stable-diffusion-v1-4}, 2023.
\newblock Accessed on March 20, 2023.

\bibitem[Rombach et~al.(2022)Rombach, Blattmann, Lorenz, Esser, and Ommer]{rombach2022high}
Robin Rombach, Andreas Blattmann, Dominik Lorenz, Patrick Esser, and Bj{\"o}rn Ommer.
\newblock High-resolution image synthesis with latent diffusion models.
\newblock In \emph{Proceedings of the IEEE/CVF Conference on Computer Vision and Pattern Recognition}, pages 10684--10695, 2022.

\bibitem[Ros et~al.(2016)Ros, Sellart, Materzynska, Vazquez, and Lopez]{ros2016synthia}
German Ros, Laura Sellart, Joanna Materzynska, David Vazquez, and Antonio~M Lopez.
\newblock The synthia dataset: A large collection of synthetic images for semantic segmentation of urban scenes.
\newblock In \emph{CVPR}, pages 3234--3243, 2016.

\bibitem[Saenko et~al.(2010)Saenko, Kulis, Fritz, and Darrell]{saenko2010adapting}
Kate Saenko, Brian Kulis, Mario Fritz, and Trevor Darrell.
\newblock Adapting visual category models to new domains.
\newblock In \emph{Computer Vision--ECCV 2010: 11th European Conference on Computer Vision, Heraklion, Crete, Greece, September 5-11, 2010, Proceedings, Part IV 11}, pages 213--226. Springer, 2010.

\bibitem[Sakaridis et~al.(2021)Sakaridis, Dai, and Van~Gool]{sakaridis2021acdc}
Christos Sakaridis, Dengxin Dai, and Luc Van~Gool.
\newblock Acdc: The adverse conditions dataset with correspondences for semantic driving scene understanding.
\newblock In \emph{Proceedings of the IEEE/CVF International Conference on Computer Vision}, pages 10765--10775, 2021.

\bibitem[Sariyildiz et~al.(2023)Sariyildiz, Alahari, Larlus, and Kalantidis]{sariyildiz2023fake}
Mert~Bulent Sariyildiz, Karteek Alahari, Diane Larlus, and Yannis Kalantidis.
\newblock Fake it till you make it: Learning transferable representations from synthetic imagenet clones.
\newblock In \emph{CVPR 2023--IEEE/CVF Conference on Computer Vision and Pattern Recognition}, 2023.

\bibitem[Schuhmann et~al.(2022)Schuhmann, Beaumont, Vencu, Gordon, Wightman, Cherti, Coombes, Katta, Mullis, Wortsman, et~al.]{schuhmann2022laion}
Christoph Schuhmann, Romain Beaumont, Richard Vencu, Cade Gordon, Ross Wightman, Mehdi Cherti, Theo Coombes, Aarush Katta, Clayton Mullis, Mitchell Wortsman, et~al.
\newblock Laion-5b: An open large-scale dataset for training next generation image-text models.
\newblock \emph{Advances in Neural Information Processing Systems}, 35:\penalty0 25278--25294, 2022.

\bibitem[Tang et~al.(2020)Tang, Gao, Zhu, Zhang, Li, and Metaxas]{tang2020selfnorm}
Zhiqiang Tang, Yunhe Gao, Yi Zhu, Zhi Zhang, Mu Li, and Dimitris~N Metaxas.
\newblock Selfnorm and crossnorm for out-of-distribution robustness.
\newblock 2020.

\bibitem[Torralba and Efros(2011)]{torralba2011unbiased}
Antonio Torralba and Alexei~A Efros.
\newblock Unbiased look at dataset bias.
\newblock In \emph{CVPR 2011}, pages 1521--1528. IEEE, 2011.

\bibitem[Touvron et~al.(2023)Touvron, Martin, Stone, Albert, Almahairi, Babaei, Bashlykov, Batra, Bhargava, Bhosale, et~al.]{touvron2023llama}
Hugo Touvron, Louis Martin, Kevin Stone, Peter Albert, Amjad Almahairi, Yasmine Babaei, Nikolay Bashlykov, Soumya Batra, Prajjwal Bhargava, Shruti Bhosale, et~al.
\newblock Llama 2: Open foundation and fine-tuned chat models.
\newblock \emph{arXiv preprint arXiv:2307.09288}, 2023.

\bibitem[Volpi et~al.(2018)Volpi, Namkoong, Sener, Duchi, Murino, and Savarese]{volpi2018generalizing}
Riccardo Volpi, Hongseok Namkoong, Ozan Sener, John~C Duchi, Vittorio Murino, and Silvio Savarese.
\newblock Generalizing to unseen domains via adversarial data augmentation.
\newblock \emph{Advances in neural information processing systems}, 31, 2018.

\bibitem[Volpi et~al.(2021)Volpi, Larlus, and Rogez]{volpi2021continual}
Riccardo Volpi, Diane Larlus, and Gr{\'e}gory Rogez.
\newblock Continual adaptation of visual representations via domain randomization and meta-learning.
\newblock In \emph{Proceedings of the IEEE/CVF Conference on Computer Vision and Pattern Recognition}, pages 4443--4453, 2021.

\bibitem[Wang et~al.(2020)Wang, Sun, Cheng, Jiang, Deng, Zhao, Liu, Mu, Tan, Wang, et~al.]{wang2020deep}
Jingdong Wang, Ke Sun, Tianheng Cheng, Borui Jiang, Chaorui Deng, Yang Zhao, Dong Liu, Yadong Mu, Mingkui Tan, Xinggang Wang, et~al.
\newblock Deep high-resolution representation learning for visual recognition.
\newblock \emph{IEEE transactions on pattern analysis and machine intelligence}, 43\penalty0 (10):\penalty0 3349--3364, 2020.

\bibitem[Wu et~al.(2023)Wu, Fu, Fang, Liu, Wang, Xu, Jin, and Arbel]{wu2023medical}
Junde Wu, Rao Fu, Huihui Fang, Yuanpei Liu, Zhaowei Wang, Yanwu Xu, Yueming Jin, and Tal Arbel.
\newblock Medical sam adapter: Adapting segment anything model for medical image segmentation.
\newblock \emph{arXiv preprint arXiv:2304.12620}, 2023.

\bibitem[Xie et~al.(2021)Xie, Wang, Yu, Anandkumar, Alvarez, and Luo]{xie2021segformer}
Enze Xie, Wenhai Wang, Zhiding Yu, Anima Anandkumar, Jose~M Alvarez, and Ping Luo.
\newblock Seg{F}ormer: Simple and efficient design for semantic segmentation with transformers.
\newblock \emph{NeurIPS}, pages 12077--12090, 2021.

\bibitem[Xu et~al.(2023)Xu, Liu, Vahdat, Byeon, Wang, and De~Mello]{xu2023open}
Jiarui Xu, Sifei Liu, Arash Vahdat, Wonmin Byeon, Xiaolong Wang, and Shalini De~Mello.
\newblock Open-vocabulary panoptic segmentation with text-to-image diffusion models.
\newblock In \emph{Proceedings of the IEEE/CVF Conference on Computer Vision and Pattern Recognition}, pages 2955--2966, 2023.

\bibitem[Xu et~al.(2021)Xu, Zhang, Zhang, Wang, and Tian]{xu2021fourier}
Qinwei Xu, Ruipeng Zhang, Ya Zhang, Yanfeng Wang, and Qi Tian.
\newblock A fourier-based framework for domain generalization.
\newblock In \emph{Proceedings of the IEEE/CVF Conference on Computer Vision and Pattern Recognition}, pages 14383--14392, 2021.

\bibitem[Yang et~al.(2023)Yang, Panagopoulou, Zhou, Jin, Callison-Burch, and Yatskar]{Yang_2023_CVPR}
Yue Yang, Artemis Panagopoulou, Shenghao Zhou, Daniel Jin, Chris Callison-Burch, and Mark Yatskar.
\newblock Language in a bottle: Language model guided concept bottlenecks for interpretable image classification.
\newblock In \emph{Proceedings of the IEEE/CVF Conference on Computer Vision and Pattern Recognition (CVPR)}, pages 19187--19197, 2023.

\bibitem[Yu et~al.(2020)Yu, Chen, Wang, Xian, Chen, Liu, Madhavan, and Darrell]{yu2020bdd100k}
Fisher Yu, Haofeng Chen, Xin Wang, Wenqi Xian, Yingying Chen, Fangchen Liu, Vashisht Madhavan, and Trevor Darrell.
\newblock Bdd100k: A diverse driving dataset for heterogeneous multitask learning.
\newblock In \emph{CVPR}, pages 2636--2645, 2020.

\bibitem[Yu et~al.(2023)Yu, He, Deng, Shen, and Chen]{yu2023convolutions}
Qihang Yu, Ju He, Xueqing Deng, Xiaohui Shen, and Liang-Chieh Chen.
\newblock Convolutions die hard: Open-vocabulary segmentation with single frozen convolutional clip.
\newblock \emph{arXiv preprint arXiv:2308.02487}, 2023.

\bibitem[Yue et~al.(2019)Yue, Zhang, Zhao, Sangiovanni-Vincentelli, Keutzer, and Gong]{yue2019domain}
Xiangyu Yue, Yang Zhang, Sicheng Zhao, Alberto Sangiovanni-Vincentelli, Kurt Keutzer, and Boqing Gong.
\newblock Domain randomization and pyramid consistency: Simulation-to-real generalization without accessing target domain data.
\newblock In \emph{ICCV}, pages 2100--2110, 2019.

\bibitem[Zara et~al.(2023)Zara, Conti, Roy, Lathuili{\`e}re, Rota, and Ricci]{zara2023unreasonable}
Giacomo Zara, Alessandro Conti, Subhankar Roy, St{\'e}phane Lathuili{\`e}re, Paolo Rota, and Elisa Ricci.
\newblock The unreasonable effectiveness of large language-vision models for source-free video domain adaptation.
\newblock In \emph{Proceedings of the IEEE/CVF International Conference on Computer Vision}, pages 10307--10317, 2023.

\bibitem[Zhang et~al.(2021)Zhang, Zhang, Zhang, Chen, Wang, and Wen]{zhang2021prototypical}
Pan Zhang, Bo Zhang, Ting Zhang, Dong Chen, Yong Wang, and Fang Wen.
\newblock Prototypical pseudo label denoising and target structure learning for domain adaptive semantic segmentation.
\newblock In \emph{Proceedings of the IEEE/CVF conference on computer vision and pattern recognition}, pages 12414--12424, 2021.

\bibitem[Zhao et~al.(2022{\natexlab{a}})Zhao, Zhong, Luo, Lee, and Sebe]{zhao2022source}
Yuyang Zhao, Zhun Zhong, Zhiming Luo, Gim~Hee Lee, and Nicu Sebe.
\newblock Source-free open compound domain adaptation in semantic segmentation.
\newblock \emph{IEEE Transactions on Circuits and Systems for Video Technology}, 32\penalty0 (10):\penalty0 7019--7032, 2022{\natexlab{a}}.

\bibitem[Zhao et~al.(2022{\natexlab{b}})Zhao, Zhong, Zhao, Sebe, and Lee]{zhao2022style}
Yuyang Zhao, Zhun Zhong, Na Zhao, Nicu Sebe, and Gim~Hee Lee.
\newblock Style-hallucinated dual consistency learning for domain generalized semantic segmentation.
\newblock In \emph{ECCV}, pages 535--552. Springer, 2022{\natexlab{b}}.

\bibitem[Zhong et~al.(2022)Zhong, Zhao, Lee, and Sebe]{zhong2022adversarial}
Zhun Zhong, Yuyang Zhao, Gim~Hee Lee, and Nicu Sebe.
\newblock Adversarial style augmentation for domain generalized urban-scene segmentation.
\newblock \emph{Advances in Neural Information Processing Systems}, 35:\penalty0 338--350, 2022.

\bibitem[Zou et~al.(2018)Zou, Yu, Kumar, and Wang]{zou2018unsupervised}
Yang Zou, Zhiding Yu, BVK Kumar, and Jinsong Wang.
\newblock Unsupervised domain adaptation for semantic segmentation via class-balanced self-training.
\newblock In \emph{Proceedings of the European conference on computer vision (ECCV)}, pages 289--305, 2018.

\bibitem[Zou et~al.(2019)Zou, Yu, Liu, Kumar, and Wang]{zou2019confidence}
Yang Zou, Zhiding Yu, Xiaofeng Liu, BVK Kumar, and Jinsong Wang.
\newblock Confidence regularized self-training.
\newblock In \emph{Proceedings of the IEEE/CVF international conference on computer vision}, pages 5982--5991, 2019.

\end{thebibliography}
}
\appendix
\clearpage
\maketitlesupplementary

\setcounter{table}{0}
\renewcommand{\thetable}{A\arabic{table}}
\setcounter{figure}{0}
\renewcommand{\thefigure}{A\arabic{figure}}
\setcounter{equation}{0}
\renewcommand{\theequation}{A\arabic{equation}}

The supplementary material is organized as follows:
We provide further details about the experiments in Sec.~\ref{sec:exp-details}, and in Sec.~\ref{sec:sota-quali} we display a side-by-side qualitative comparison of \method and other state of the art methods. In Sec.~\ref{sec:components-quali}, we examine the qualitative effects of incorporating different foundation models into our framework. Sec.~\ref{sec:sam-quali} shows the effect of SAM in refining the pseudo labels for self-training. Sec.~\ref{sec:dm-quali} presents a collection of images produced by diffusion models, each guided by a distinct text prompt generated by the LLM.
\vspace{-2mm}
\section{Experimental details}
\label{sec:exp-details}
\vspace{-1mm}
\paragraph{Training losses:}
As mentionned in Sec.~\ref{sub:clip}, the source loss $\mathcal{L}_{S}$ is a linear combination of the mask loss $\mathcal{L}_{\text{mask}}$ and classification loss $\mathcal{L}_{\text{cls}}$. The mask loss can be expressed as follows : $\mathcal{L}_{\text{mask}}=\lambda_{\text{ce}} \mathcal{L}_{\text{ce}} + \lambda_{\text{dice}} \mathcal{L}_{\text{dice}}$, and we maintain the same values of $\lambda_{\text{ce}}$, $\lambda_{\text{dice}}$ and $\lambda_{\text{dice}}$ used in Mask2Former~\cite{cheng2022masked}.
The self-training loss $\mathcal{L}_{ST}$ is exactly similar to $\mathcal{L}_{S}$, with the key difference being its application to the images generated by the model.
\vspace{-15pt}
\paragraph{Prompting of the LLM:} We used the Llama2-70b-Chat version of LLama2, which is optimized for dialogue use cases, for more details about the architecture and training details, refer to~\cite{touvron2023llama}.
The full prompt that was used to prompt Llama2 is the following: \textit{I want a list of prompts that can be used by an image generation model to generate synthetic images. The prompt should strictly follow this template: "a photo of X in Z" where X contains one or multiple class names within {road, building, sidewalk, wall, fence, pole, traffic light, traffic sign, vegetation, grass, sky, person, rider, car, truck, bus, train, motorcycle, bicycle} put in the context of an urban street scene, you use other synonyms to increase diversity. Z contains a brief description of regular lighting and weather conditions. Can you provide 100 diverse and simple prompts.}
\section{Comparison with state of the art}
\label{sec:sota-quali}
\begin{figure*}[ht]
    \centering
          \begin{subfigure}[b]{0.19\linewidth}
            \includegraphics[width=\linewidth]{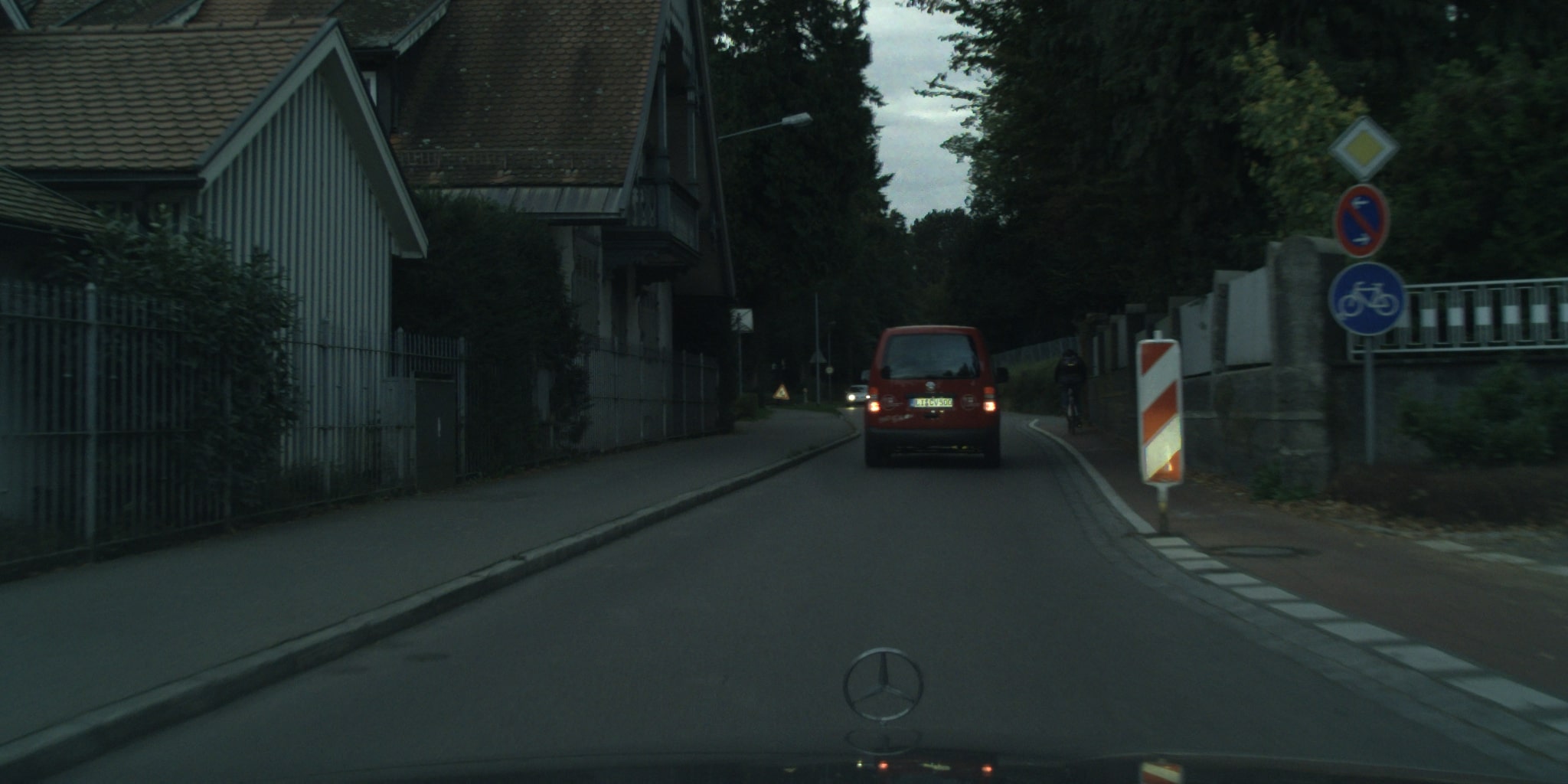}
          \end{subfigure}
          \hfill
          \begin{subfigure}[b]{0.19\linewidth}
            \includegraphics[width=\linewidth]{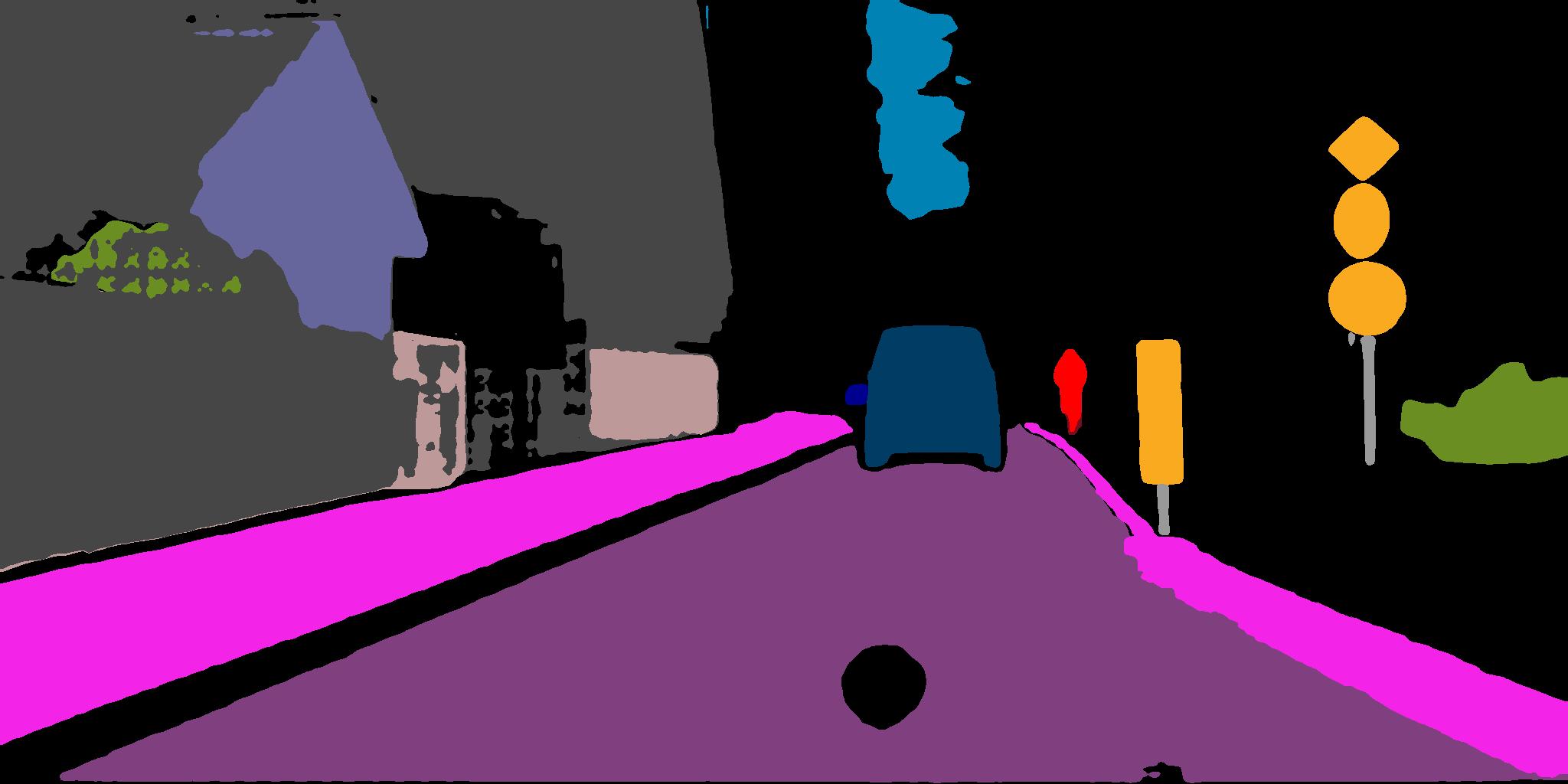}

          \end{subfigure}
          \hfill
          \begin{subfigure}[b]{0.19\linewidth}
            \includegraphics[width=\linewidth]{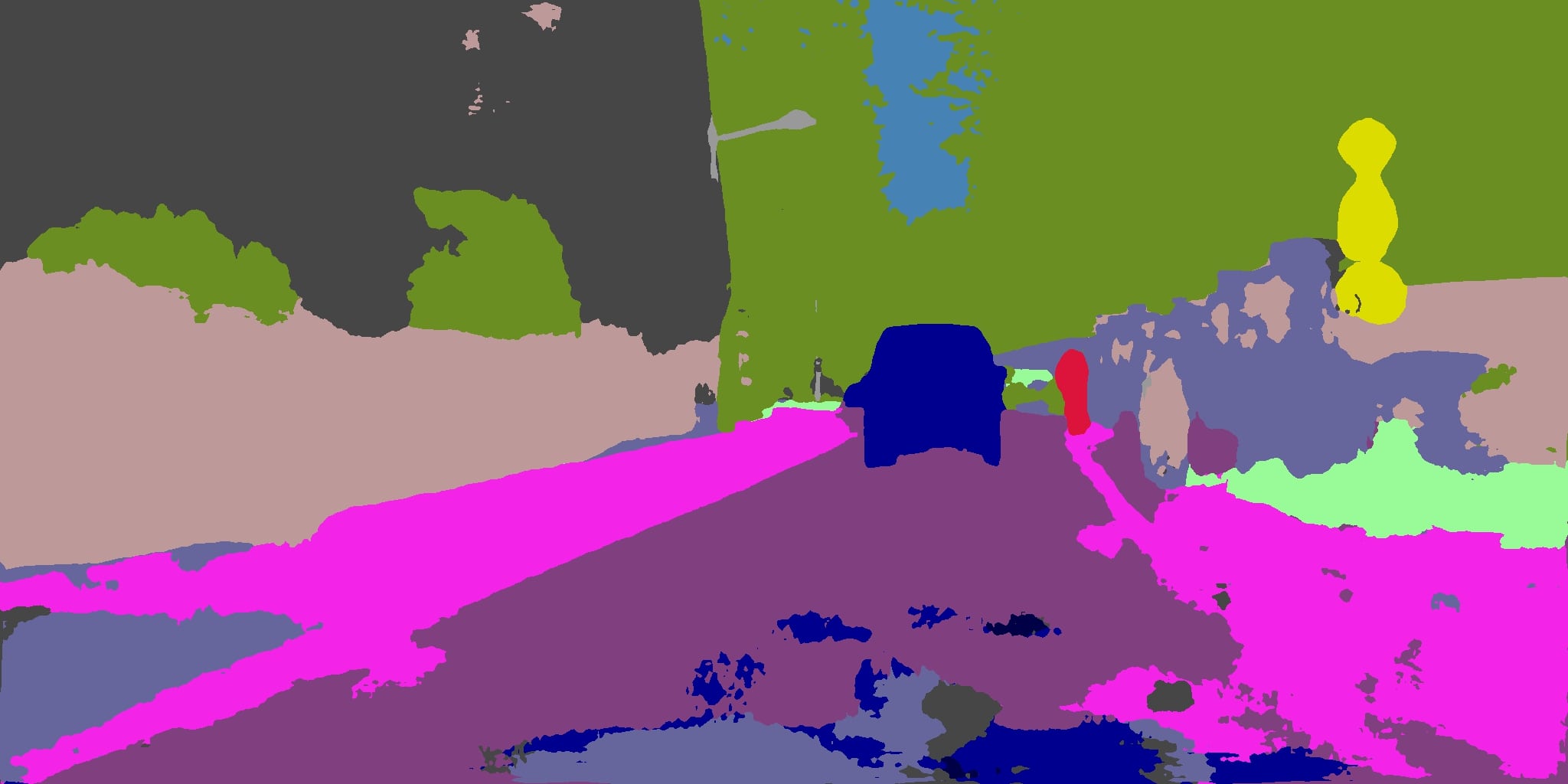}

          \end{subfigure}
          \hfill
          \begin{subfigure}[b]{0.19\linewidth}
            \includegraphics[width=\linewidth]{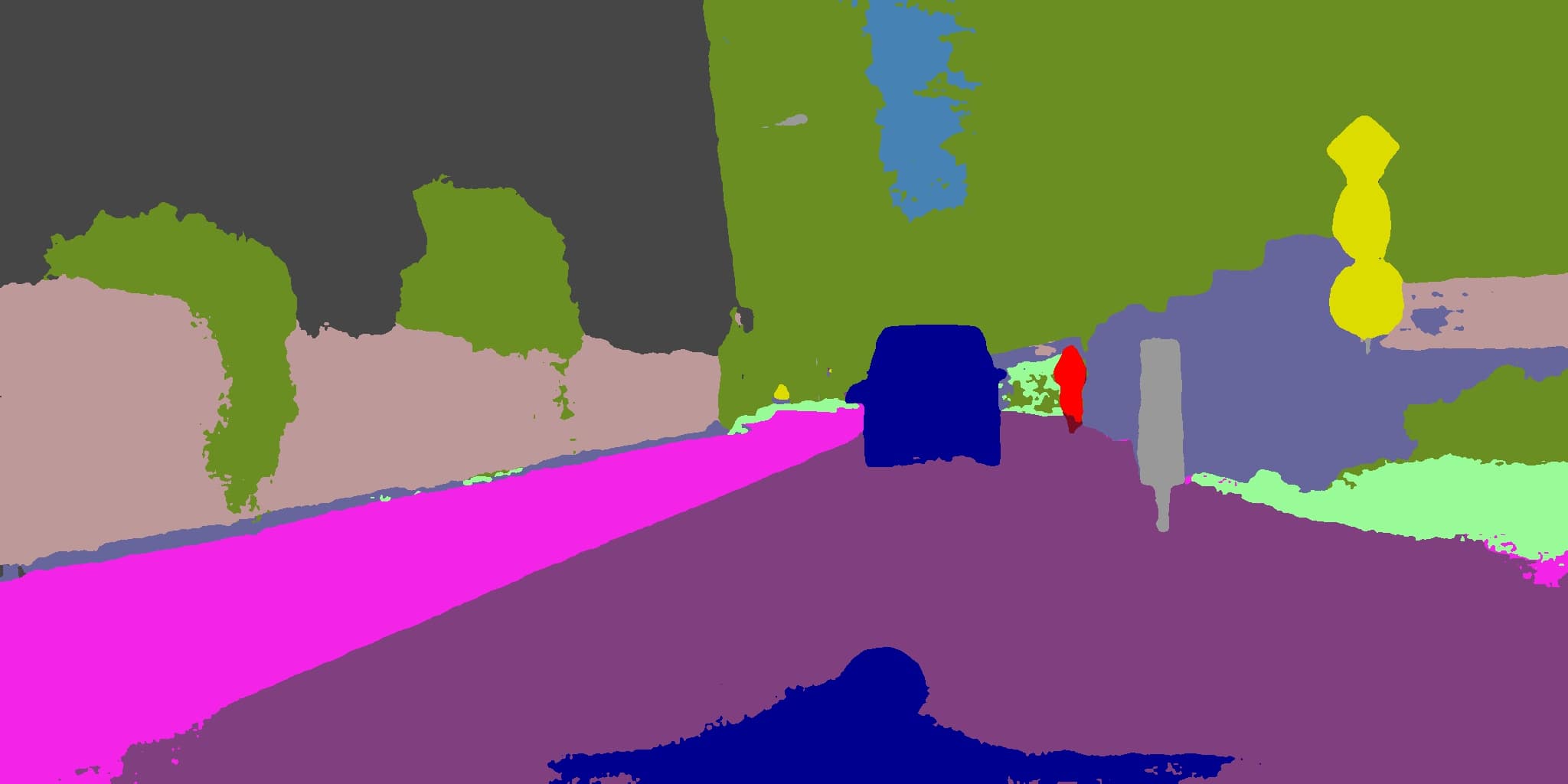}

          \end{subfigure}
          \hfill
          \begin{subfigure}[b]{0.19\linewidth}
            \includegraphics[width=\linewidth]{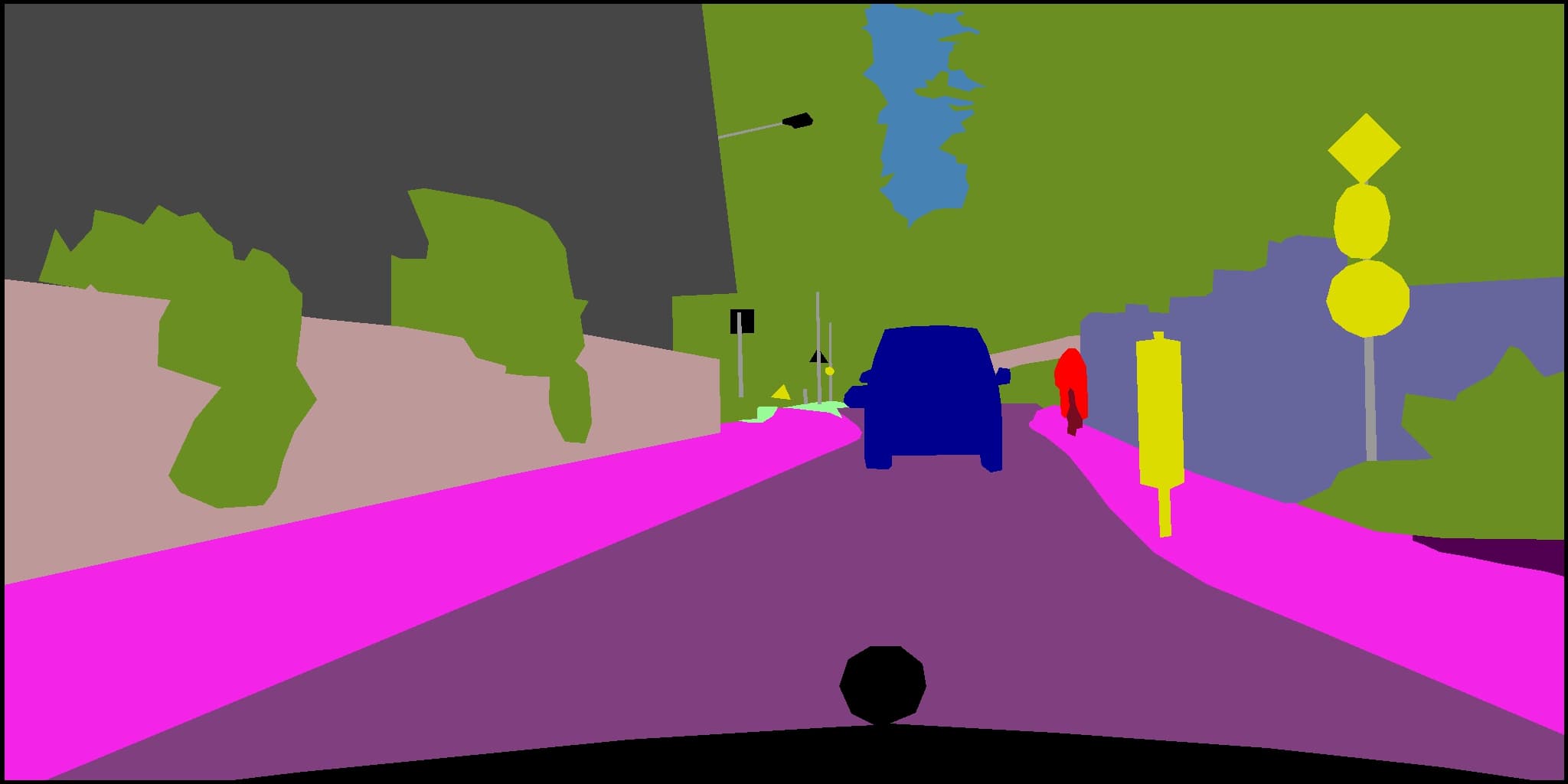}

          \end{subfigure} \\
          \begin{subfigure}[b]{0.19\linewidth}
            \includegraphics[width=\linewidth]{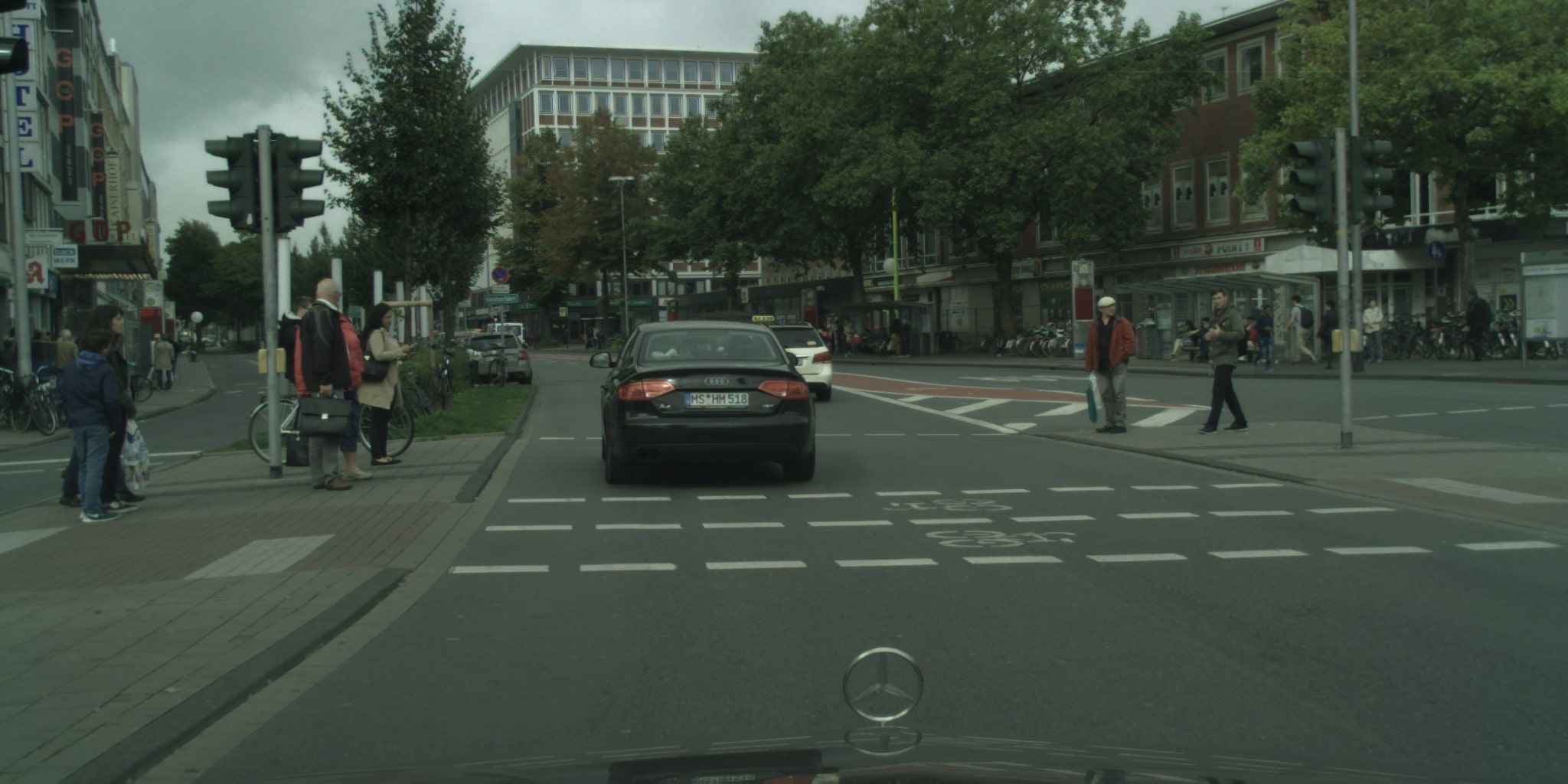}

          \end{subfigure}
          \hfill
          \begin{subfigure}[b]{0.19\linewidth}
            \includegraphics[width=\linewidth]{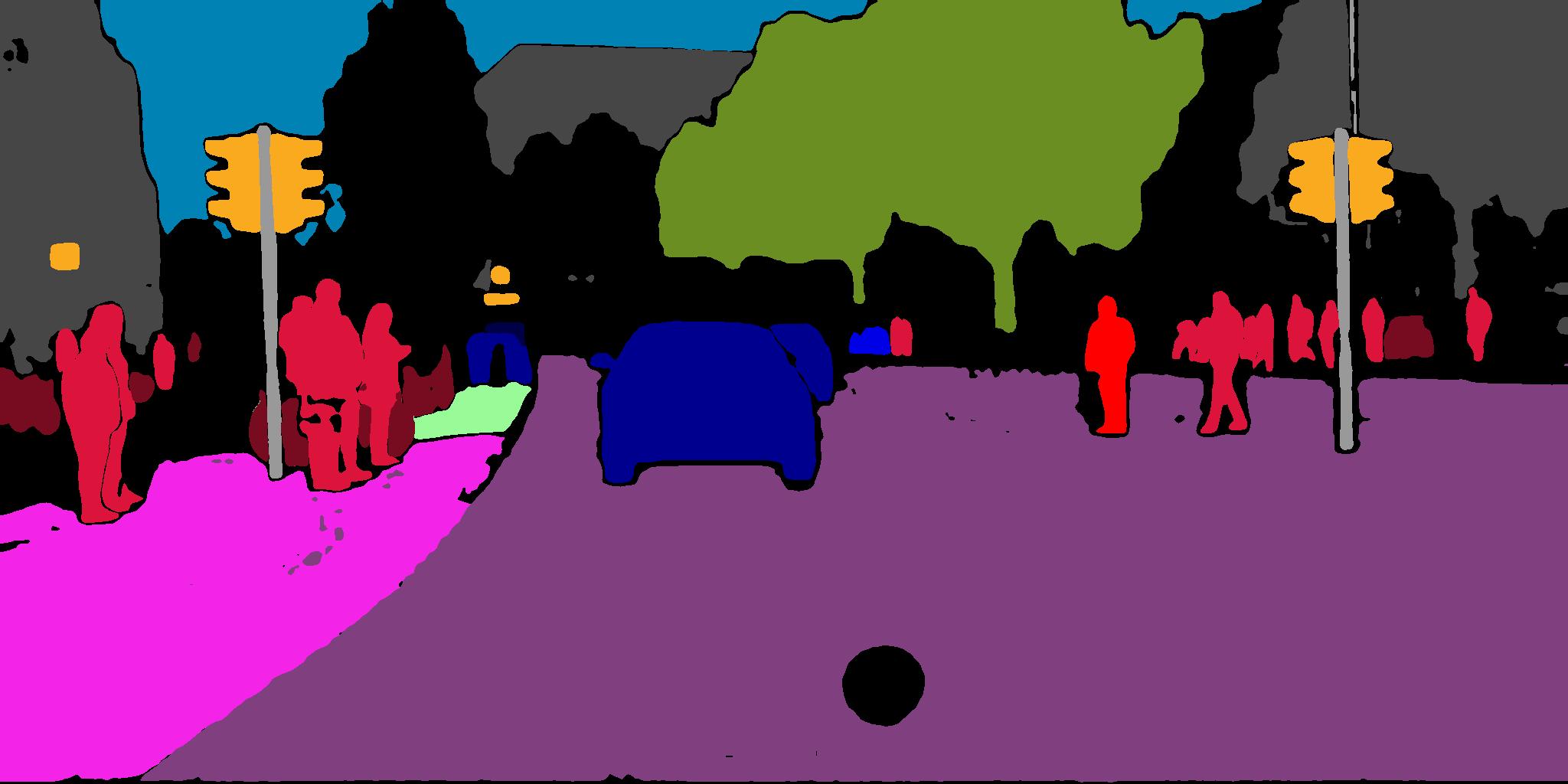}

          \end{subfigure}
          \hfill
          \begin{subfigure}[b]{0.19\linewidth}
            \includegraphics[width=\linewidth]{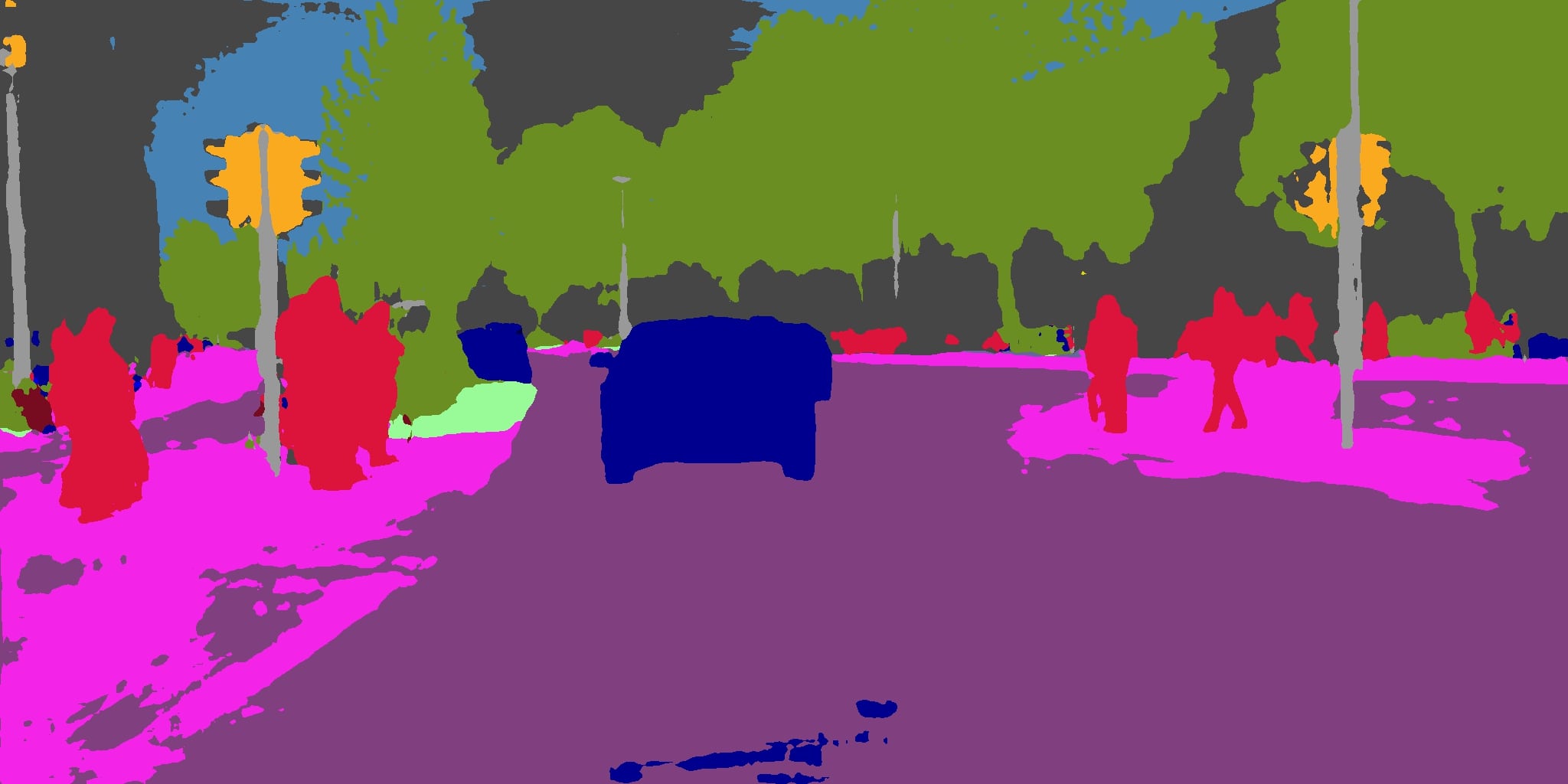}

          \end{subfigure}
          \hfill
          \begin{subfigure}[b]{0.19\linewidth}
            \includegraphics[width=\linewidth]{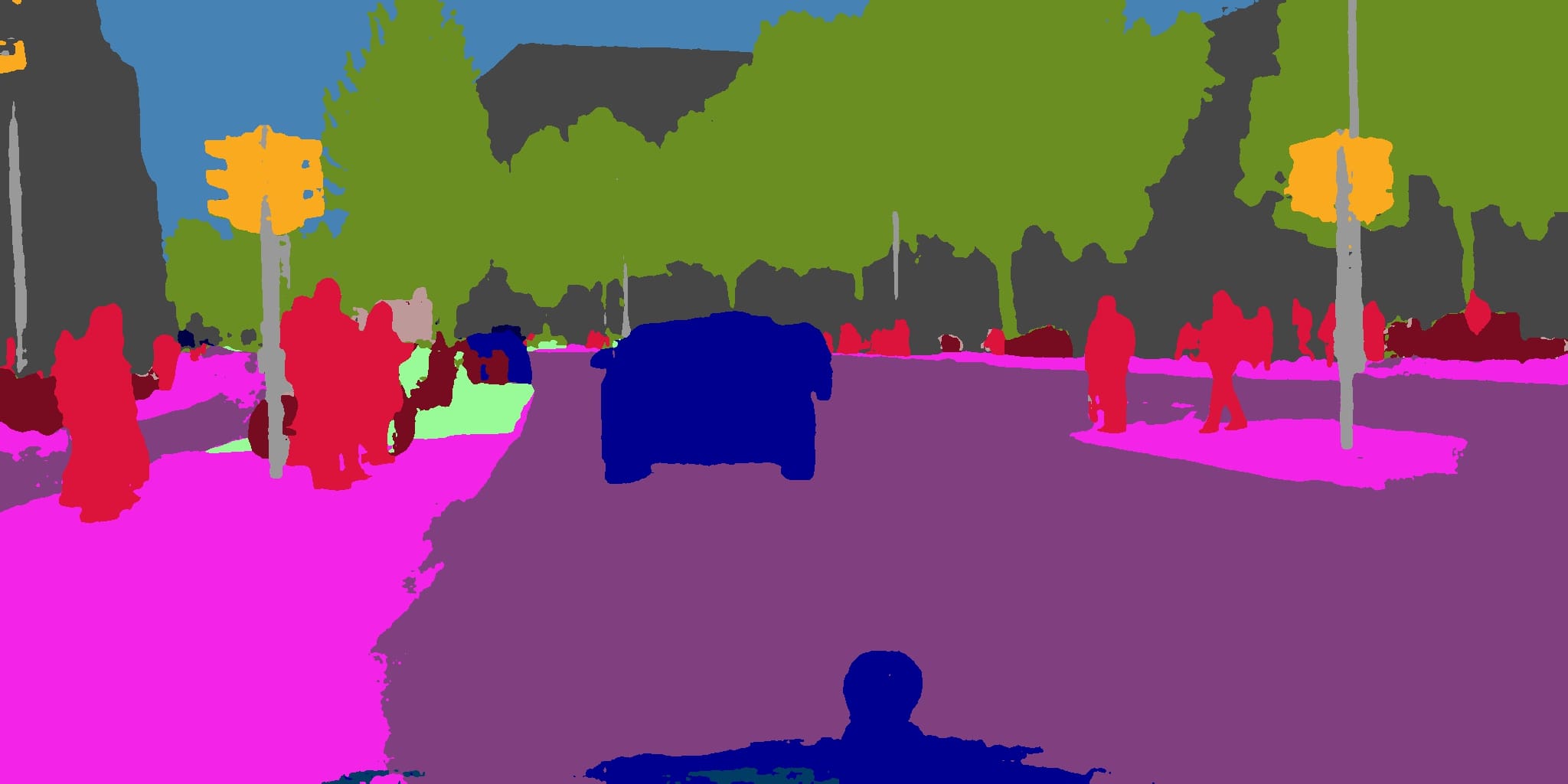}

          \end{subfigure}
          \hfill
          \begin{subfigure}[b]{0.19\linewidth}
            \includegraphics[width=\linewidth]{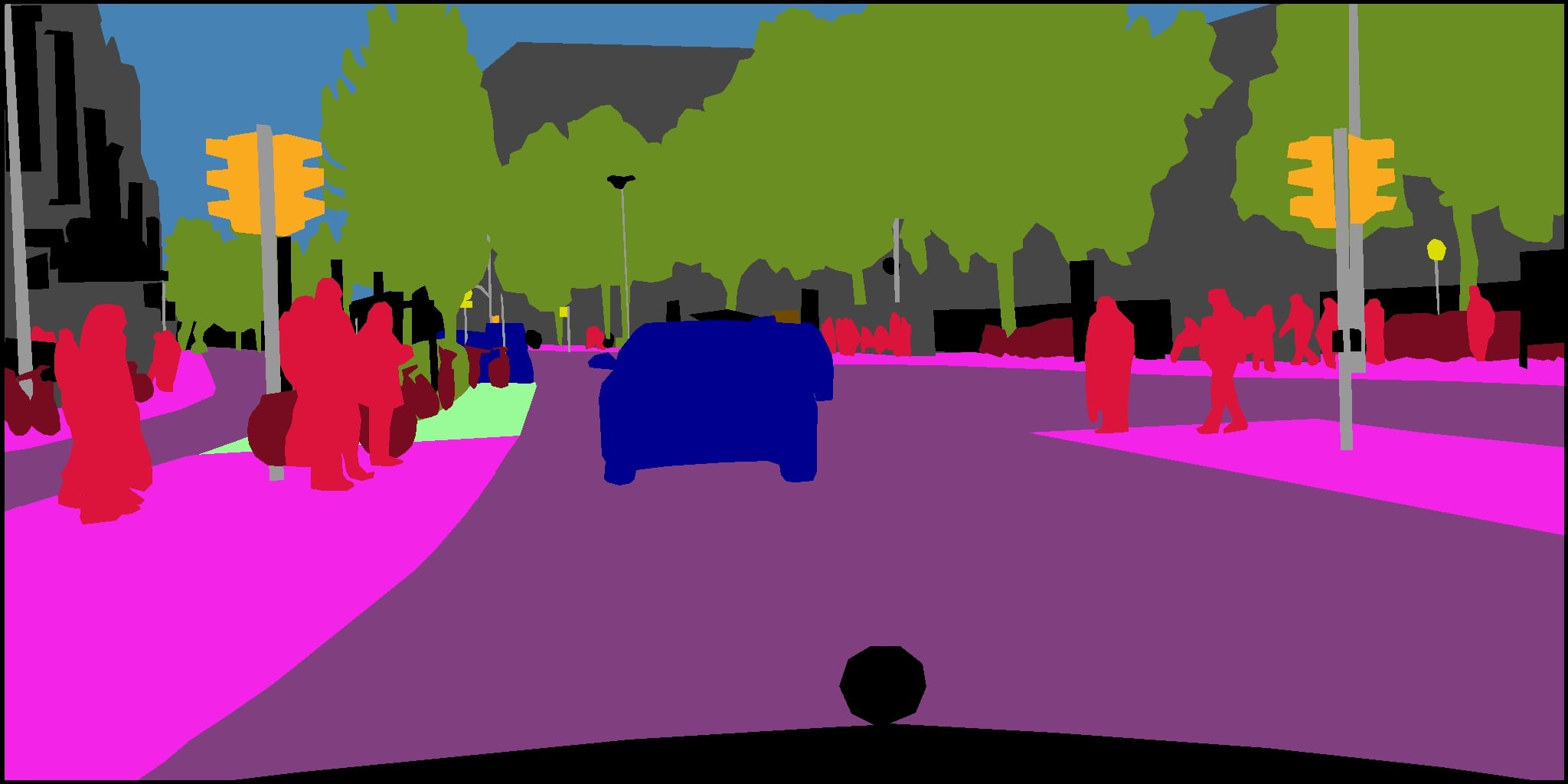}

          \end{subfigure}\\
          \begin{subfigure}[b]{0.19\linewidth}
            \includegraphics[width=\linewidth]{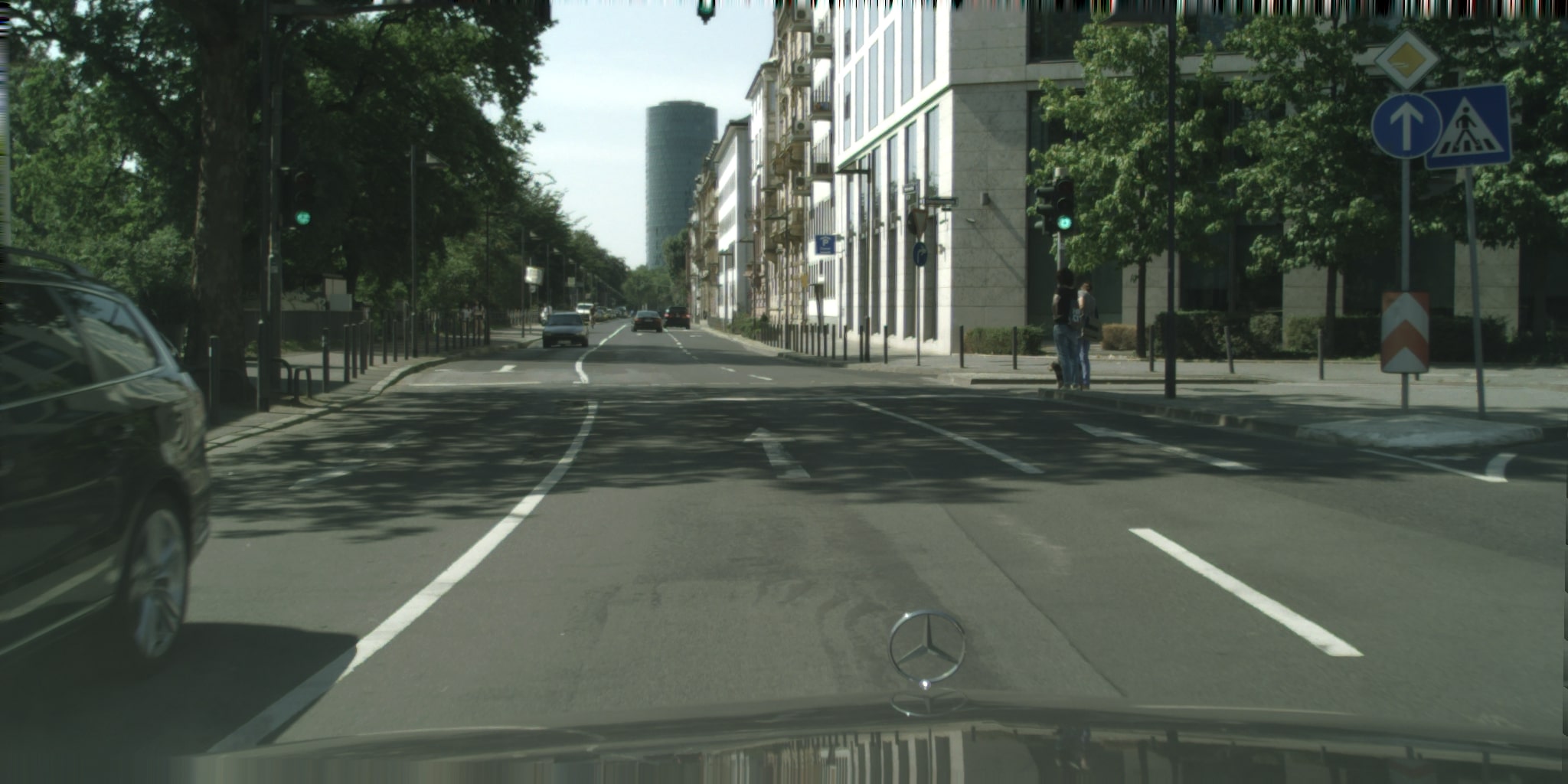}

          \end{subfigure}
          \hfill
          \begin{subfigure}[b]{0.19\linewidth}
            \includegraphics[width=\linewidth]{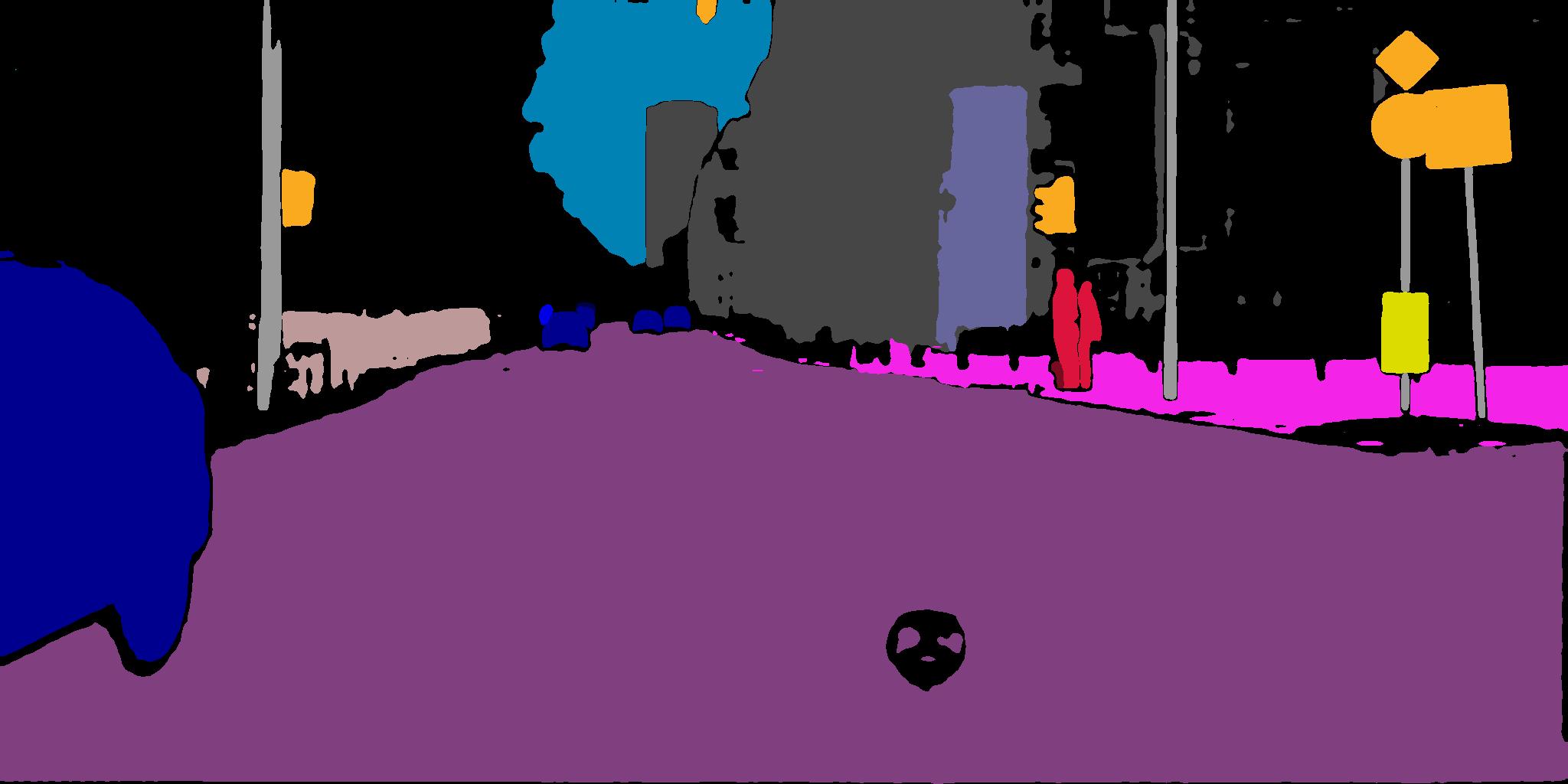}

          \end{subfigure}
          \hfill
          \begin{subfigure}[b]{0.19\linewidth}
            \includegraphics[width=\linewidth]{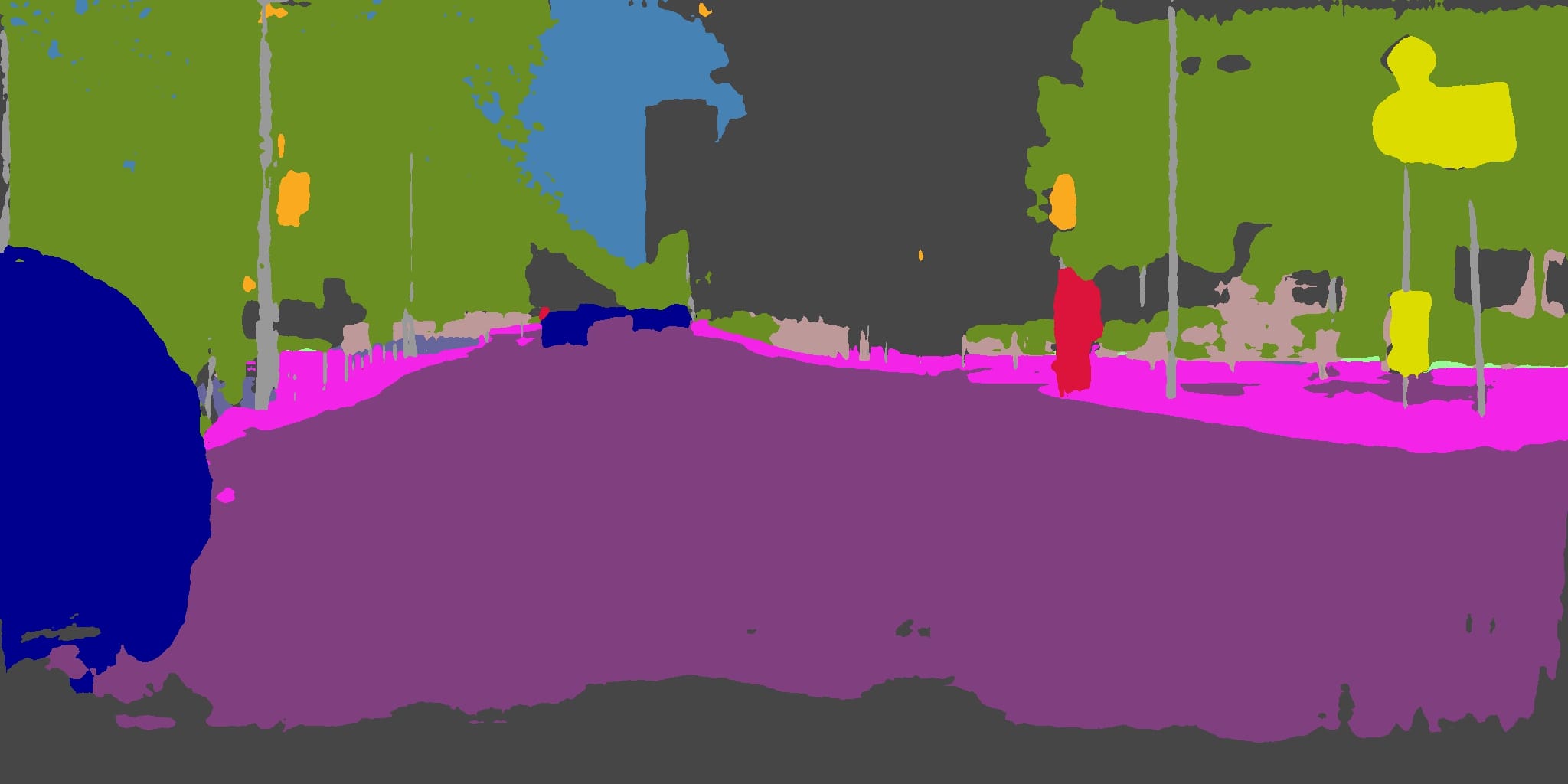}
 
          \end{subfigure}
          \hfill
          \begin{subfigure}[b]{0.19\linewidth}
            \includegraphics[width=\linewidth]{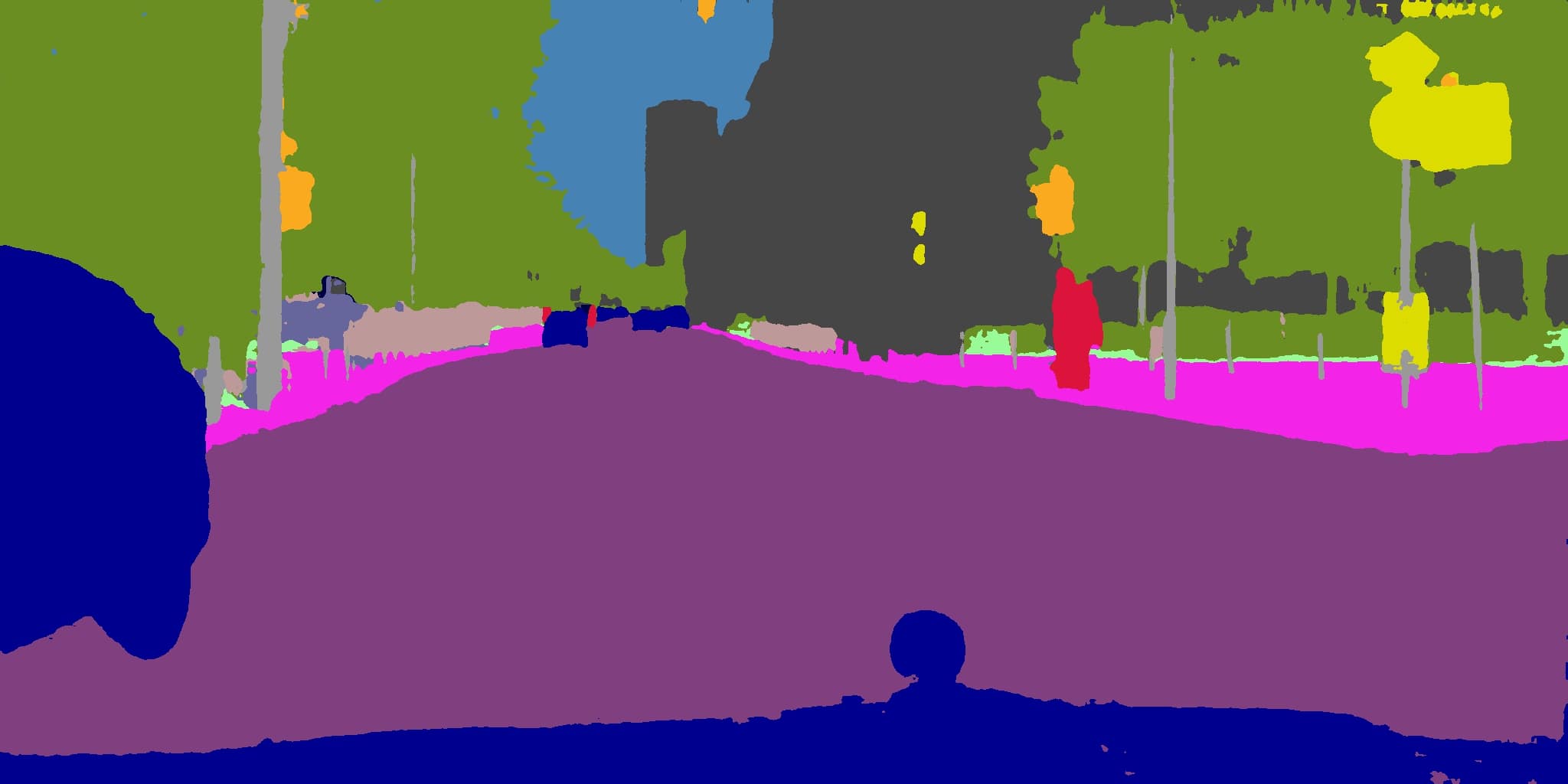}

          \end{subfigure}
          \hfill
          \begin{subfigure}[b]{0.19\linewidth}
            \includegraphics[width=\linewidth]{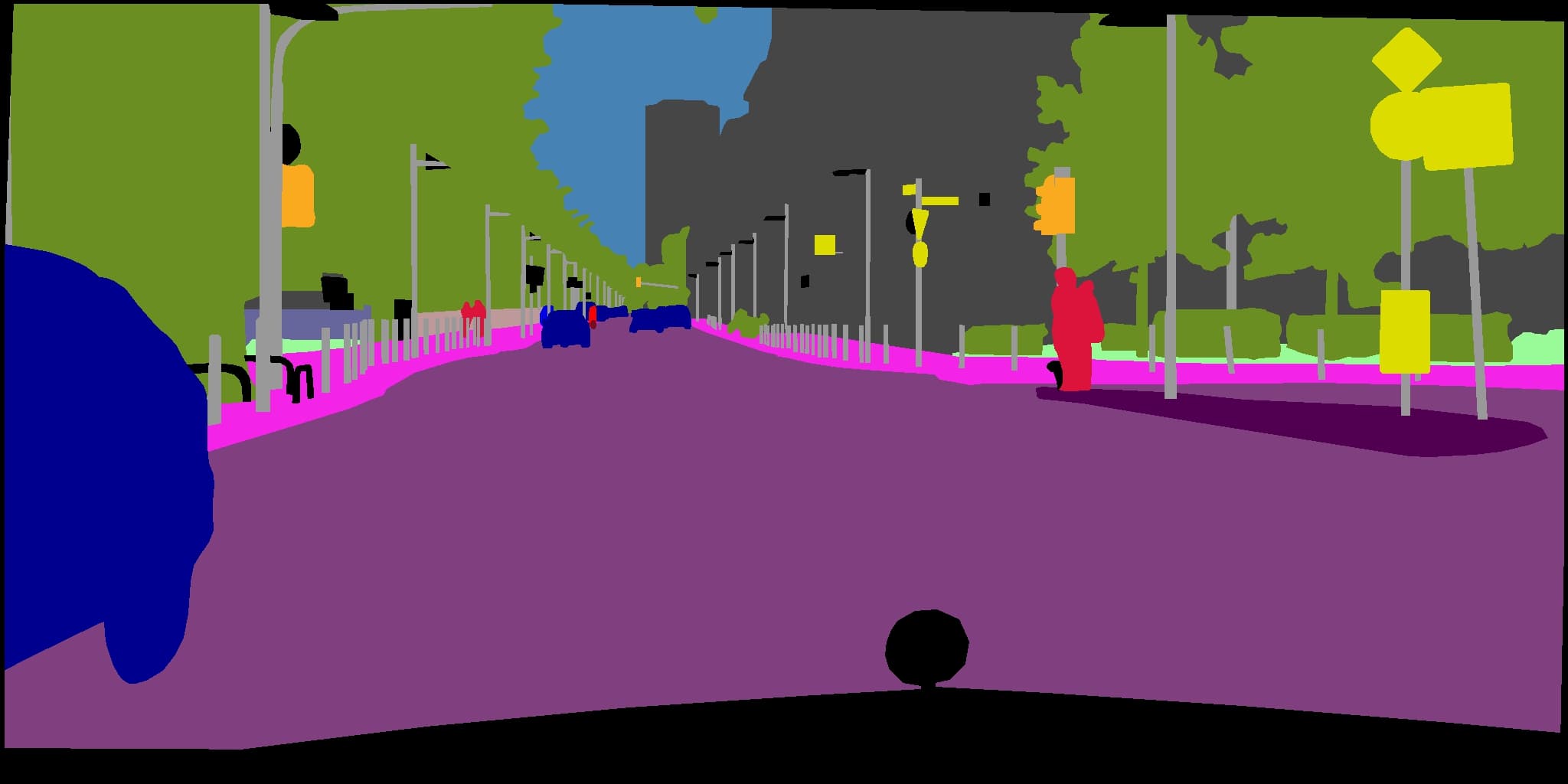}

          \end{subfigure}\\
          \begin{subfigure}[b]{0.19\linewidth}
    \includegraphics[width=\linewidth]{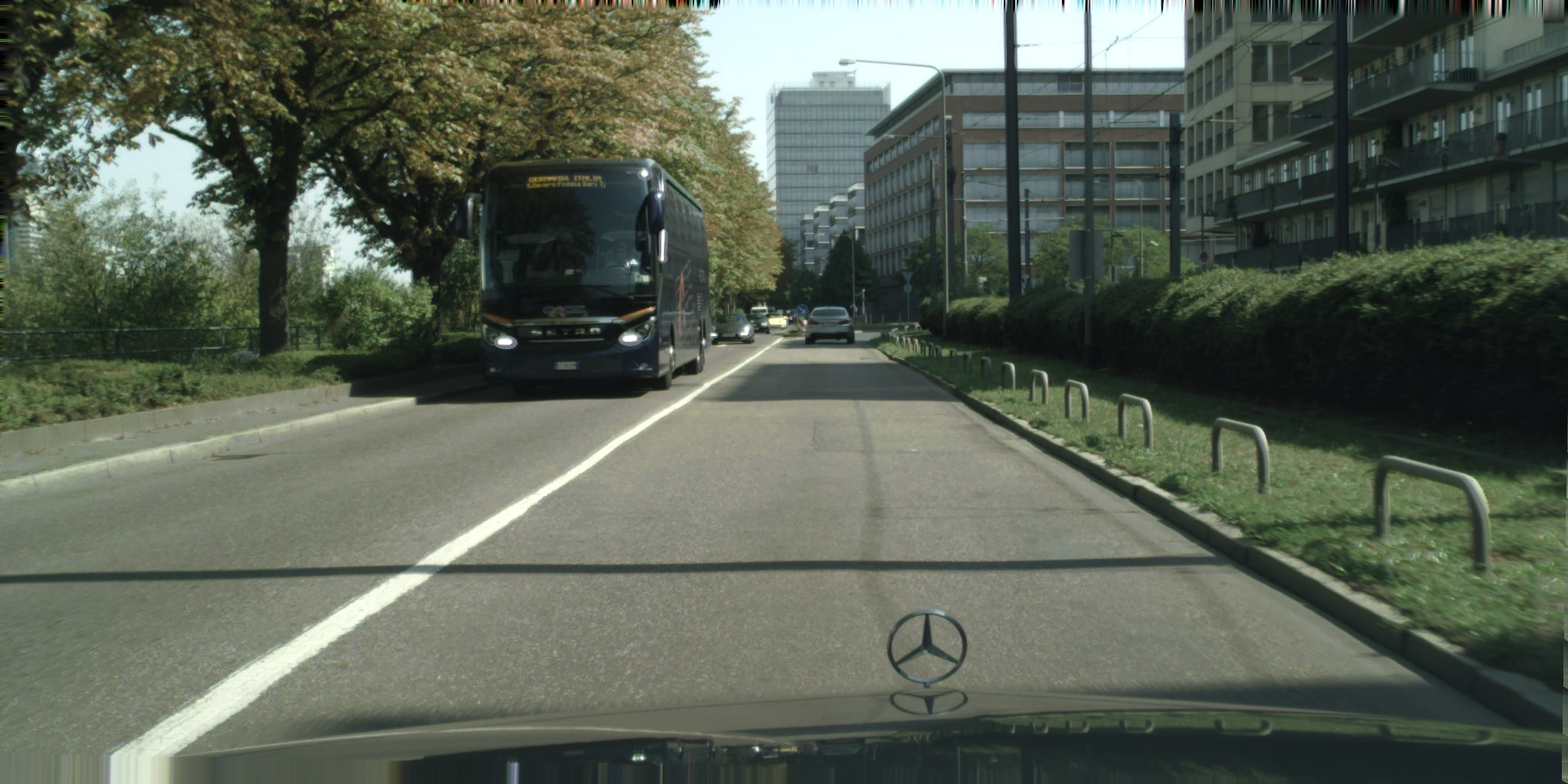}
    \subcaption{RGB image}
    \label{fig:supp_rgb}
  \end{subfigure}
          \hfill
          \begin{subfigure}[b]{0.19\linewidth}
            \includegraphics[width=\linewidth]{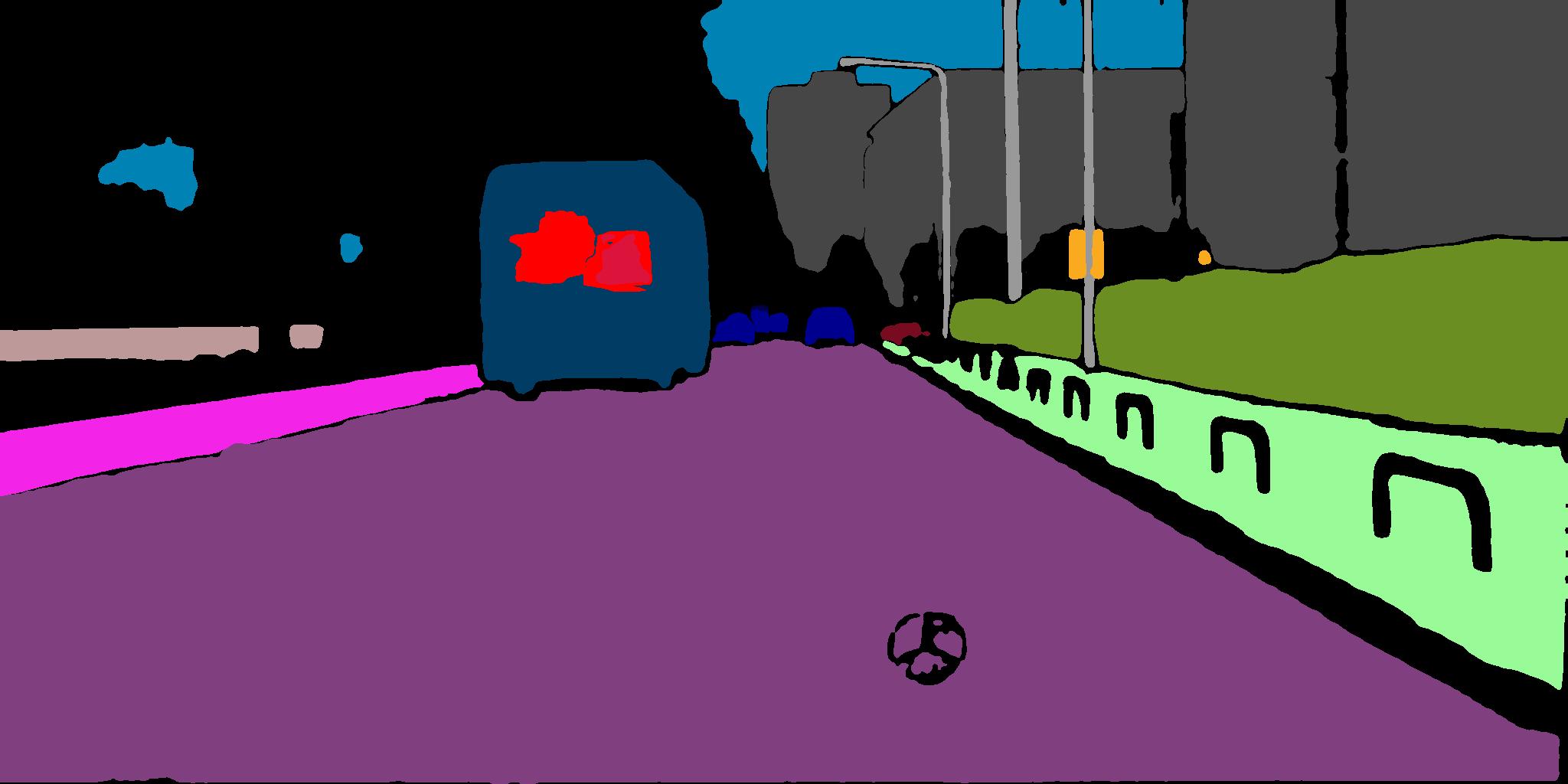}
            \subcaption{GroundingSAM~\cite{liu2023grounding,kirillov2023segment}}
          \end{subfigure}
          \hfill
          \begin{subfigure}[b]{0.19\linewidth}
            \includegraphics[width=\linewidth]{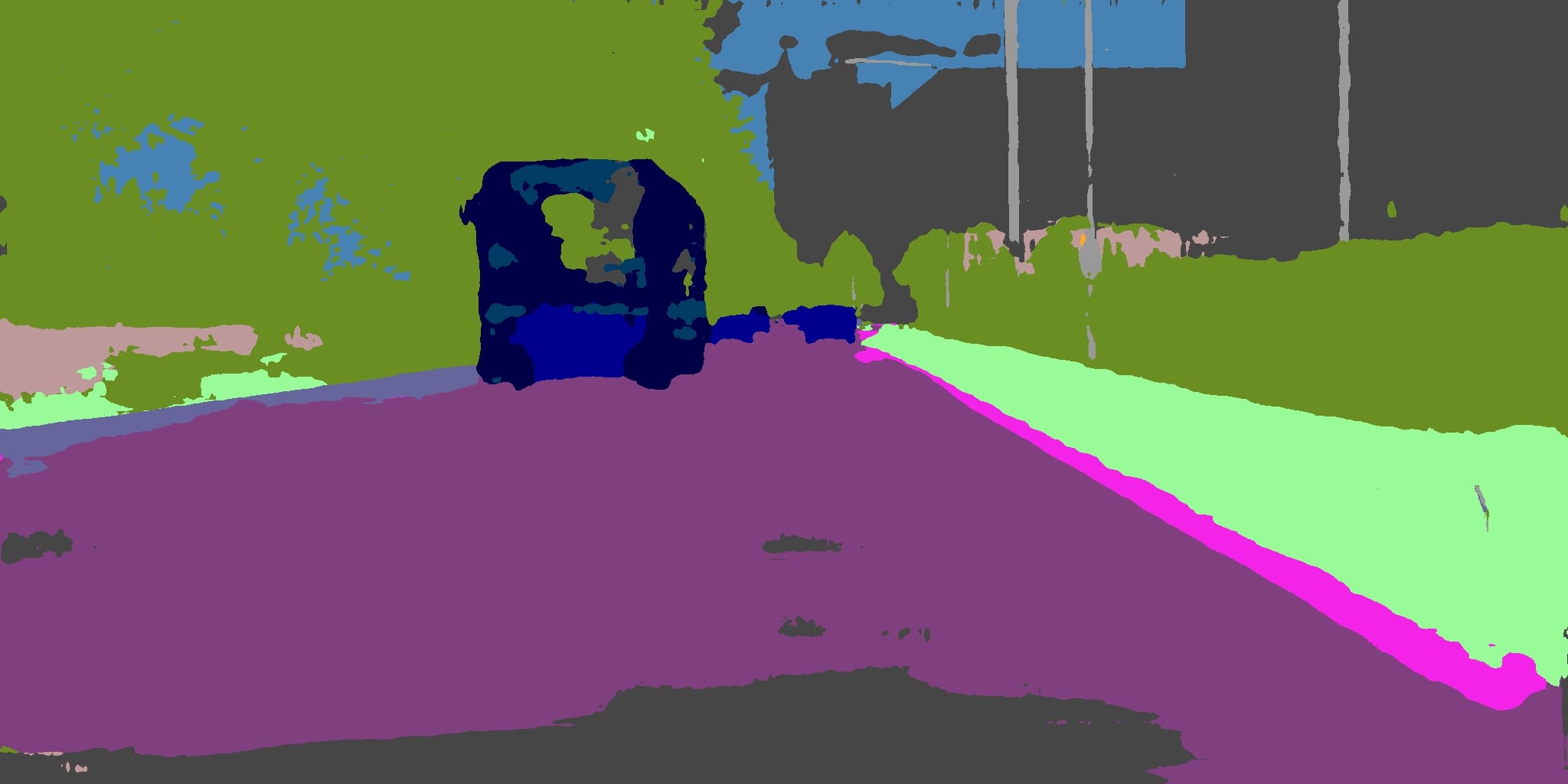}
            \subcaption{SHADE~\cite{zhao2022style}}
          \end{subfigure}
          \hfill
          \begin{subfigure}[b]{0.19\linewidth}
            \includegraphics[width=\linewidth]{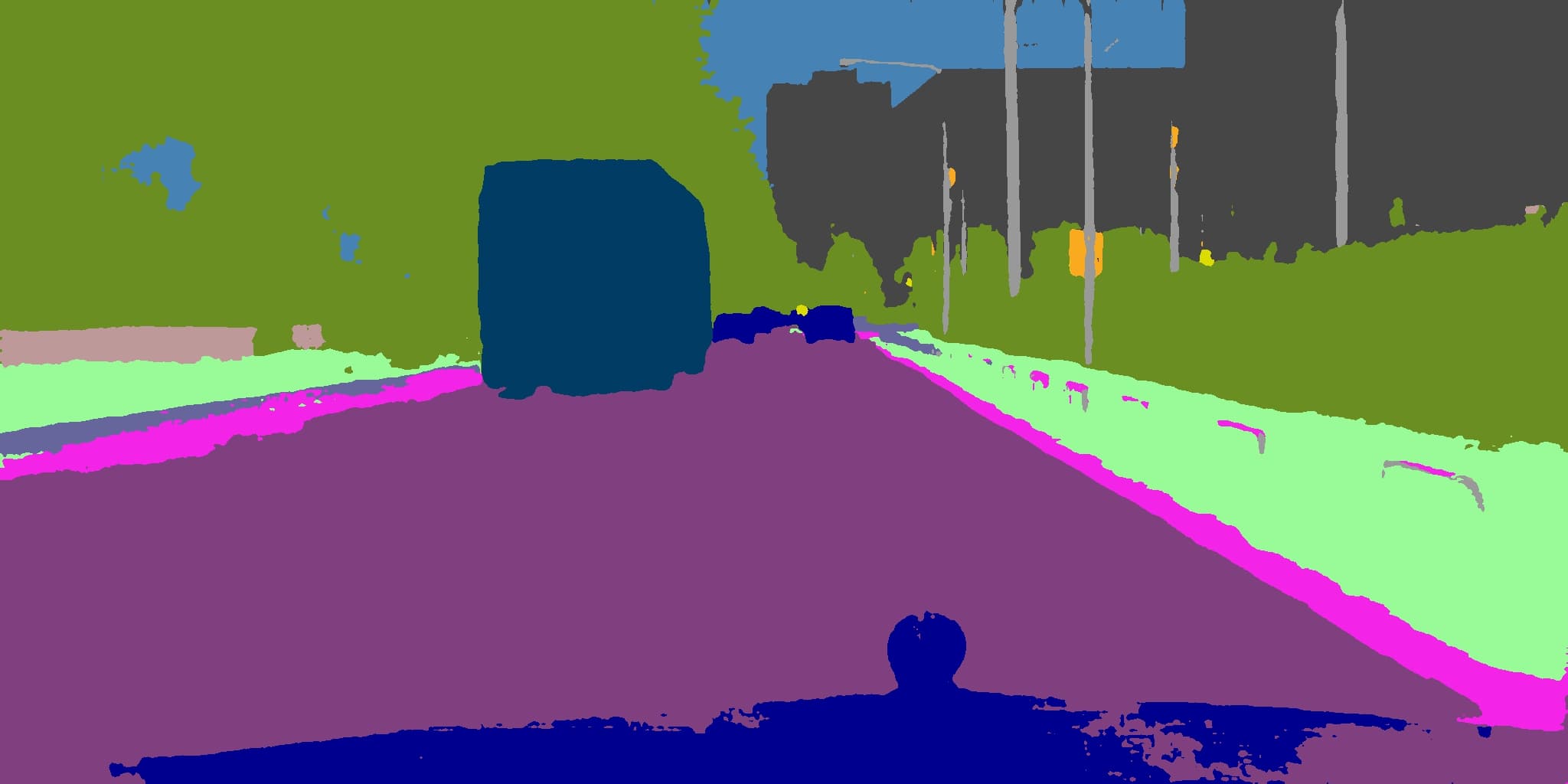}
            \subcaption{\textbf{\method}}
          \end{subfigure}
          \hfill
          \begin{subfigure}[b]{0.19\linewidth}
            \includegraphics[width=\linewidth]{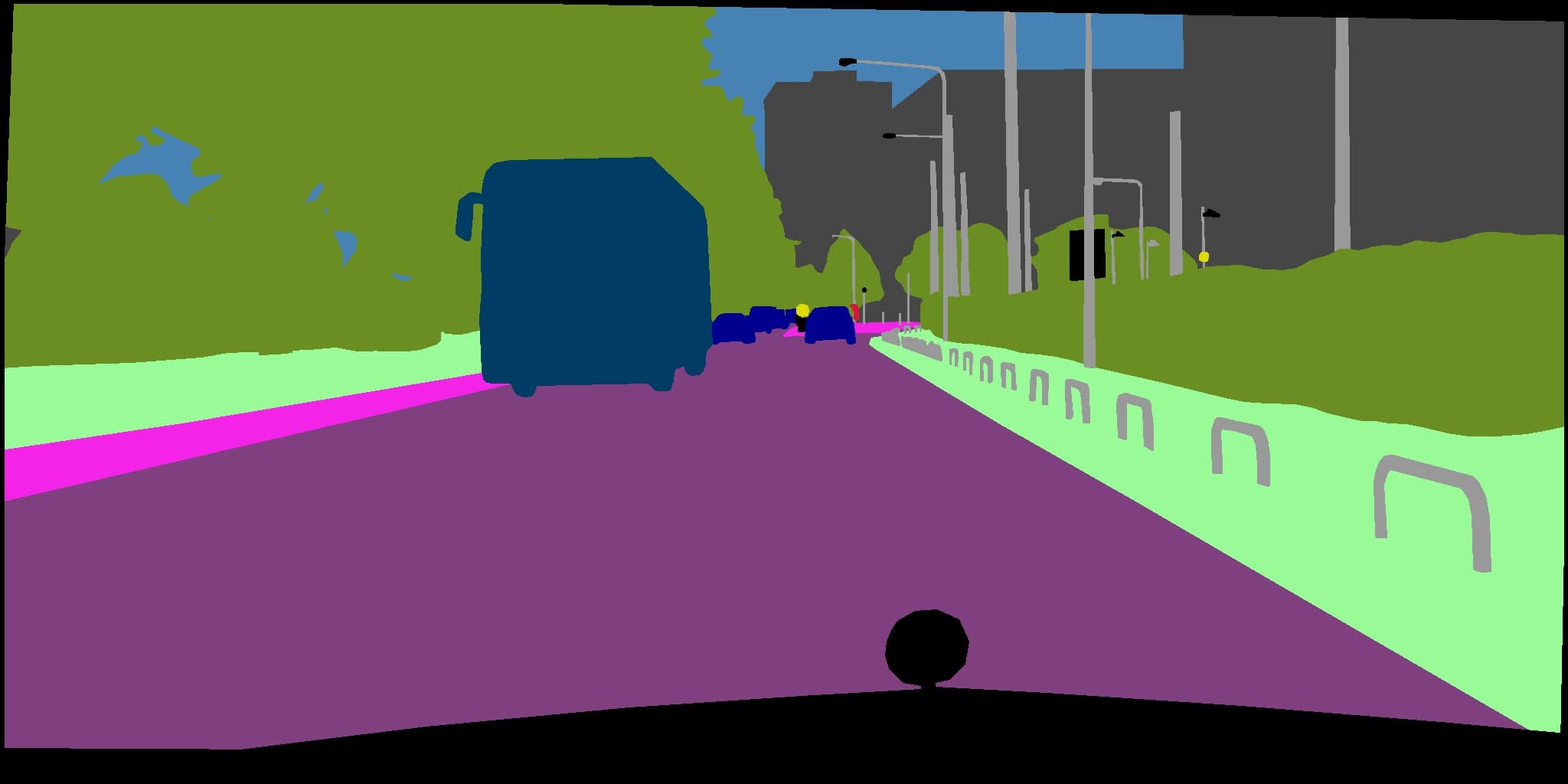}
            \subcaption{GT}

          \end{subfigure}\\

    \caption{Qualitative study on the GTA $\rightarrow$ \{C, B, M\} scenario using ConvNext-L architecture. For each RGB image (a), we show (b) ~\cite{liu2023grounding,kirillov2023segment} the segmentation map predicted by GroundingSAM~\cite{liu2023grounding,kirillov2023segment}, (c) SHADE~\cite{zhao2022style}, (d) \method and (e) the Groundtruth associated. }
    \label{fig:quali_pred}
\end{figure*}

The qualitative analysis in Figure \ref{fig:quali_pred} clearly shows the differences between \method and other approaches like GroundingSAM and SHADE. GroundingSAM struggles with fully segmenting all the regions due to its reliance on text prompts, which becomes challenging when objects are difficult to describe textually. Conversely, SHADE lacks robustness in distinguishing classes with similar features, such as "road" and "sidewalk", resulting in the creation of noisy pseudo labels.

\section{Impact of each component}
\label{sec:components-quali}
\begin{figure*}[ht]
    \centering
          \begin{subfigure}[b]{0.19\linewidth}
            \includegraphics[width=\linewidth]{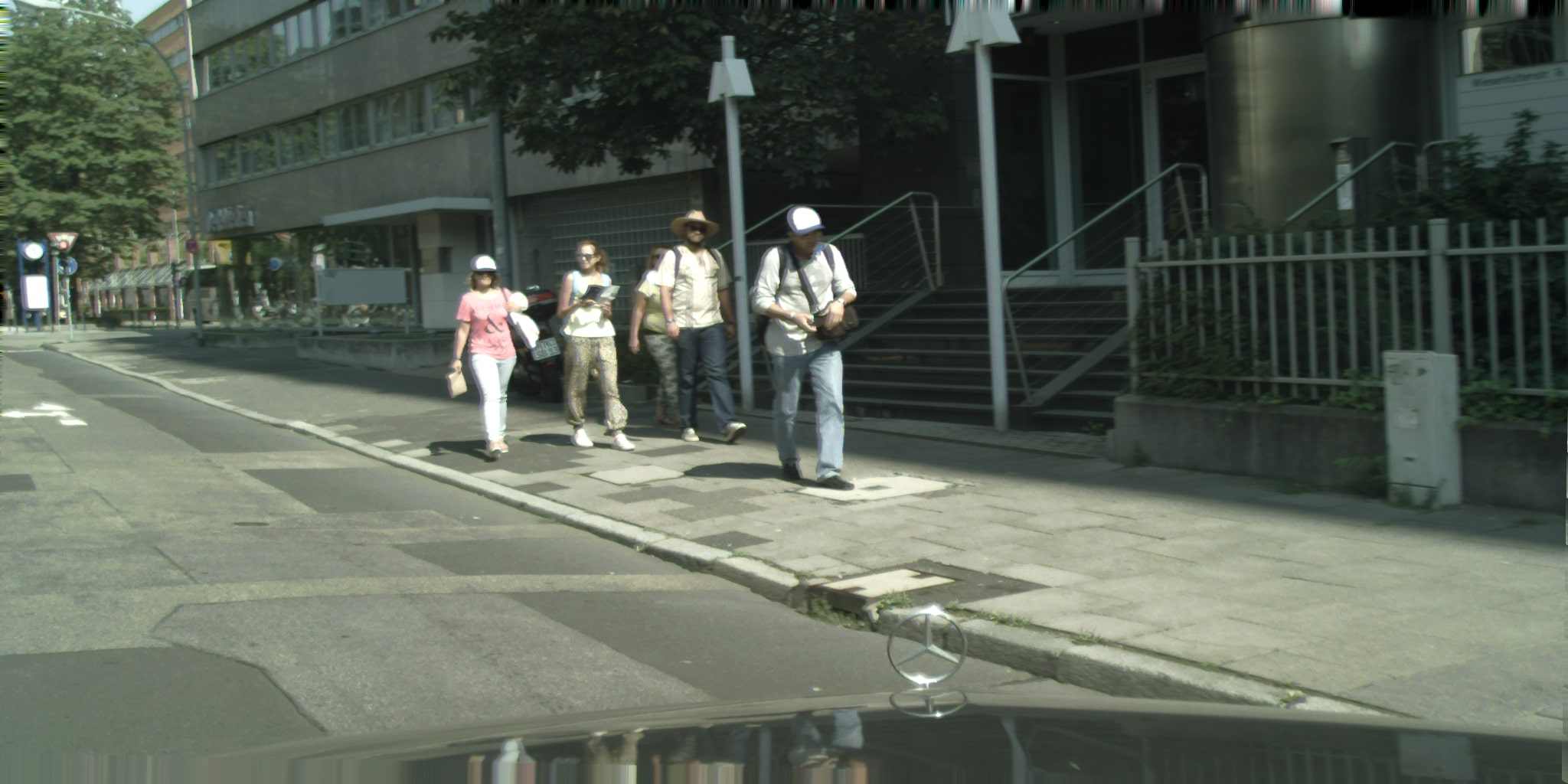}
          \end{subfigure}
          \hfill
          \begin{subfigure}[b]{0.19\linewidth}
            \includegraphics[width=\linewidth]{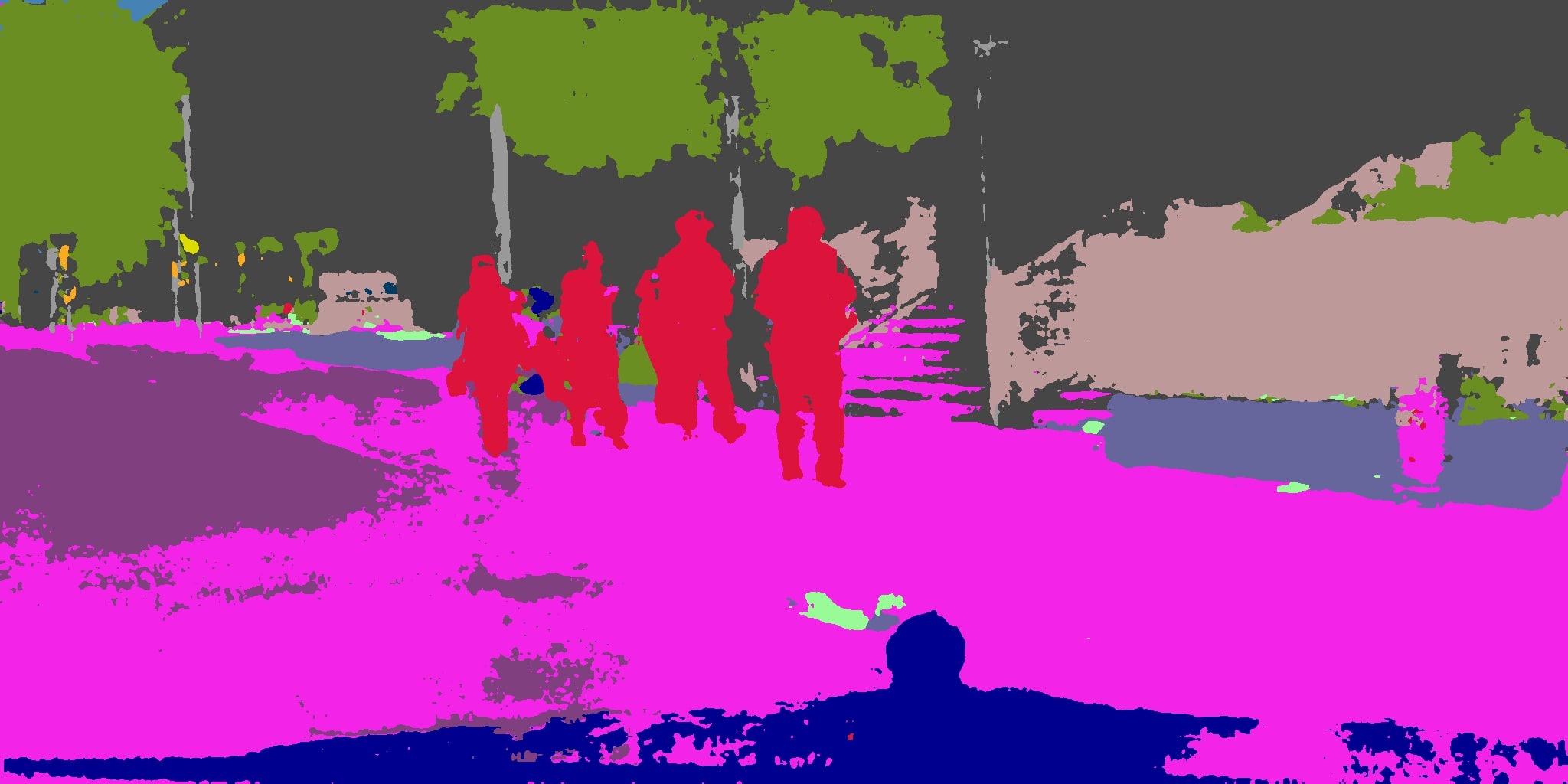}
          \end{subfigure}
          \hfill
          \begin{subfigure}[b]{0.19\linewidth}
            \includegraphics[width=\linewidth]{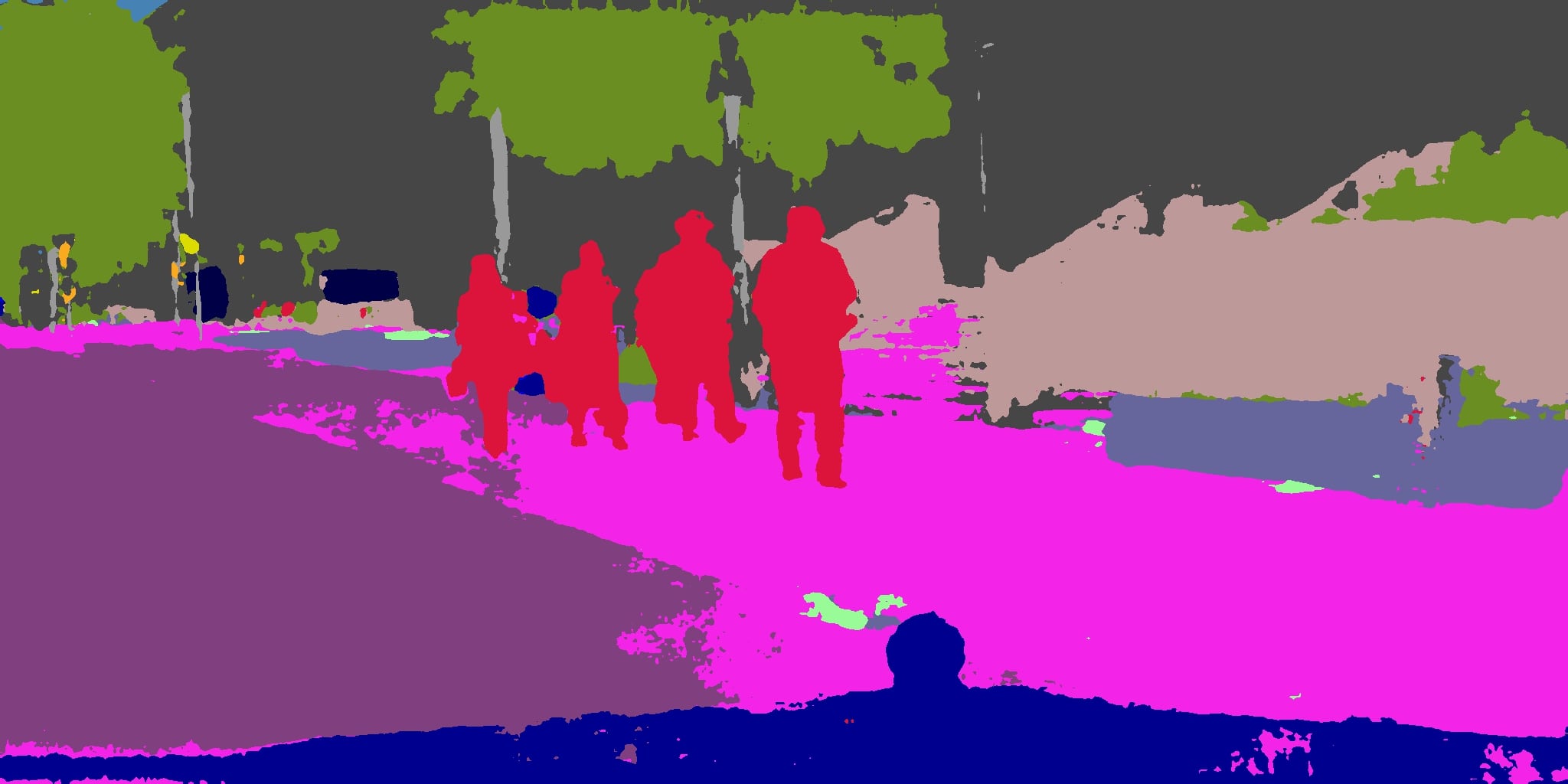}
          \end{subfigure}
          \hfill
          \begin{subfigure}[b]{0.19\linewidth}
            \includegraphics[width=\linewidth]{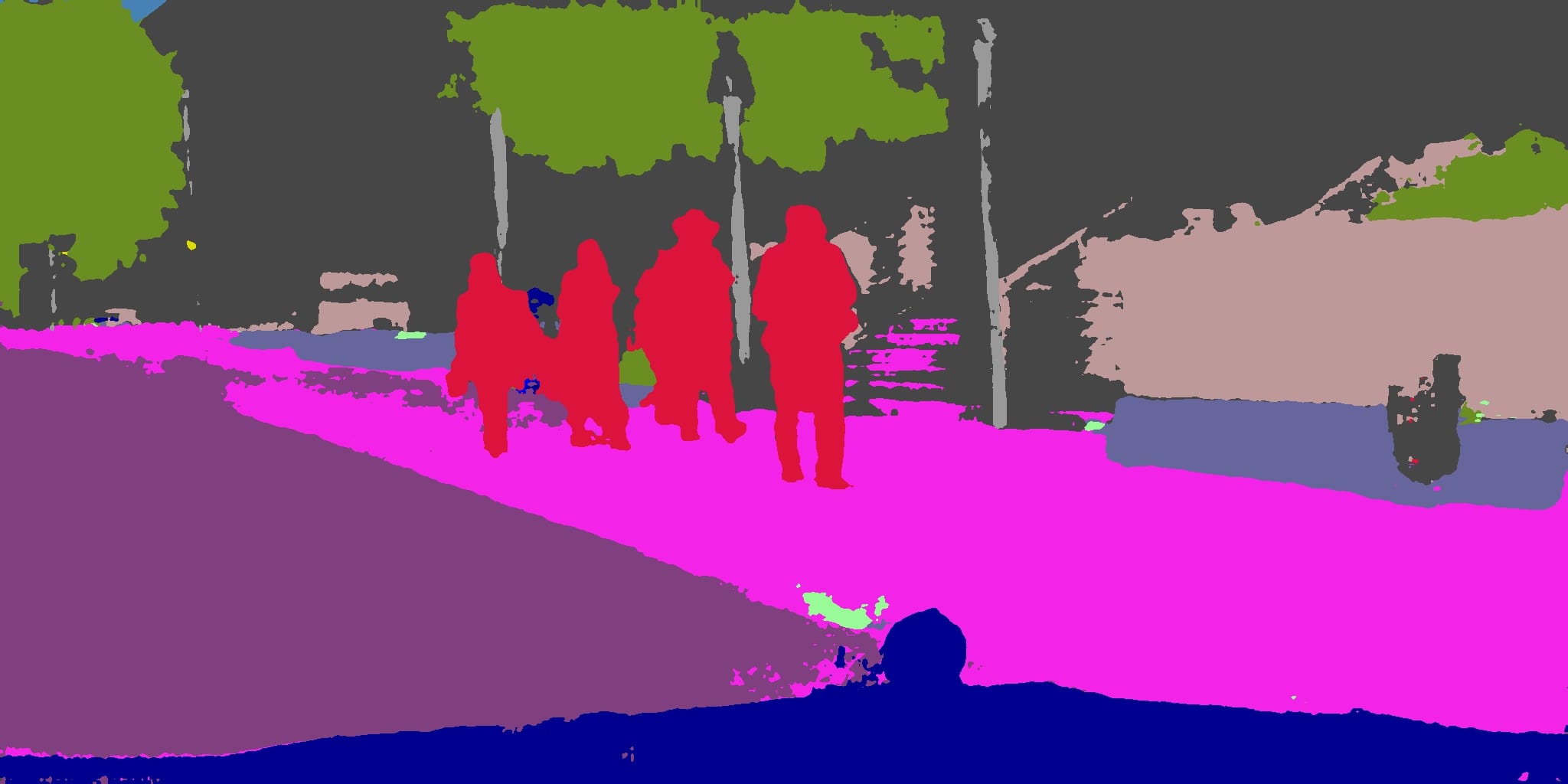}
          \end{subfigure}
          \hfill
          \begin{subfigure}[b]{0.19\linewidth}
            \includegraphics[width=\linewidth]{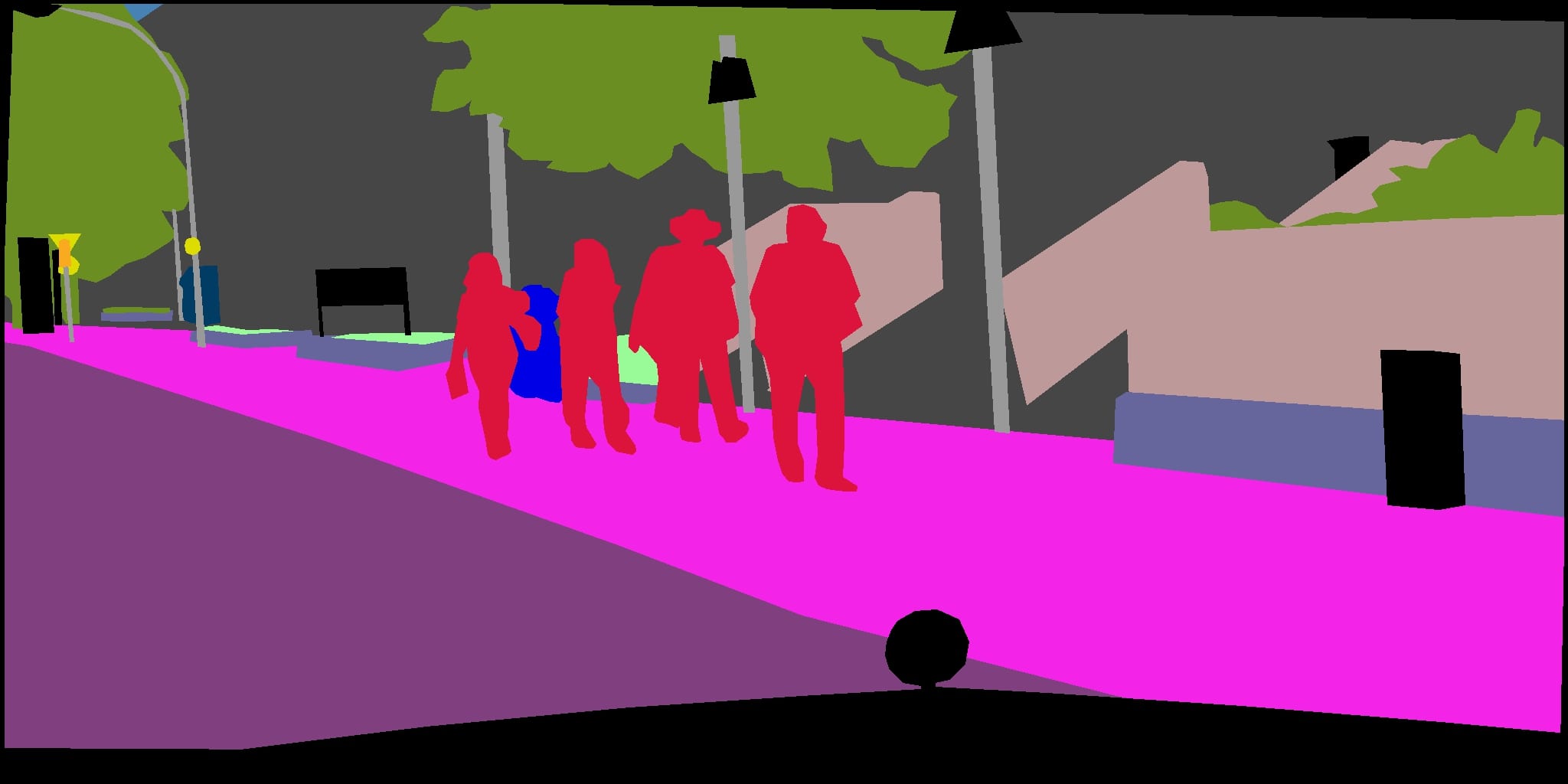}
          \end{subfigure} \\
          \begin{subfigure}[b]{0.19\linewidth}
            \includegraphics[width=\linewidth]{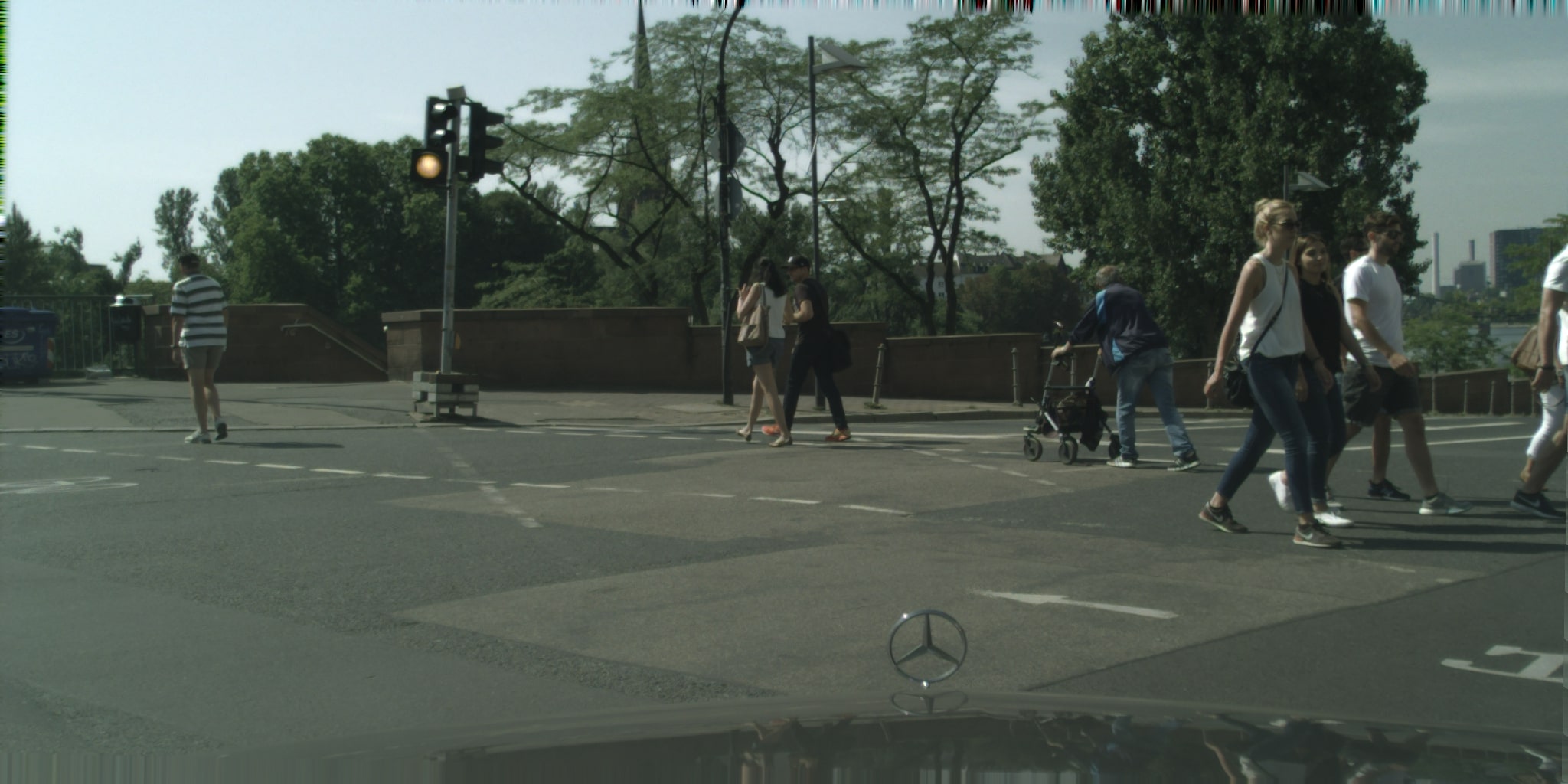}
          \end{subfigure}
          \hfill
          \begin{subfigure}[b]{0.19\linewidth}
            \includegraphics[width=\linewidth]{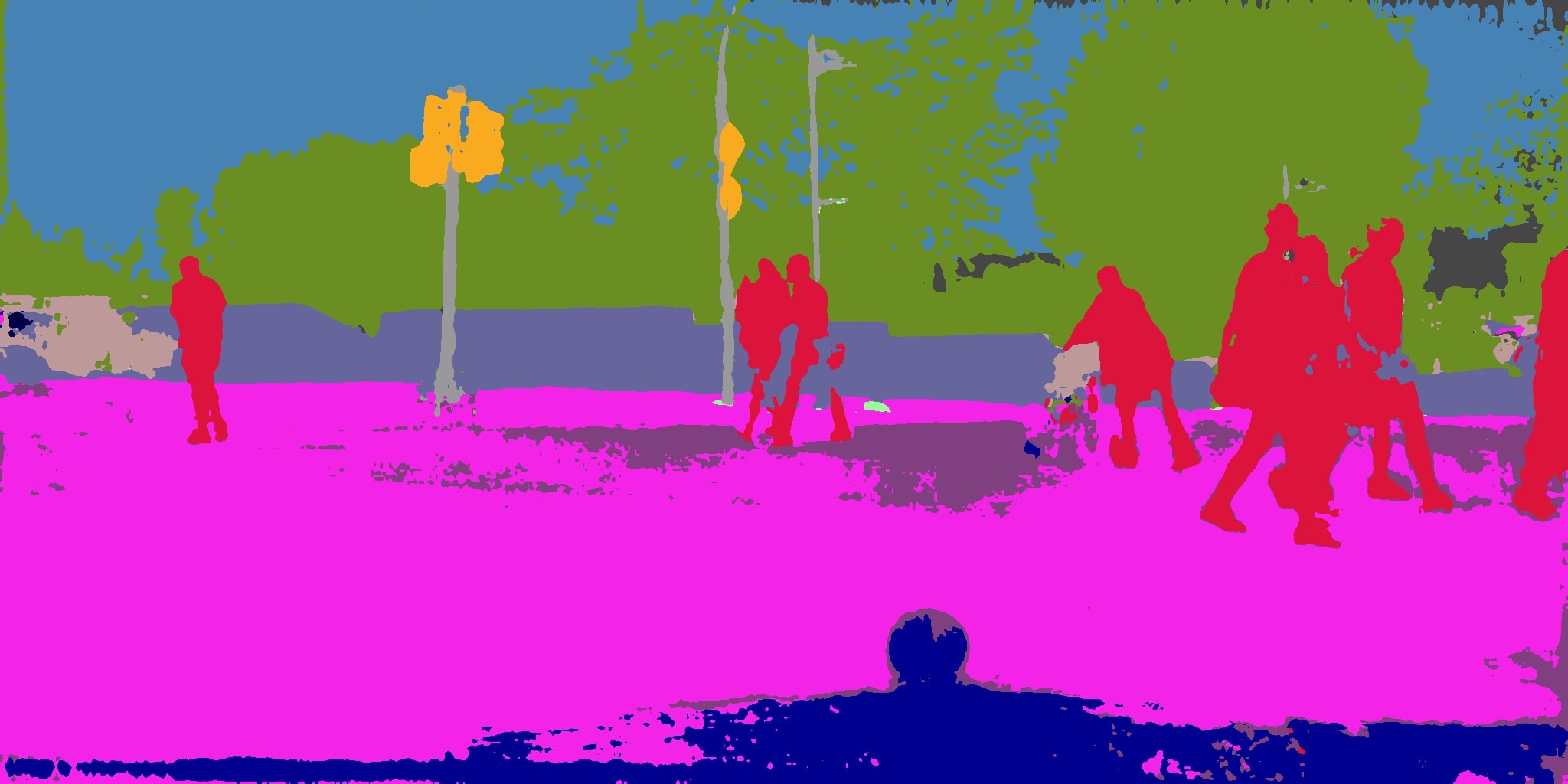}
          \end{subfigure}
          \hfill
          \begin{subfigure}[b]{0.19\linewidth}
            \includegraphics[width=\linewidth]{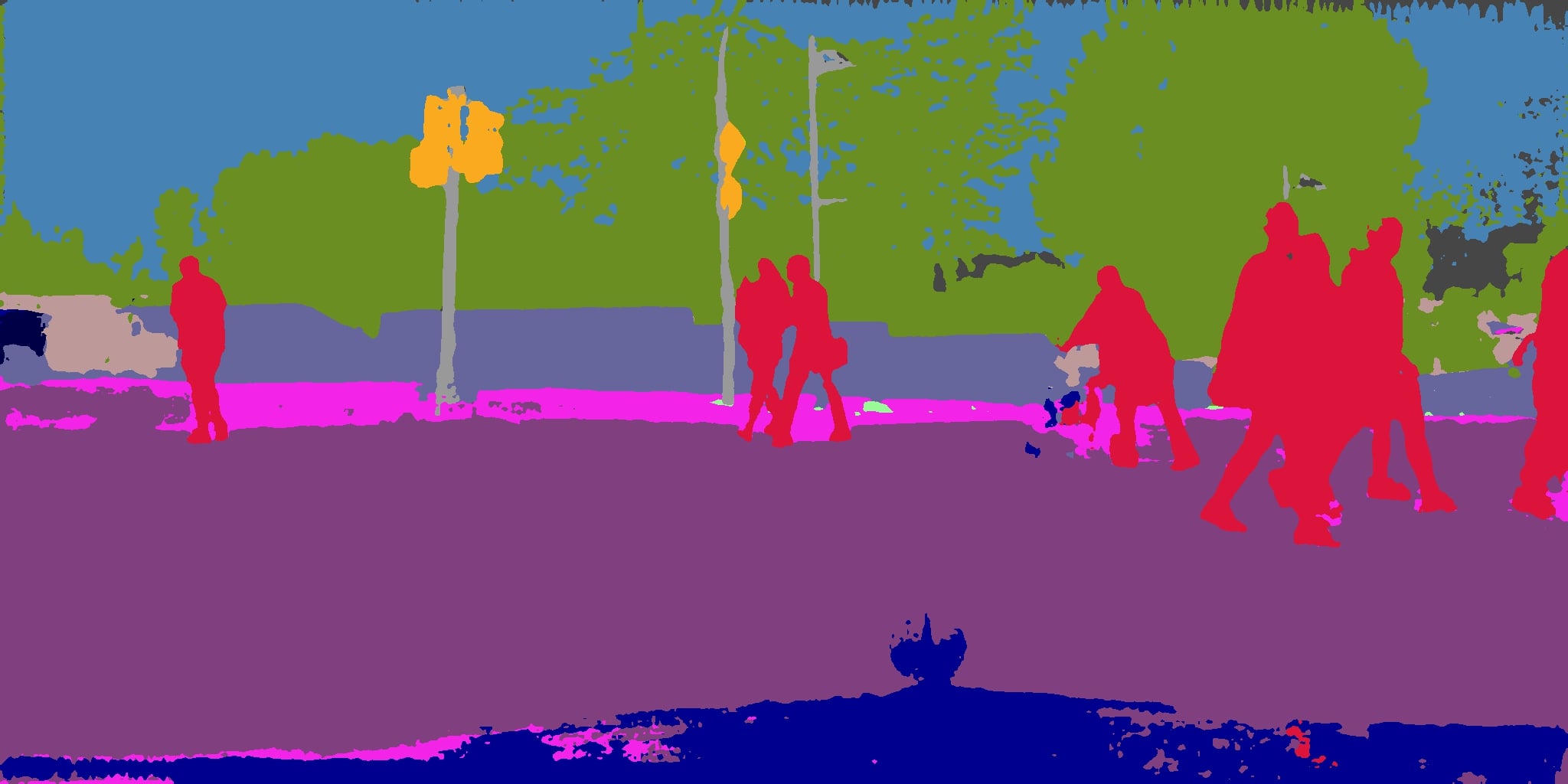}
          \end{subfigure}
          \hfill
          \begin{subfigure}[b]{0.19\linewidth}
            \includegraphics[width=\linewidth]{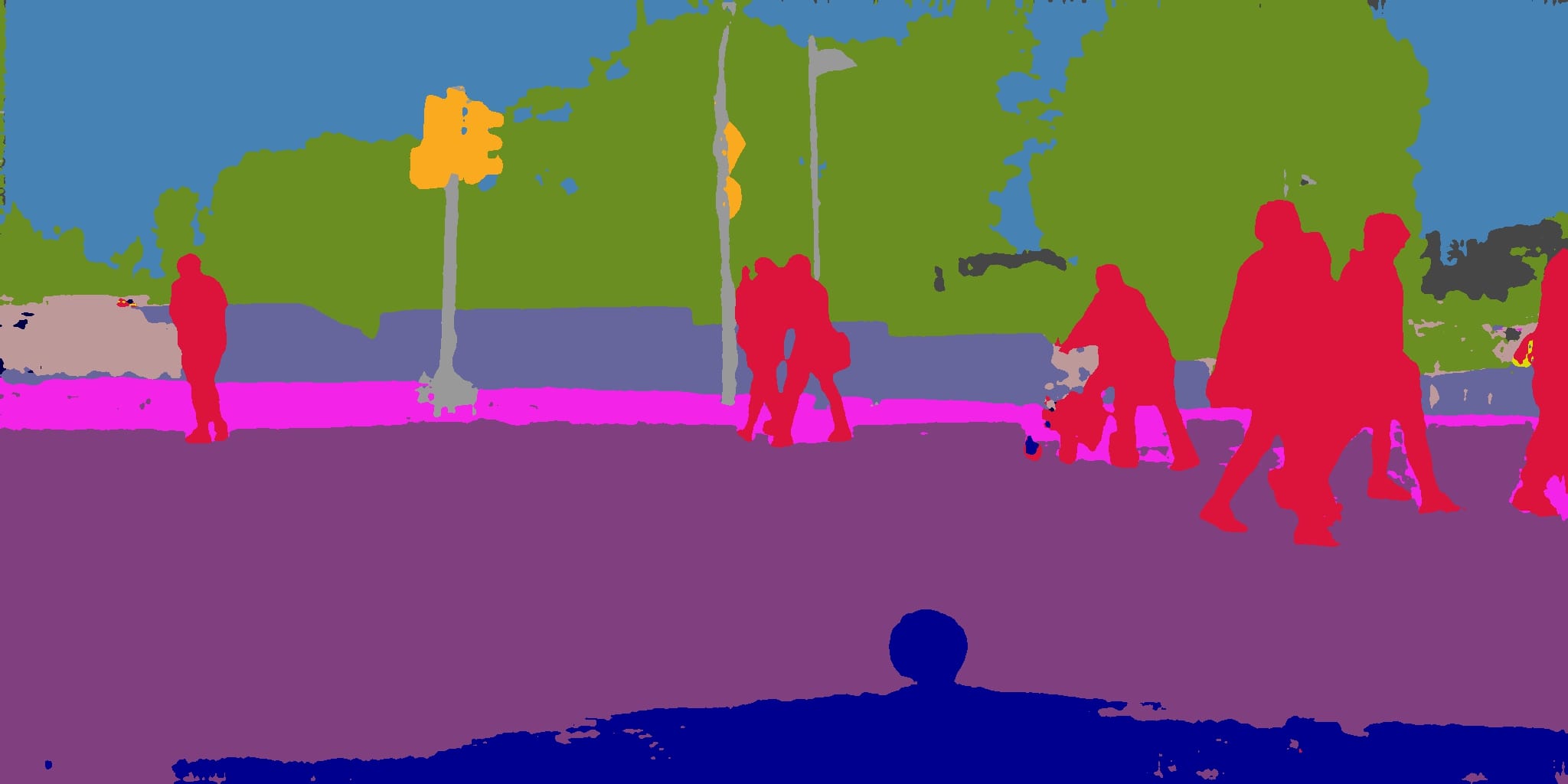}
          \end{subfigure}
          \hfill
          \begin{subfigure}[b]{0.19\linewidth}
            \includegraphics[width=\linewidth]{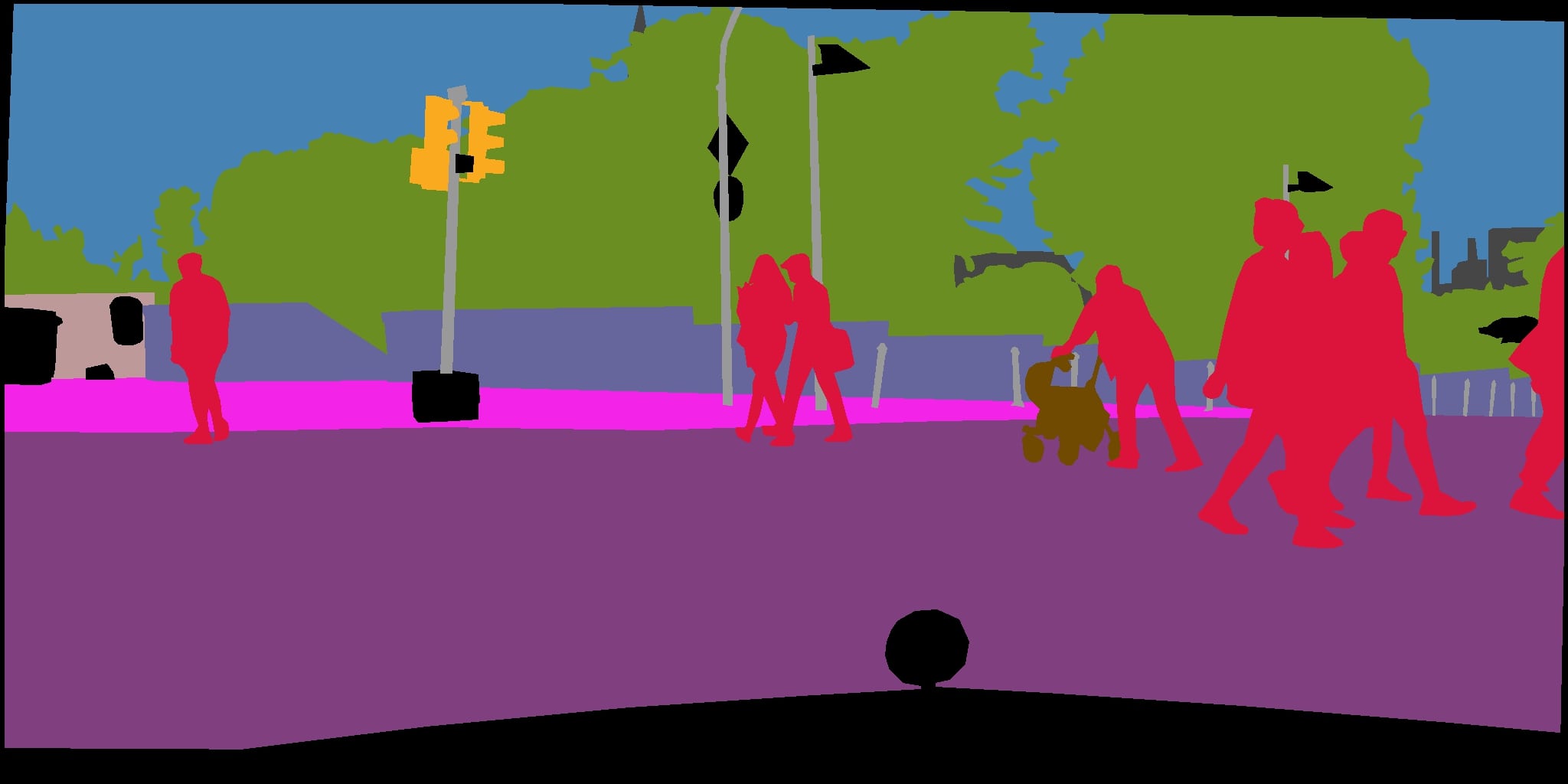}
          \end{subfigure} \\
          \begin{subfigure}[b]{0.19\linewidth}
            \includegraphics[width=\linewidth]{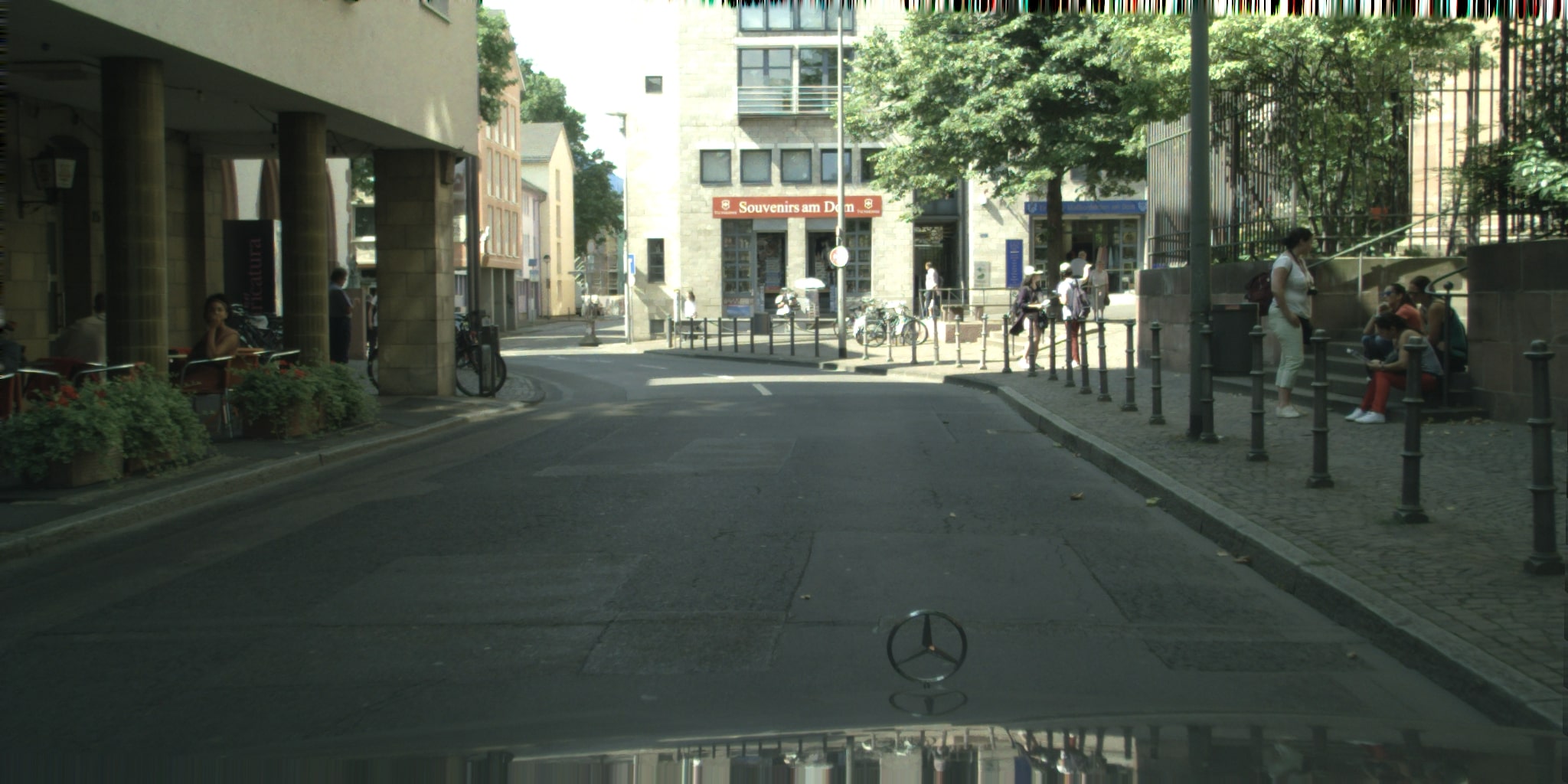}
          \end{subfigure}
          \hfill
          \begin{subfigure}[b]{0.19\linewidth}
            \includegraphics[width=\linewidth]{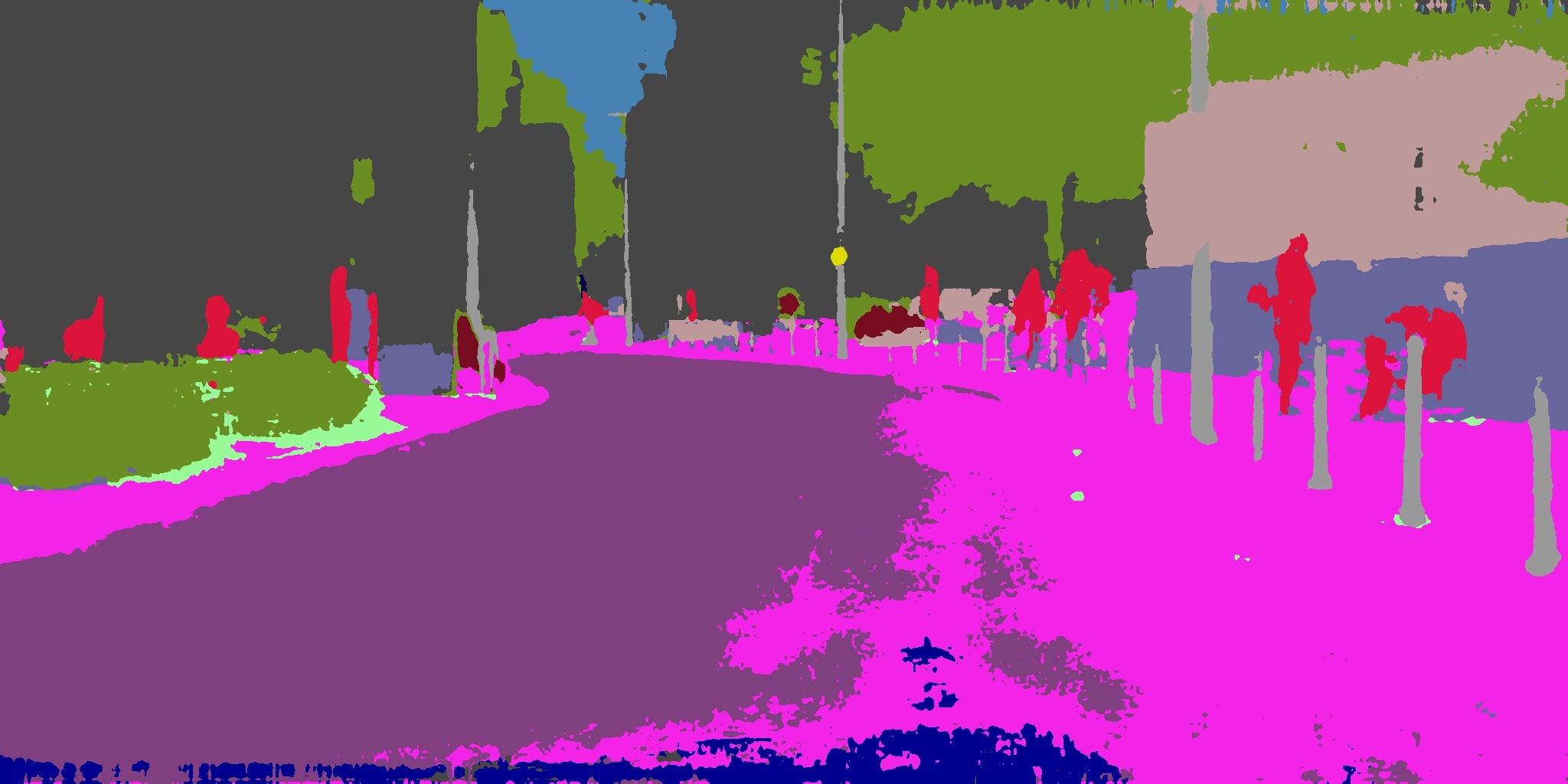}
          \end{subfigure}
          \hfill
          \begin{subfigure}[b]{0.19\linewidth}
            \includegraphics[width=\linewidth]{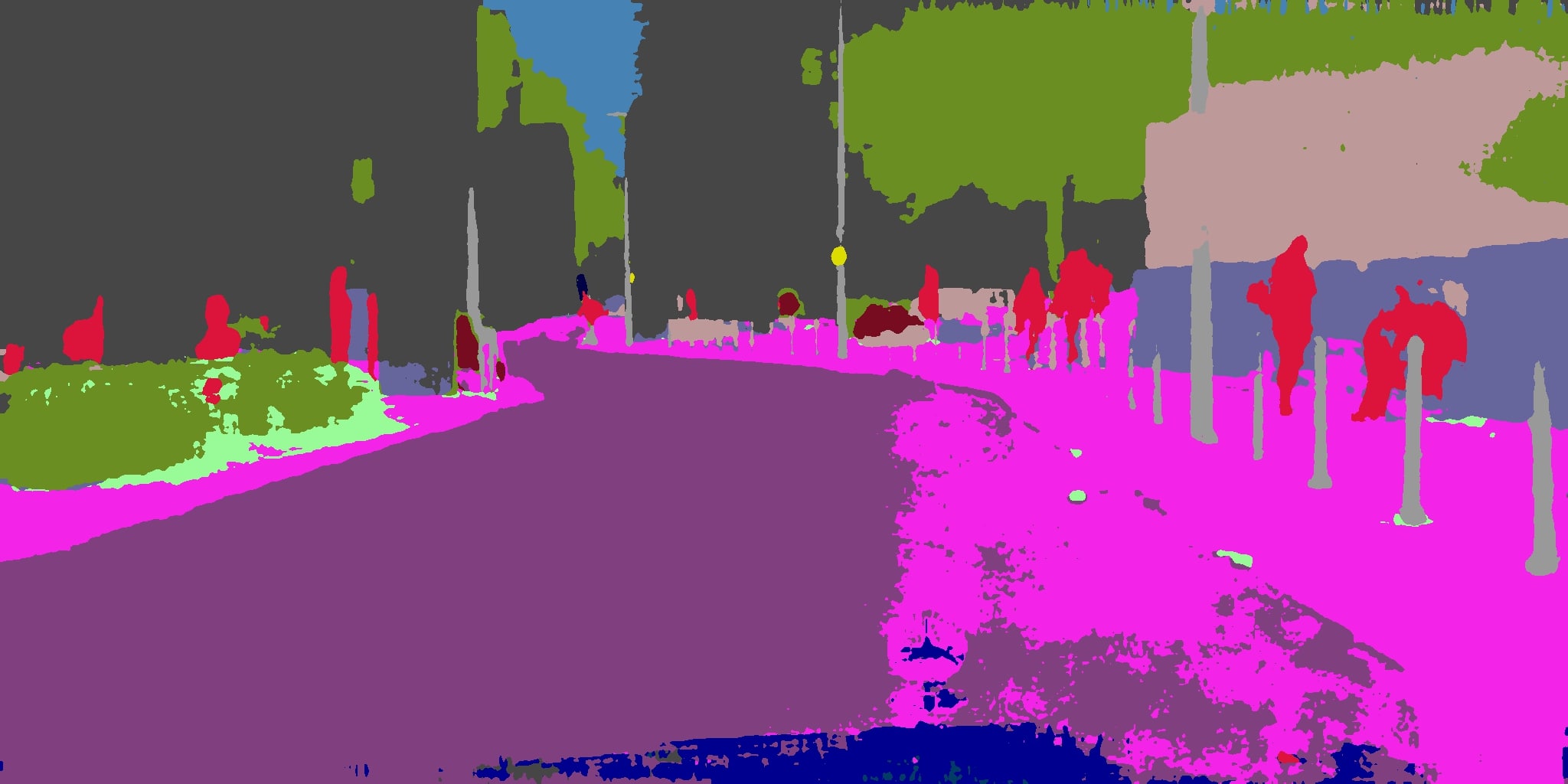}
          \end{subfigure}
          \hfill
          \begin{subfigure}[b]{0.19\linewidth}
            \includegraphics[width=\linewidth]{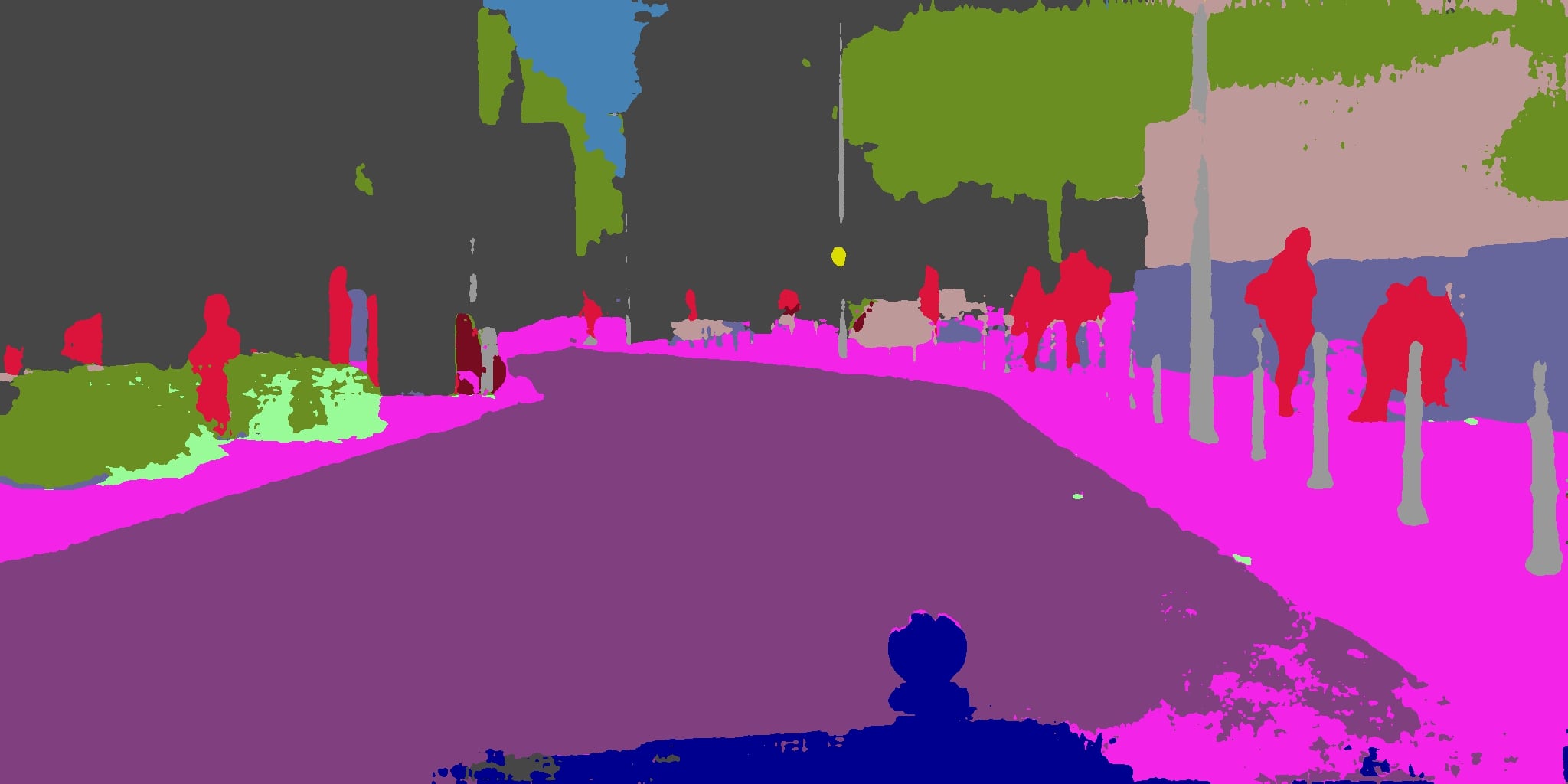}
          \end{subfigure}
          \hfill
          \begin{subfigure}[b]{0.19\linewidth}
            \includegraphics[width=\linewidth]{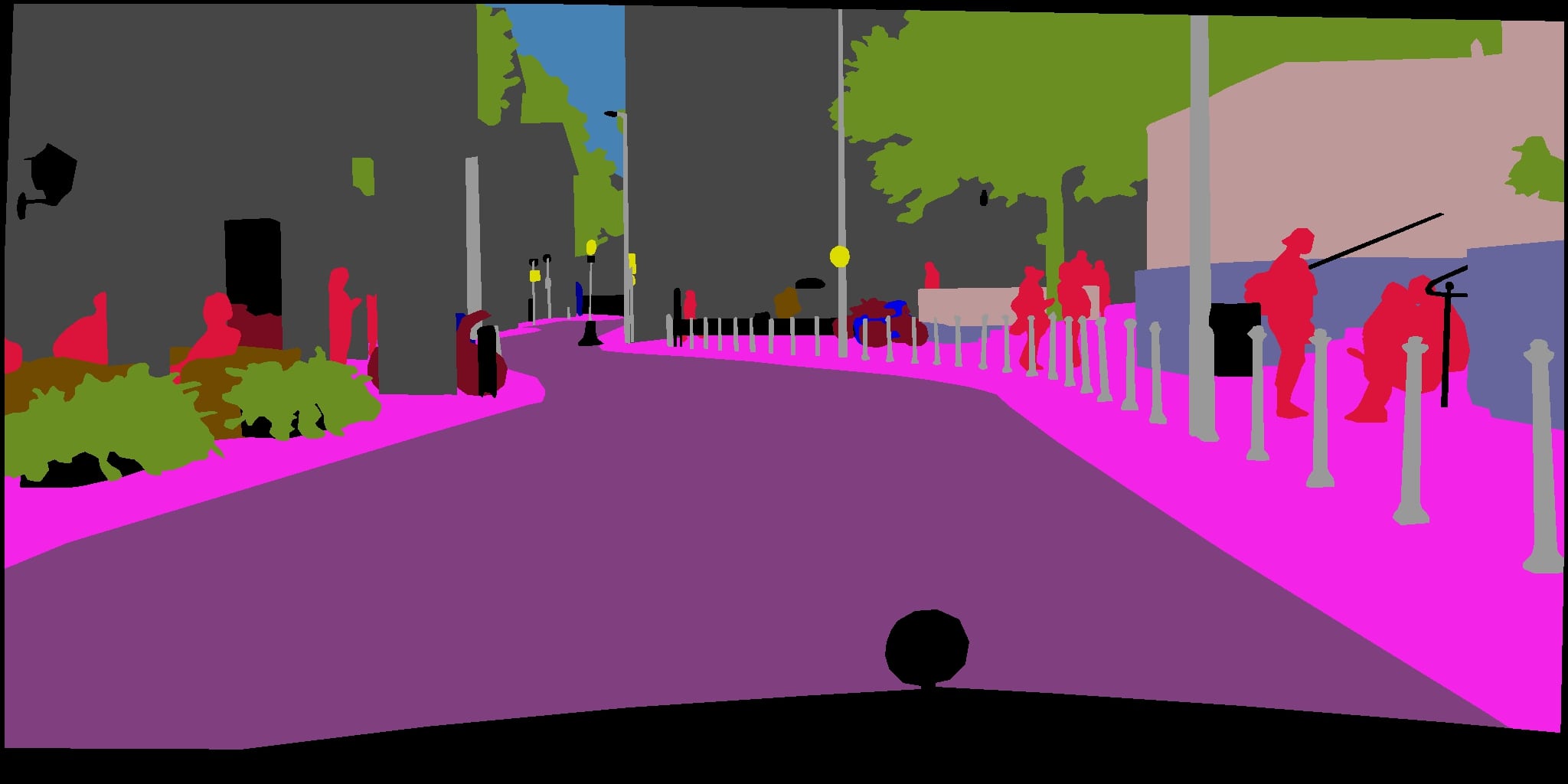}
          \end{subfigure} \\
          \begin{subfigure}[b]{0.19\linewidth}
            \includegraphics[width=\linewidth]{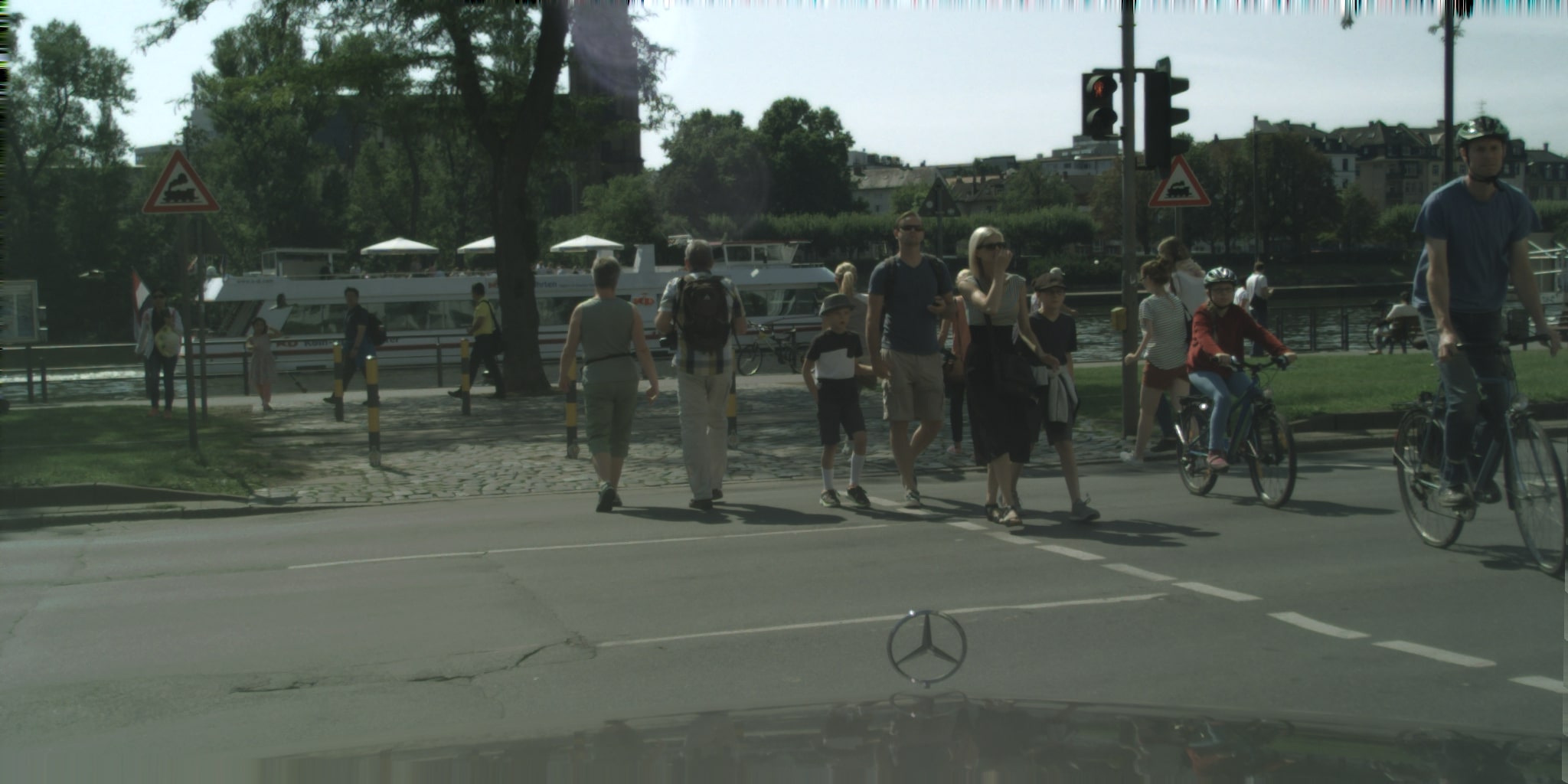}
            \subcaption{RGB image}
          \end{subfigure}
          \hfill
          \begin{subfigure}[b]{0.19\linewidth}
            \includegraphics[width=\linewidth]{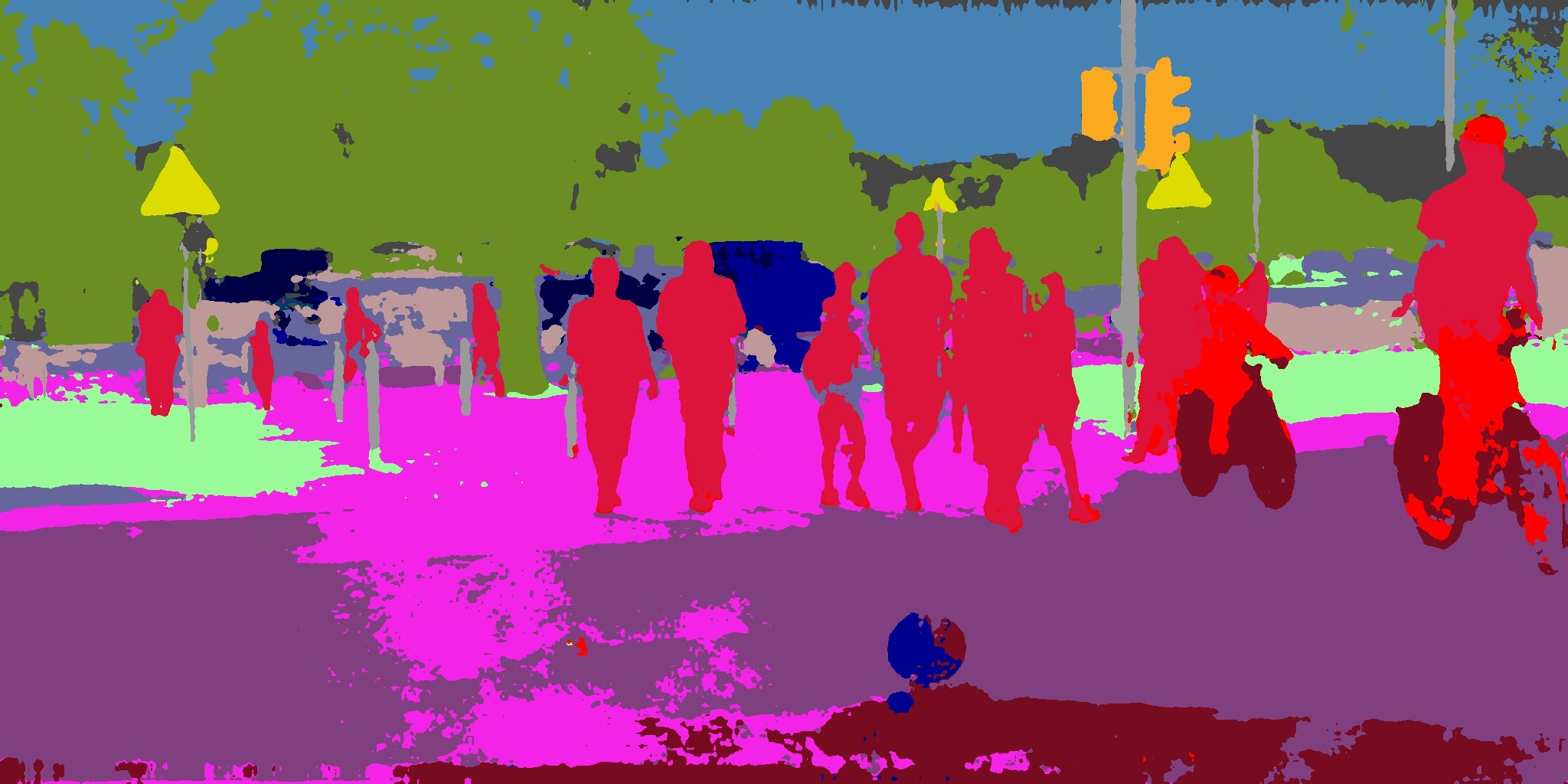}
            \subcaption{CLIP}
          \end{subfigure}
          \hfill
          \begin{subfigure}[b]{0.19\linewidth}
            \includegraphics[width=\linewidth]{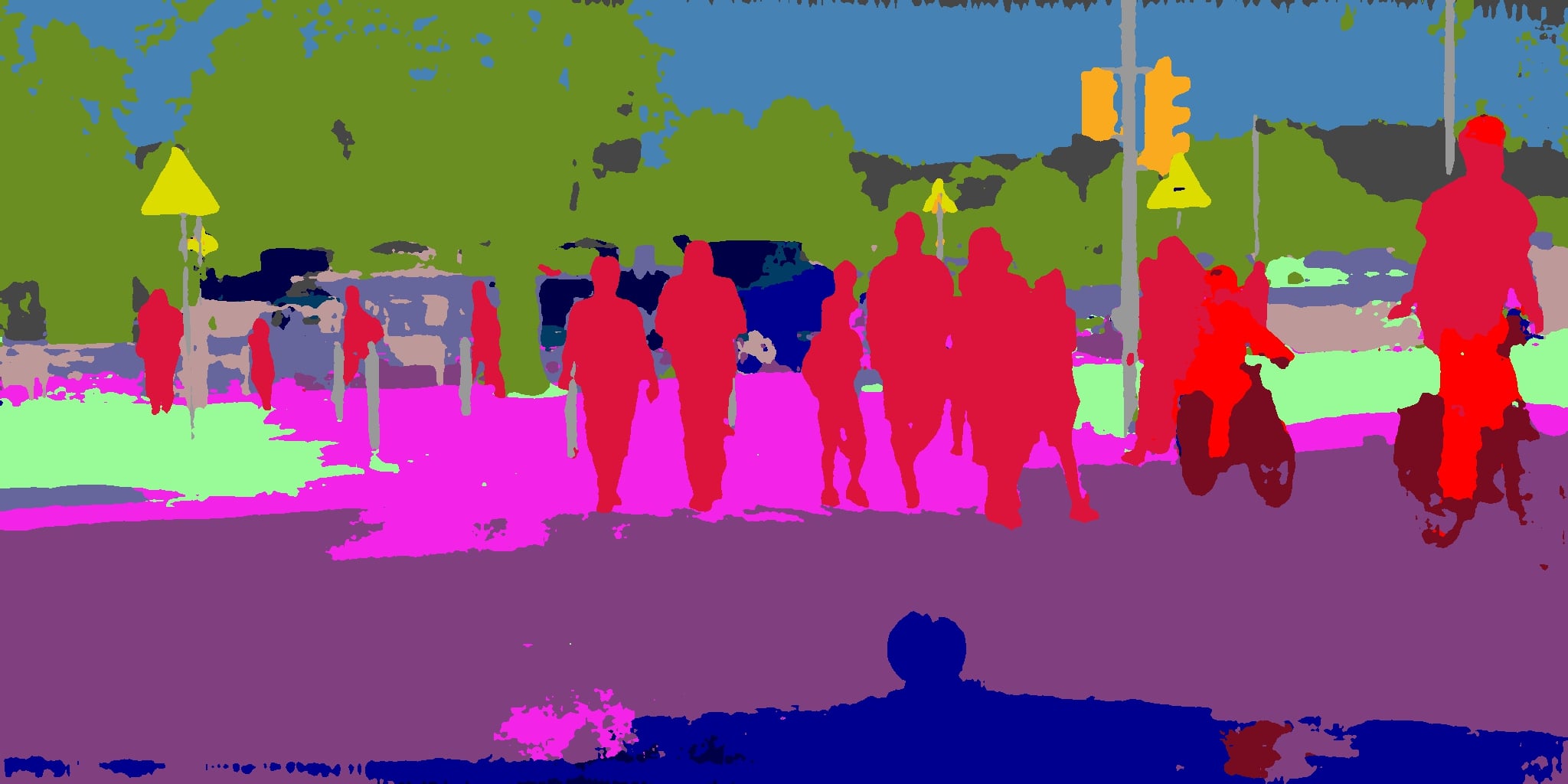}
            \subcaption{CLIP + \{L, D\}}
          \end{subfigure}
          \hfill
          \begin{subfigure}[b]{0.19\linewidth}
            \includegraphics[width=\linewidth]{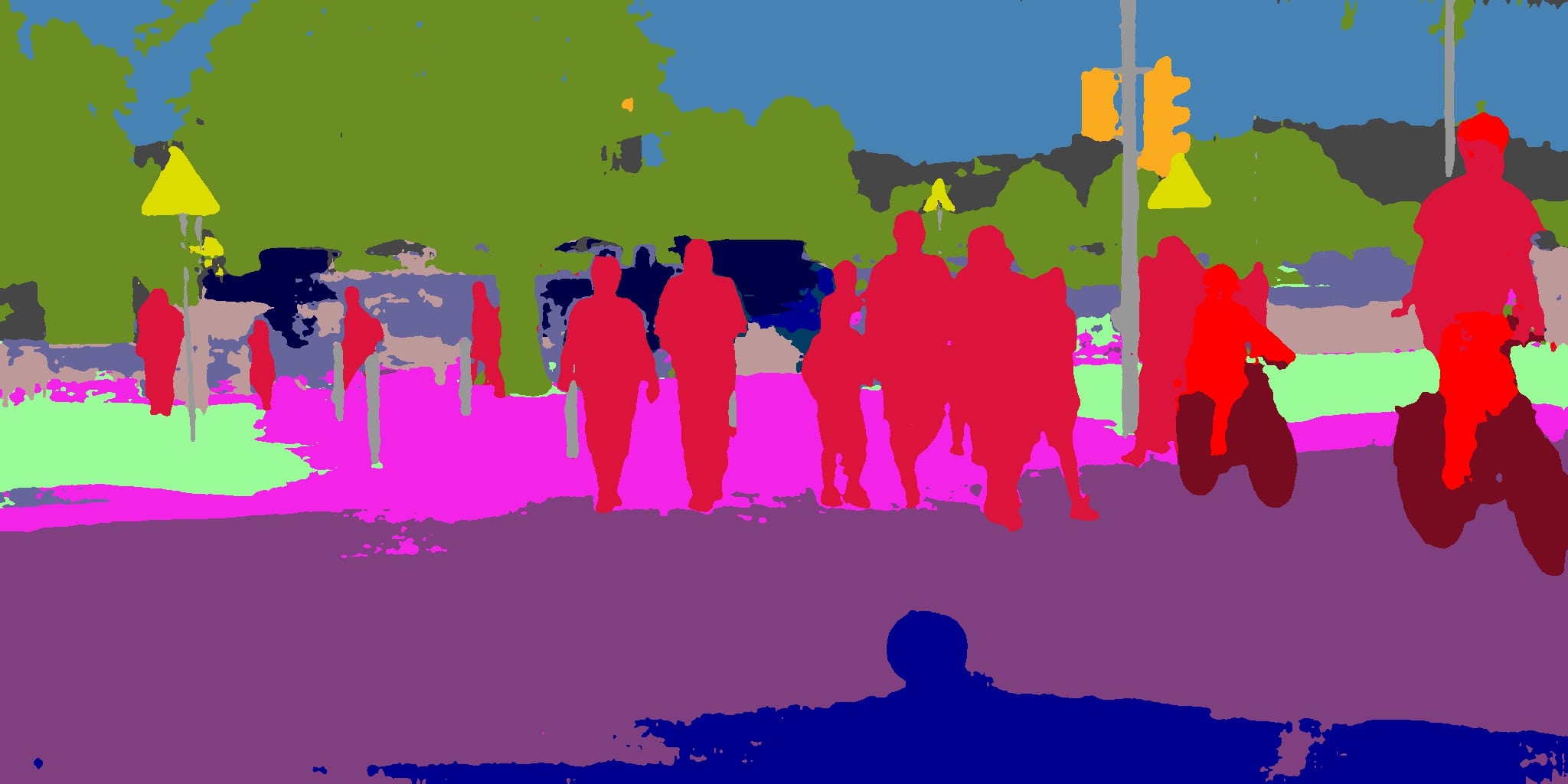}
            \subcaption{CLIP + \{L, D\} + SAM}
          \end{subfigure}
          \hfill
          \begin{subfigure}[b]{0.19\linewidth}
            \includegraphics[width=\linewidth]{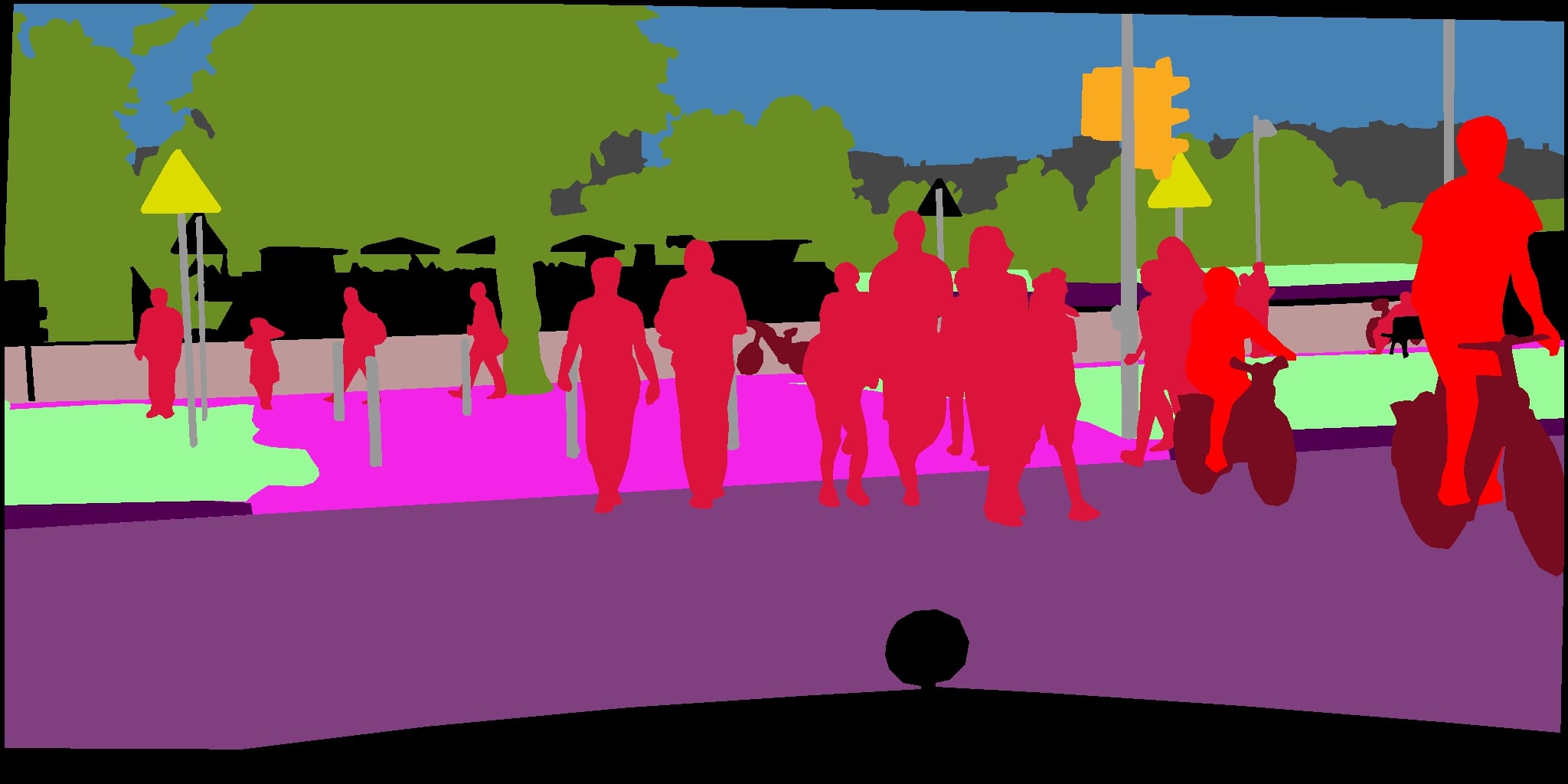}
            \subcaption{GT}
          \end{subfigure} \\
    \caption{Qualitative ablation study on the GTA $\rightarrow$ \{C, B, M\} scenario using ConvNext-L architecture. We demonstrate the impact of each foundation model in \method. For each RGB image (a), we show: (b) the segmentation maps predicted by the model using CLIP only as a feature extractor, (c) CLIP + \{LLM+Diffusion\} where self-training is done using the original pseudo labels, and (d) CLIP + \{LLM+Diffusion\} + SAM where the pseudo labels are refined using SAM}
    \label{fig:quali_ablation}
\end{figure*}

The qualitative analysis presented in Figure \ref{fig:quali_ablation} illustrates the progressive impact of integrating various foundation models into our system. The initial employment of CLIP~\cite{radford2021learning} as a sole feature extractor paired with Mask2Former~\cite{cheng2022masked} decoder yields a quantitative leap forward over previous DGSS methods, as shown in Table \ref{tab:mainablation}. However, this configuration, as visible in the second column of Figure \ref{fig:quali_ablation}, struggles to distinguish between similar classes. This issue is similarly observed in with SHADE~\cite{zhao2022style}, as shown in Figure \ref{fig:quali_pred}.
The incorporation of the \{LLM~\cite{touvron2023llama} + Diffusion~\cite{sdv1_4}\} models with CLIP, showcased in the third column, improves segmentation quality and minimizes artifacts. This enhancement results from the introduction of a self-training loss, leveraging the original pseudo labels provided by the Teacher model.
In the fourth column, we see the final enhancement: incorporating SAM to improve pseudo labels used in self-training. This leads to the distinctive sharp and detailed segmentation maps of our finalized model, \method.
This step-by-step enhancement highlights the effectiveness of sequentially adding foundation models, each contributing to increasingly accurate and detailed segmentation outcomes.

\section{Pseudo label refinement using SAM}
\begin{figure*}[!t]
    \centering
          \begin{subfigure}[b]{0.25\linewidth}
            \includegraphics[width=\linewidth]{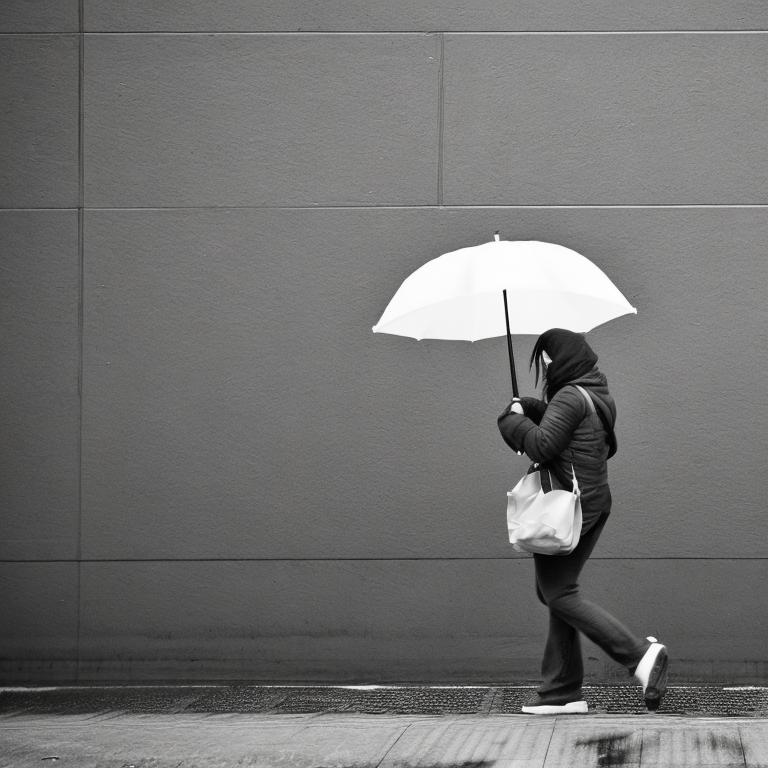}
          \end{subfigure}
          \hfill
          \begin{subfigure}[b]{0.25\linewidth}
            \includegraphics[width=\linewidth]{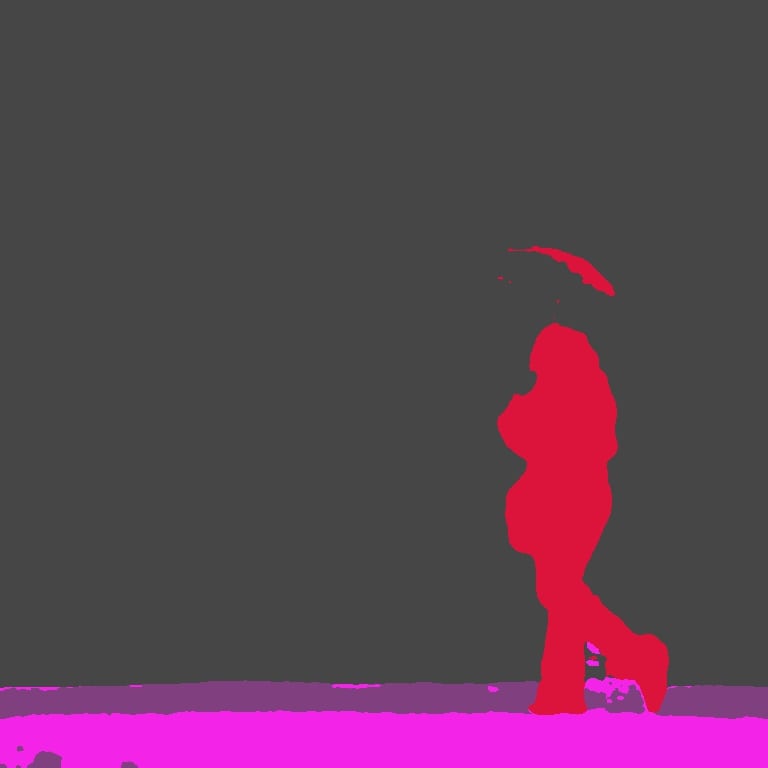}
          \end{subfigure}
          \hfill
          \begin{subfigure}[b]{0.25\linewidth}
            \includegraphics[width=\linewidth]{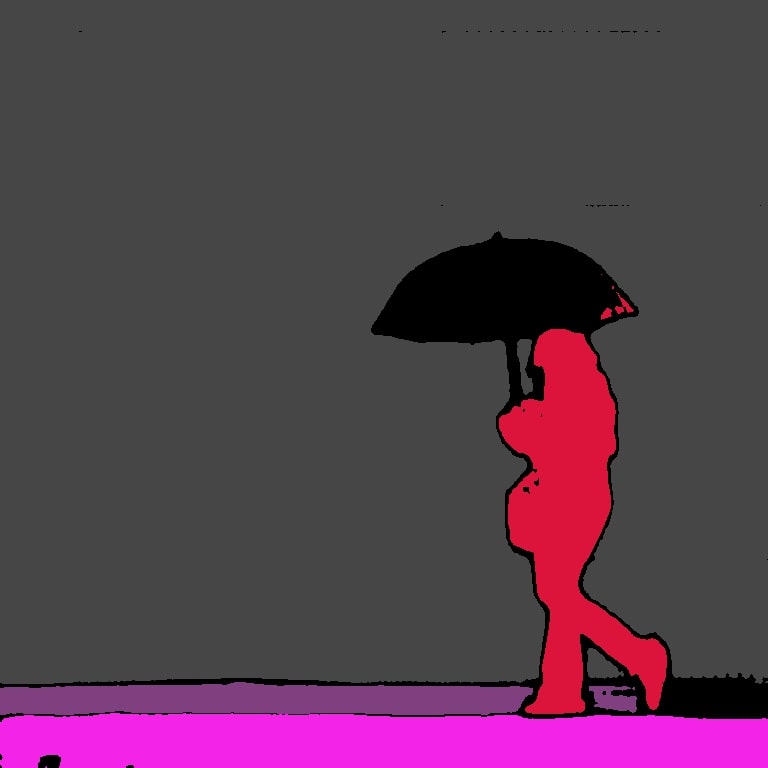}
          \end{subfigure}
          \hfill\\
          \begin{subfigure}[b]{0.25\linewidth}
            \includegraphics[width=\linewidth]{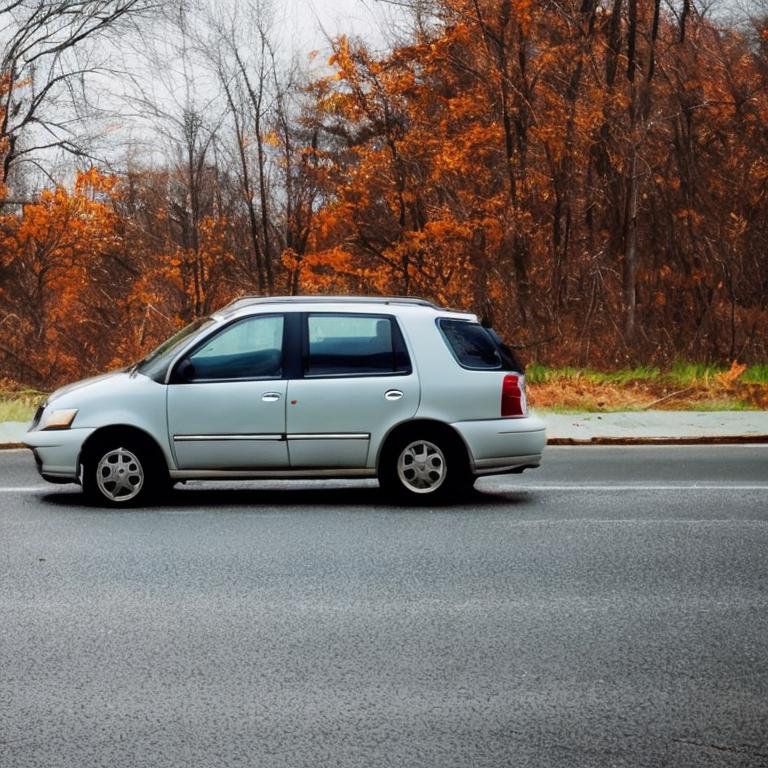}
          \end{subfigure}
          \hfill
          \begin{subfigure}[b]{0.25\linewidth}
            \includegraphics[width=\linewidth]{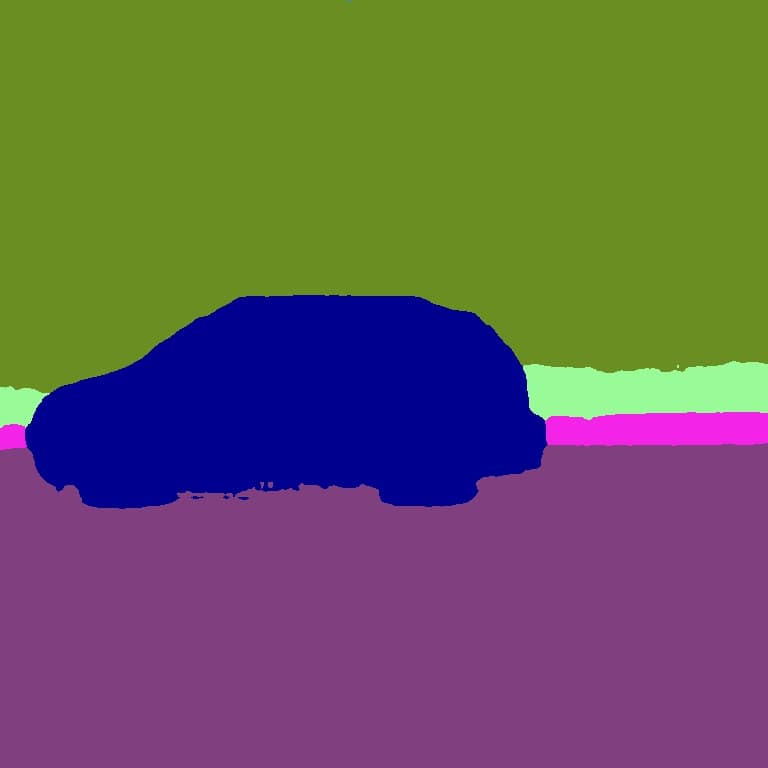}
          \end{subfigure}
          \hfill
          \begin{subfigure}[b]{0.25\linewidth}
            \includegraphics[width=\linewidth]{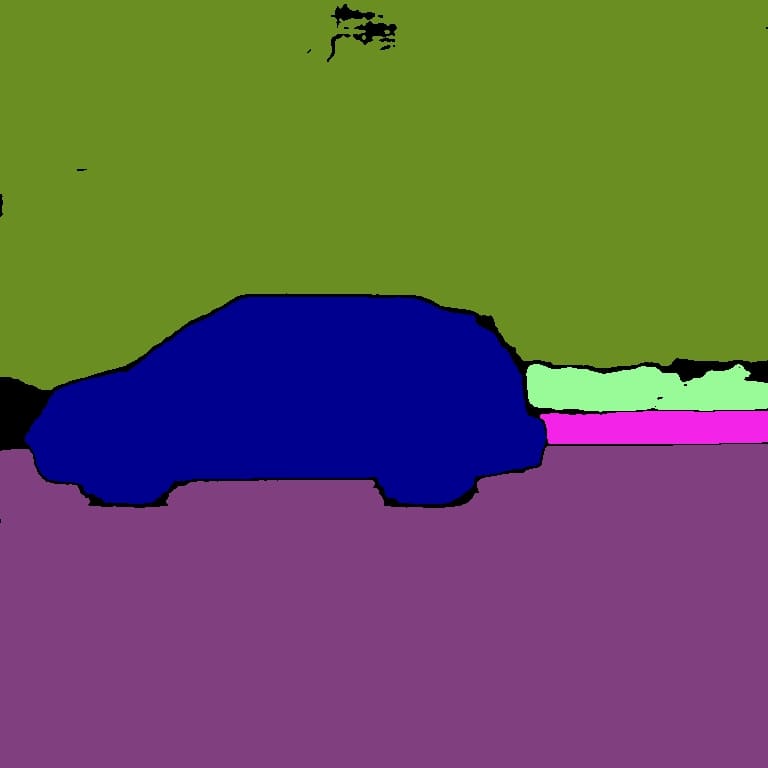}
          \end{subfigure}
          \hfill\\
          \begin{subfigure}[b]{0.25\linewidth}
            \includegraphics[width=\linewidth]{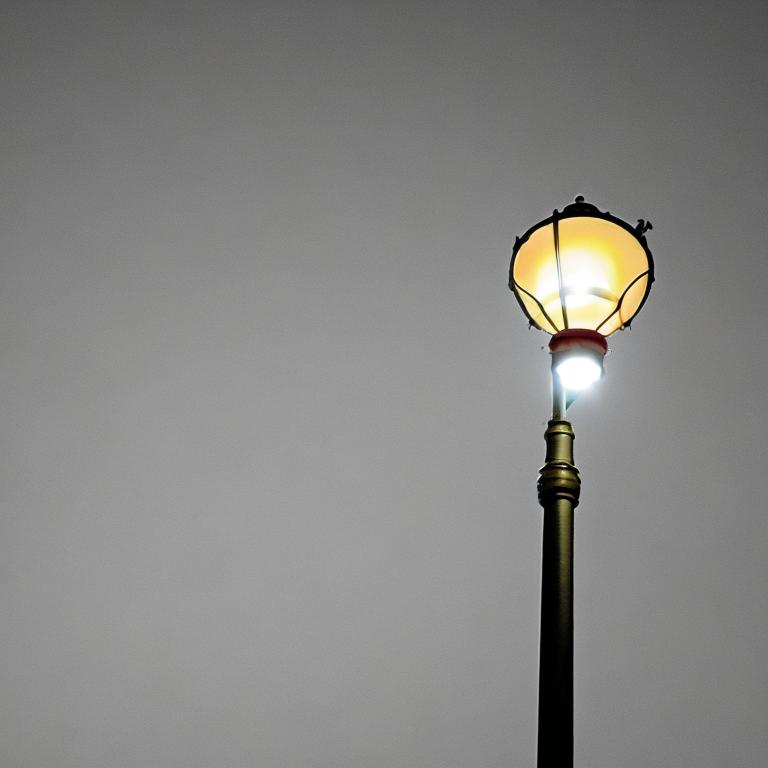}
          \end{subfigure}
          \hfill
          \begin{subfigure}[b]{0.25\linewidth}
            \includegraphics[width=\linewidth]{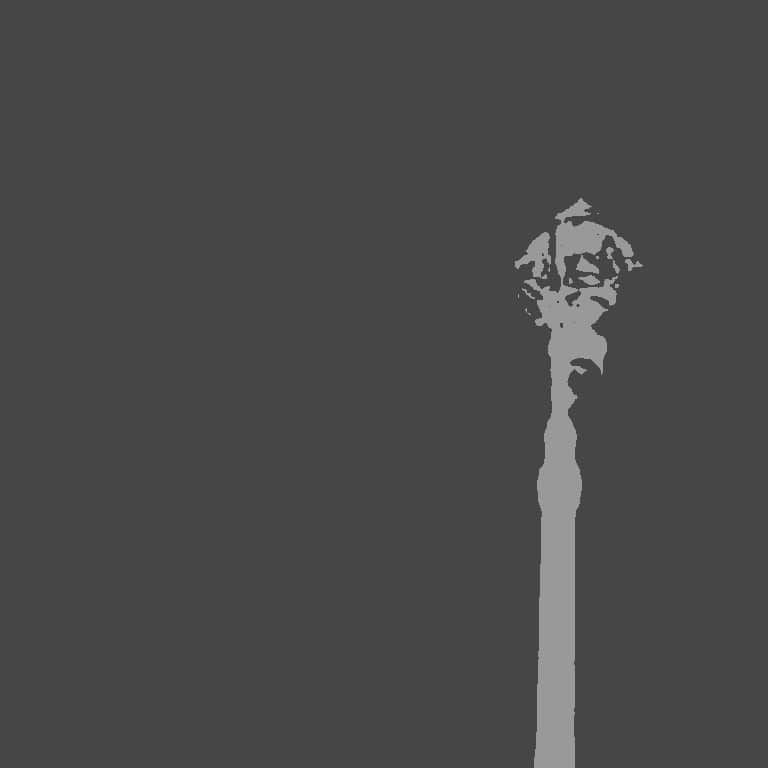}
          \end{subfigure}
          \hfill
          \begin{subfigure}[b]{0.25\linewidth}
            \includegraphics[width=\linewidth]{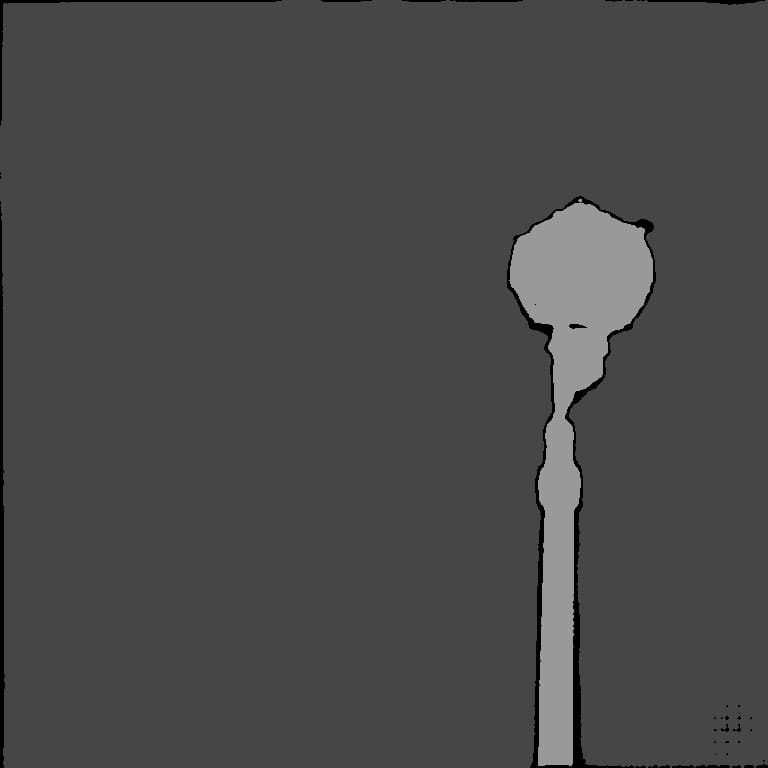}
          \end{subfigure}
          \hfill\\
          \begin{subfigure}[b]{0.25\linewidth}
            \includegraphics[width=\linewidth]{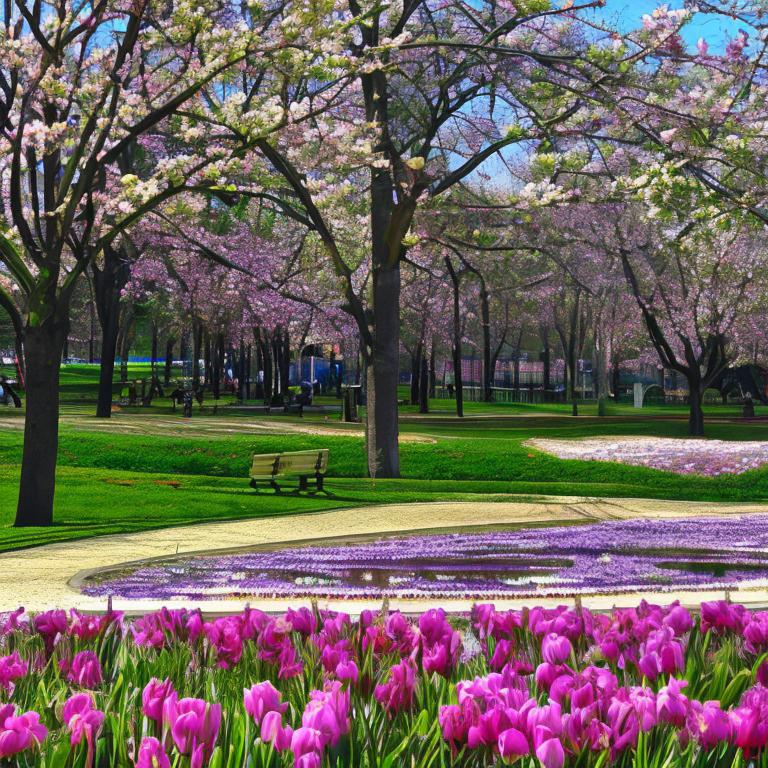}
            \subcaption{RGB image}
          \end{subfigure}
          \hfill
          \begin{subfigure}[b]{0.25\linewidth}
            \includegraphics[width=\linewidth]{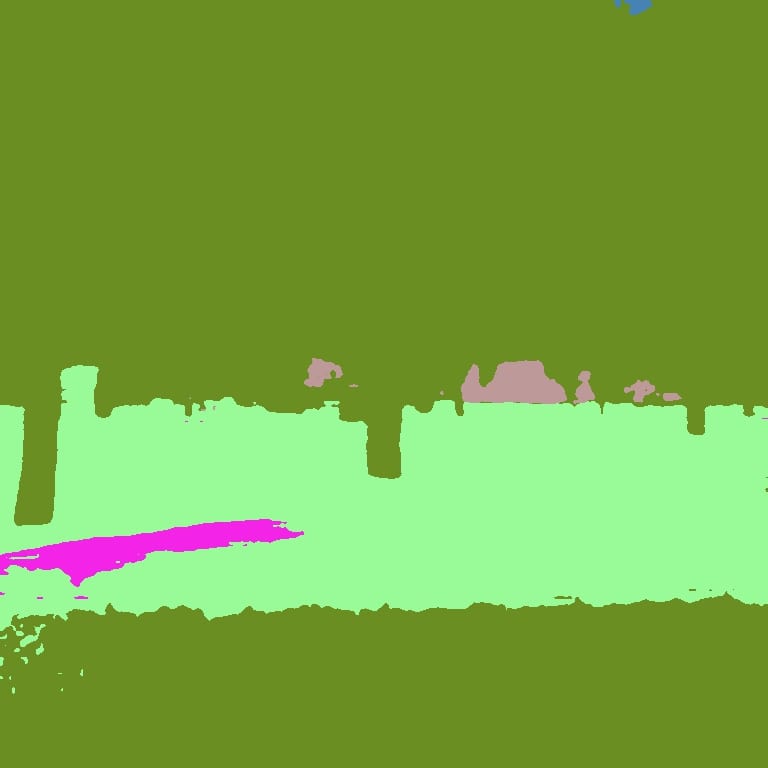}
            \subcaption{Pseudo label before SAM}
          \end{subfigure}
          \hfill
          \begin{subfigure}[b]{0.25\linewidth}
            \includegraphics[width=\linewidth]{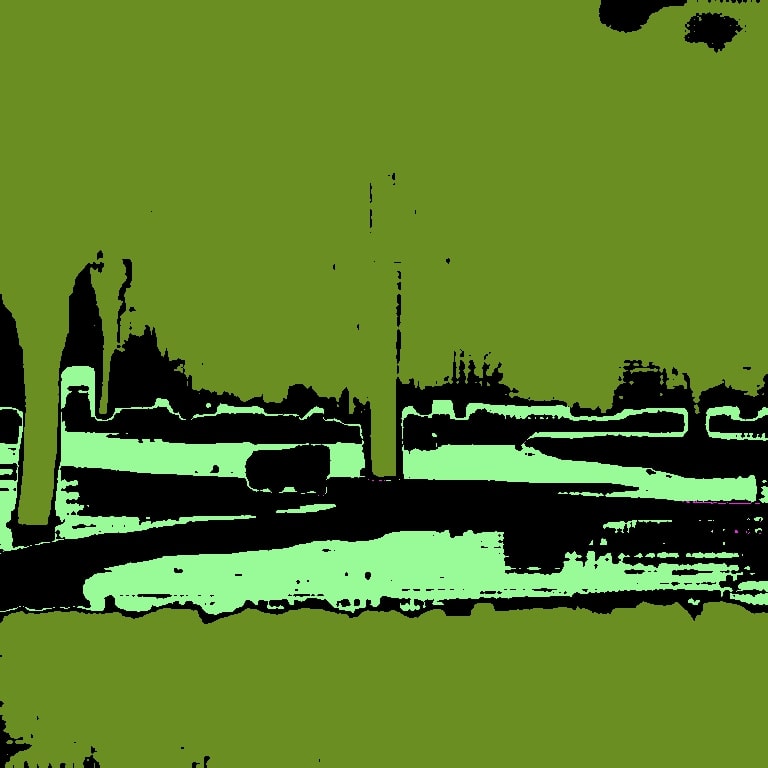}
            \subcaption{Pseudo label after SAM}
          \end{subfigure}
          \hfill\\

    \caption{Pseudo label refinement using SAM. We show: (a) The generated image using the Diffusion model conditioned by the LLM, (b) the pseudo label before using SAM, and (c) the pseudo label after the refinement by SAM}
    \label{fig:sam_ref}
\end{figure*}

In Figure \ref{fig:sam_ref}, we show some examples of initial and refined pseudo-labels. This comparison reveals that the refined pseudo labels have more accurate object boundaries. The refinement process can effectively eliminate ambiguous objects, such as the "umbrella" depicted in the first row of Figure \ref{fig:sam_ref}. However, this method is not without its limitations. Occasionally, it incorrectly designates some pixels as undefined (marked in black), excluding them from supervision despite their correct initial segmentation. It tends to discard numerous pixels, especially at the intersections of masks. Despite this, the pseudo label refinement reduces false positives within the pseudo labels.

\label{sec:sam-quali}

\section{Generated images using the Diffusion model}
\label{sec:dm-quali}
\begin{figure*}[!t]
    \centering
          \begin{subfigure}[b]{0.25\linewidth}
            \includegraphics[width=\linewidth]{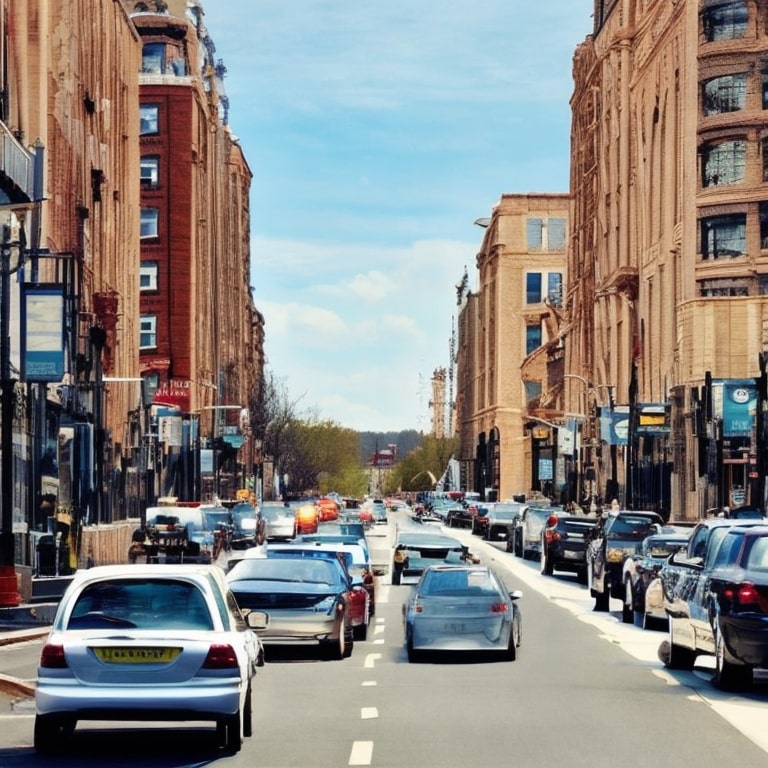}
            \subcaption{A photo of a busy road in daylight with sunny weather.}
          \end{subfigure}
          \hfill
          \begin{subfigure}[b]{0.25\linewidth}
            \includegraphics[width=\linewidth]{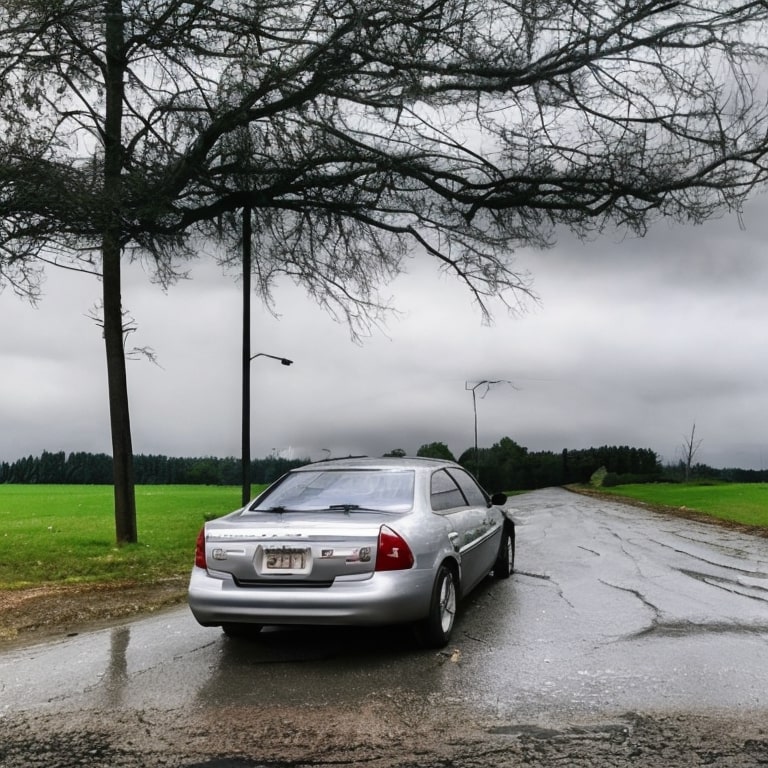}
            \subcaption{A picture of a parked car on the side of the road on a cloudy day.}
          \end{subfigure}
          \hfill
          \begin{subfigure}[b]{0.25\linewidth}
            \includegraphics[width=\linewidth]{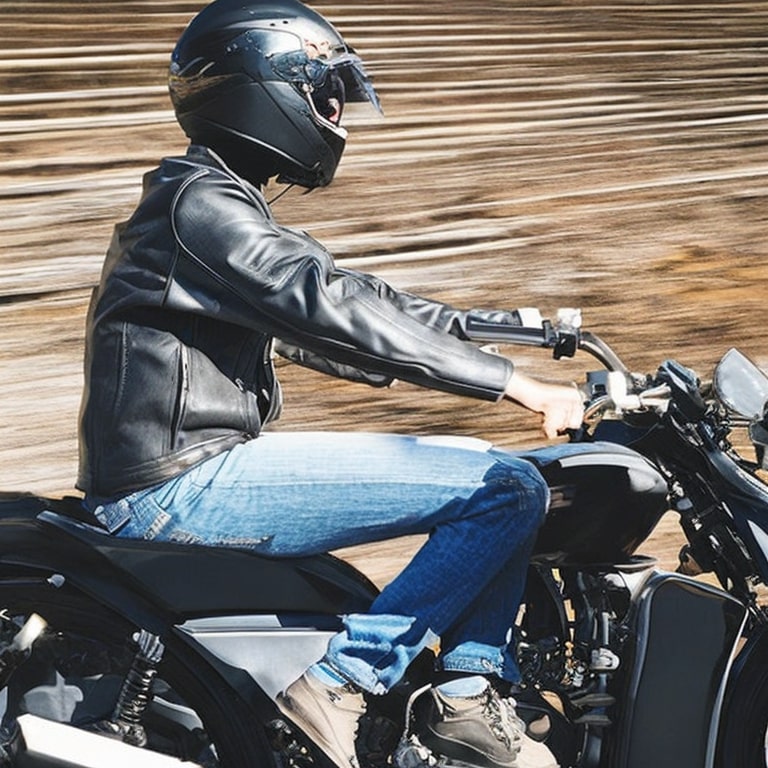}
            \subcaption{A picture of a person on a motorcycle in the sun.}
          \end{subfigure}
          \hfill\\
          \begin{subfigure}[b]{0.25\linewidth}
            \includegraphics[width=\linewidth]{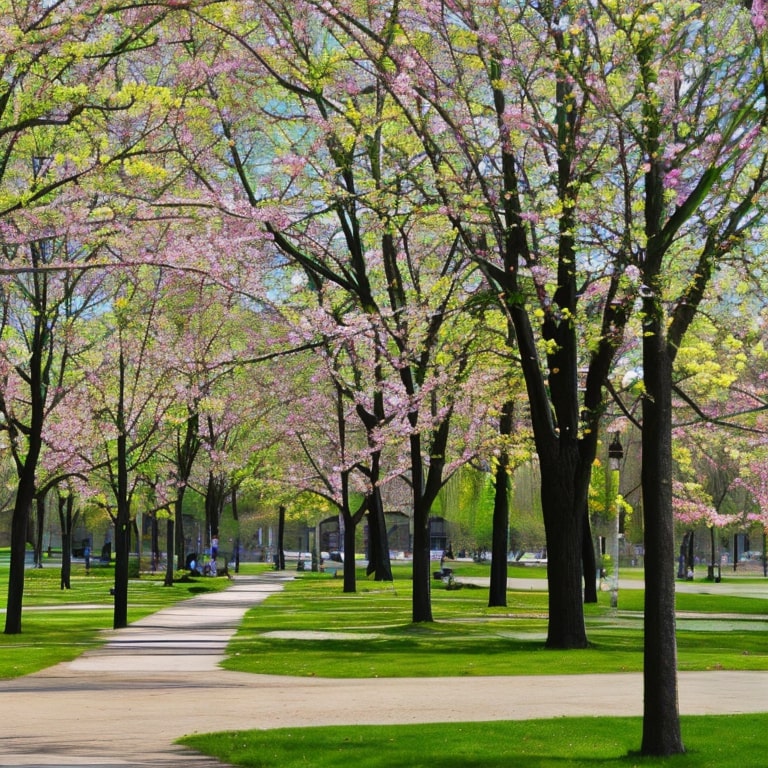}
            \subcaption{A picture of a city park in the spring on a clear day}
          \end{subfigure}
          \hfill
          \begin{subfigure}[b]{0.25\linewidth}
            \includegraphics[width=\linewidth]{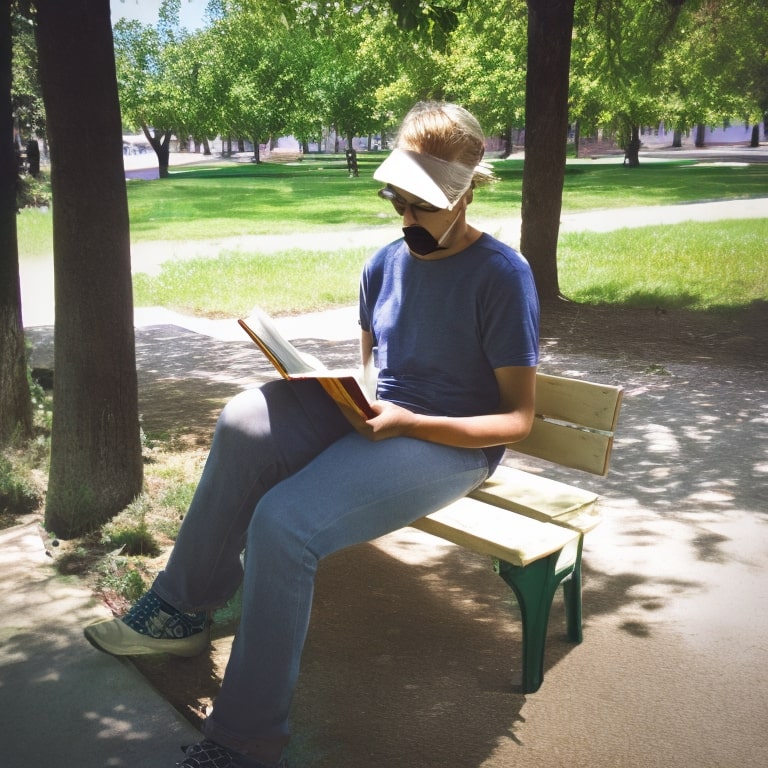}
            \subcaption{A photo of a person sitting on a bench reading a book in the shade.}
          \end{subfigure}
          \hfill
          \begin{subfigure}[b]{0.25\linewidth}
            \includegraphics[width=\linewidth]{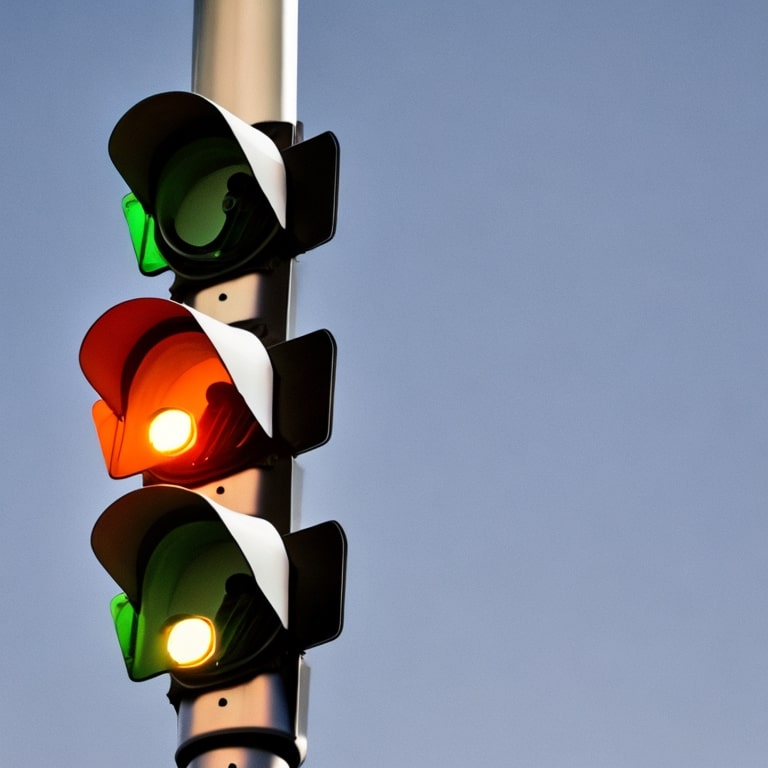}
            \subcaption{A photo of a traffic light in the morning}
          \end{subfigure}
          \hfill\\
          \begin{subfigure}[b]{0.25\linewidth}
            \includegraphics[width=\linewidth]{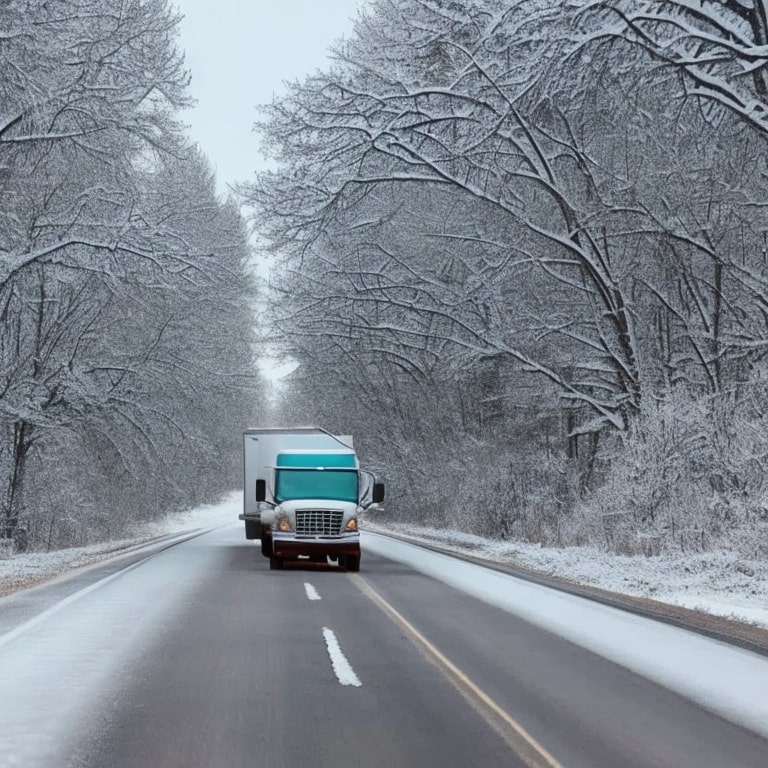}
            \subcaption{A snapshot of a truck driving down the road in the winter}
          \end{subfigure}
          \hfill
          \begin{subfigure}[b]{0.25\linewidth}
            \includegraphics[width=\linewidth]{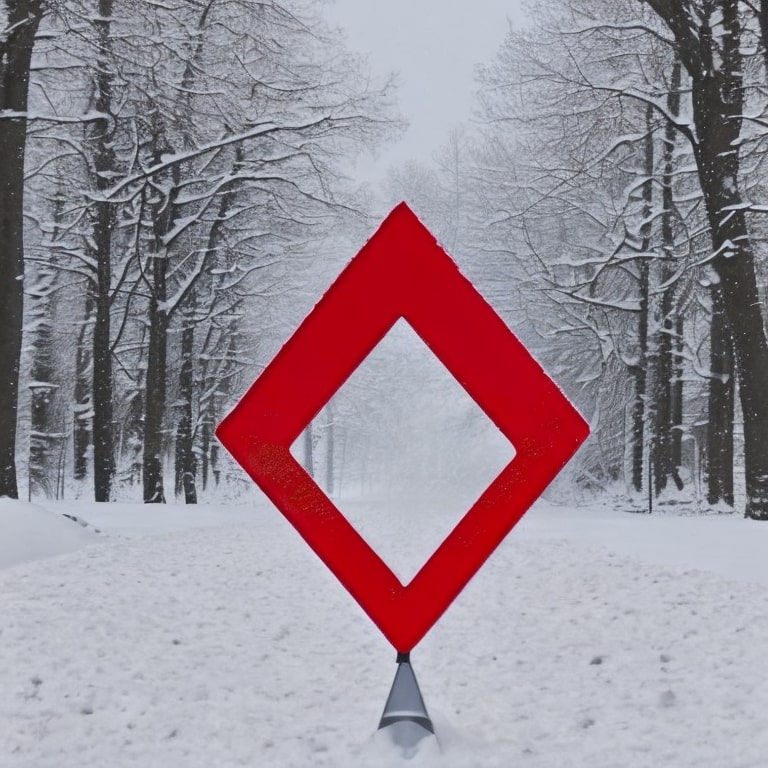}
            \subcaption{A photo of a traffic sign in the shape of a yield sign in the snow}
          \end{subfigure}
          \hfill
          \begin{subfigure}[b]{0.25\linewidth}
            \includegraphics[width=\linewidth]{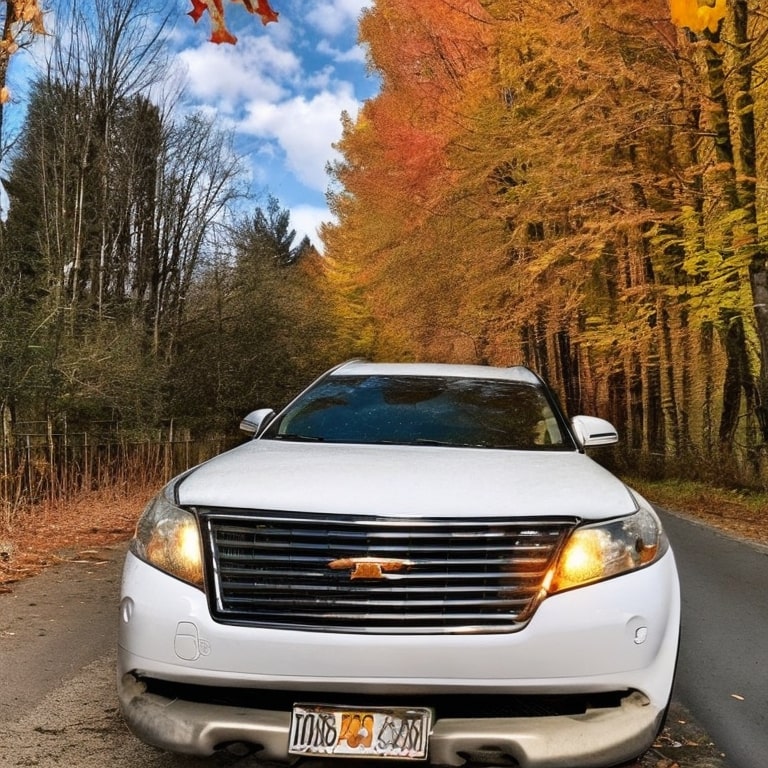}
            \subcaption{A picture of a car parked on the side of the road in the fall}
          \end{subfigure}
          \hfill\\
          \begin{subfigure}[b]{0.25\linewidth}
            \includegraphics[width=\linewidth]{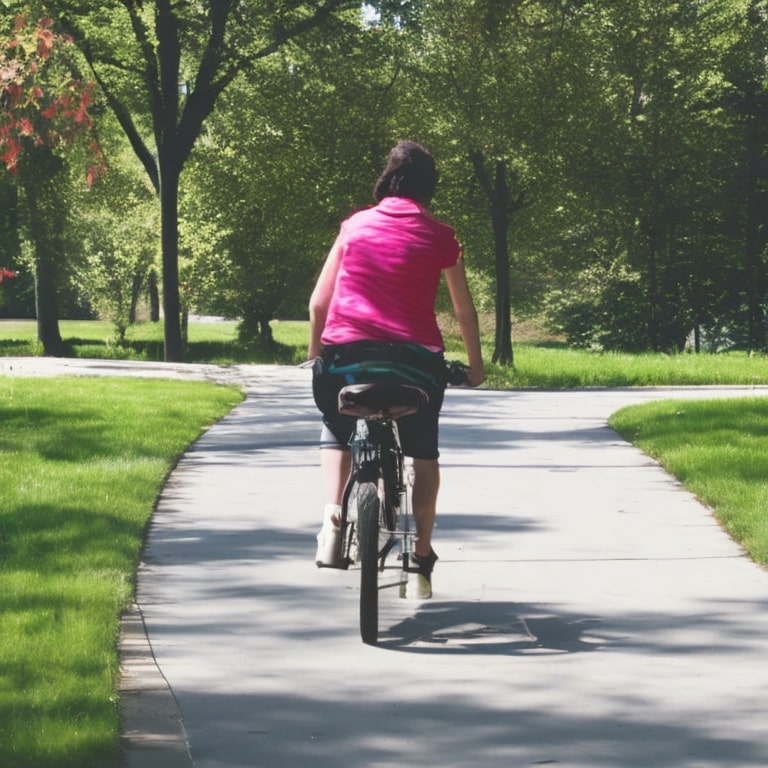}
            \subcaption{A photo of a person riding a bicycle in the park on a sunny day}
          \end{subfigure}
          \hfill
          \begin{subfigure}[b]{0.25\linewidth}
            \includegraphics[width=\linewidth]{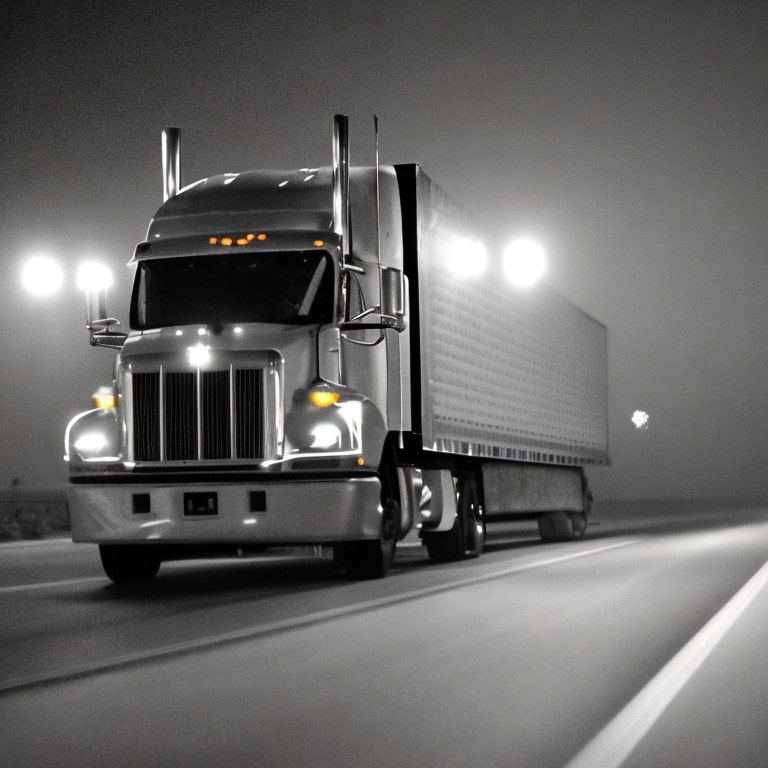}
            \subcaption{A snapshot of a truck driving down a highway in the night}
          \end{subfigure}
          \hfill
          \begin{subfigure}[b]{0.25\linewidth}
            \includegraphics[width=\linewidth]{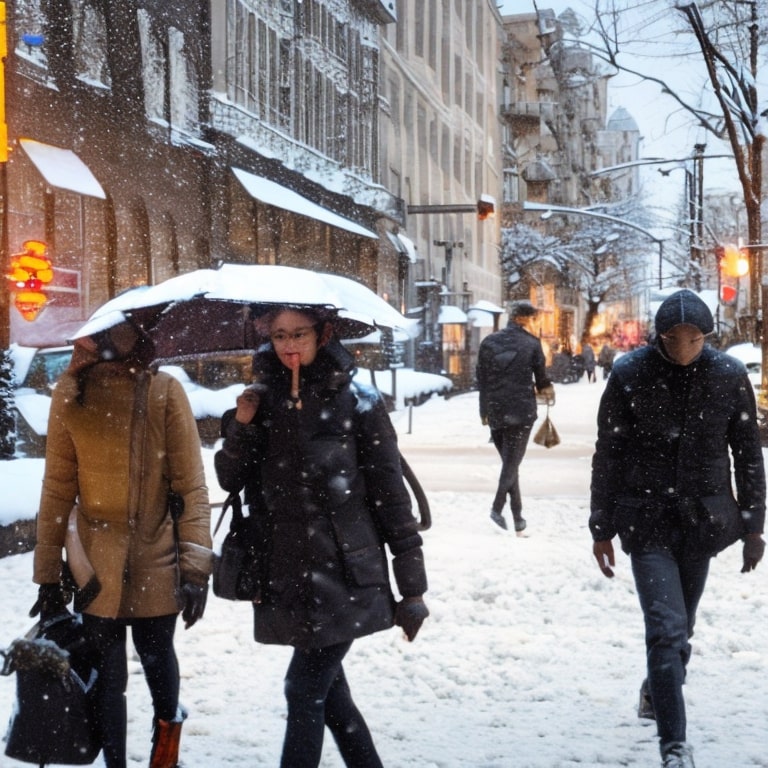}
            \subcaption{A photo of a group of pedestrians crossing the street in the snow}
          \end{subfigure}
          \hfill\\
    \caption{Examples of generated images using the diffusion model along with the text conditioning prompt generated by the LLM}
    \label{fig:gen_img}
\end{figure*}

In Figure \ref{fig:gen_img}, we show some generated images using the diffusion model along with their caption that were generated using LLM and used as text conditioning during generation.

\end{document}